%% file: main.tex
\documentclass[onefignum,onetabnum]{siamonline220329}

\usepackage{pifont}
\usepackage{lipsum}
\usepackage{amsfonts}
\usepackage{graphicx}
\usepackage{epstopdf}
\usepackage{algorithmic}
\usepackage{amssymb}
\usepackage{amsmath}
\usepackage{hyperref}       
\usepackage{booktabs}       
\usepackage{nicefrac}       
\usepackage{microtype}      
\usepackage{xcolor}         
\usepackage{multicol}
\usepackage{subcaption}
\usepackage{bbm}
\usepackage{multirow}
\usepackage{verbatim}

\newcommand{\norm}[1]{\left\lVert#1\right\rVert}
\newcommand\inner[2]{\left\langle #1, #2 \right\rangle}
\newcommand{\prth}[1]{\left(#1\right)}
\newcommand{\brac}[1]{\left[#1\right]}
\newcommand{\set}[1]{\left\{#1\right\}}

\newcommand{\mc}{\mathcal}
\newcommand{\mbb}{\mathbb}
\newcommand{\mb}{\mathbf}

\newcommand{\pt}{{\pi_\theta}}
\newcommand{\ptt}{{\pi_{\theta_t}}}
\newcommand{\dt}{{\diamond, t}}
\newcommand{\dtmone}{{\diamond, t-1}}

\ifpdf
\hypersetup{
  pdftitle={Policy-based Methods for Concave CMDP},
  pdfauthor={D. Ying, MA. Guo, H. Lee, Y. Ding, J. Lavaei, ZM. Shen}
}
\fi

\ifpdf
  \DeclareGraphicsExtensions{.eps,.pdf,.png,.jpg}
\else
  \DeclareGraphicsExtensions{.eps}
\fi

\usepackage{enumitem}
\setlist[enumerate]{leftmargin=.5in}
\setlist[itemize]{leftmargin=.5in}

\newsiamremark{remark}{Remark}
\newsiamremark{hypothesis}{Hypothesis}
\crefname{hypothesis}{Hypothesis}{Hypotheses}
\newsiamthm{claim}{Claim}
\newtheorem{example}{Example}
\newtheorem{assumption}{Assumption}

\headers{Policy-based Methods for Concave CMDP}{D. Ying, MA. Guo, H. Lee, Y. Ding, J. Lavaei, ZM. Shen}

\title{Policy-based Primal-Dual Methods for Concave CMDP with Variance Reduction\thanks{This work is an extension of \cite{ying2023policy}, which only derived the convergence rate of Primal-Dual Policy Gradient algorithm (PDPG) in the exact setting.
In this work, however, we study a more practical setting where the gradients of the Lagrangian need to be estimated from data.
Under this sample-based setting, we incorporate variance reduction techniques and propose the Variance-Reduced PDPG algorithm (VR-PDPG), whose sample complexity is rigorously analyzed.
We also conduct new experiments to validate the effectiveness of VR-PDPG.
}}

\author{Donghao Ying\thanks{Department of Industrial Engineering and Operations Research, UC Berkeley, CA (\email{donghaoy, mengzi\_guo, yuhao\_ding, lavaei, maxshen@berkeley.edu}).}
\and Mengzi Amy Guo\footnotemark[2]
\and Hyunin Lee\thanks{Department of Mechanical Engineering, UC Berkeley, CA (\email{hyunin@berkeley.edu}).}
\and Yuhao Ding\footnotemark[2]
\and  Javad Lavaei\footnotemark[2]
\and Zuo-Jun Max Shen\footnotemark[2]}

\usepackage{amsopn}

\begin{document}

\maketitle

\begin{abstract}
We study Concave Constrained Markov Decision Processes (Concave CMDPs) where both the objective and constraints are defined as concave functions of the state-action occupancy measure. We propose the Variance-Reduced Primal-Dual Policy Gradient Algorithm (VR-PDPG), which updates the primal variable via policy gradient ascent and the dual variable via projected sub-gradient descent. Despite the challenges posed by the loss of additivity structure and the nonconcave nature of the problem, we establish the global convergence of VR-PDPG by exploiting a form of hidden concavity. In the exact setting, we prove an $\mathcal{O}(T^{-1/3})$ convergence rate for both the average optimality gap and constraint violation, which further improves to $\mathcal{O}(T^{-1/2})$ under strong concavity of the objective in the occupancy measure. In the sample-based setting, we demonstrate that VR-PDPG achieves an $\widetilde{\mc{O}}(\epsilon^{-4})$ sample complexity for $\epsilon$-global optimality. Moreover, by incorporating a diminishing pessimistic term into the constraint, we show that VR-PDPG can attain a zero constraint violation without compromising the convergence rate of the optimality gap. Finally, we validate the effectiveness of our methods through numerical experiments.
\end{abstract}




\input{file/1_intro}
\input{file/2_formulation}
\input{file/3_algorithm}

\input{file/4_convergence}

\input{file/5_zero_violation}

\input{file/6_experiment2}
\input{file/7_conclusion}


\bibliographystyle{siamplain}
\bibliography{references}

\newpage
\appendix
\input{app/appendix_1}

\input{app/appendix_2_3}
\input{app/appendix_4}
\input{app/appendix_5}

\input{app/appendix_aux}

\input{app/appendix_exp}

\end{document}

%% file: file/1_intro.tex
\section{Introduction}\label{sec:intro}
Reinforcement Learning (RL), typically formulated by Markov Decision Processes (MDPs), focuses on learning a policy that maps situations to actions to maximize the expected cumulative sum of rewards. 
Mathematically, this objective can be reformulated as maximizing the inner product between the policy-induced state-action occupancy measure and a policy-independent reward for each state-action pair. 
Nonetheless, many decision-making problems of practical interests take a form beyond the cumulative reward.
For instance, in apprenticeship learning, the objective shifts towards minimizing the distance between the occupancy measures of an expert and the learning agent \cite{abbeel2004apprenticeship}.
In the pursuit of diverse skill discovery, the agent seeks to maximize the Kullback–Leibler (KL) divergence between the state occupancy measures that correspond to different skills \cite{eysenbach2018diversity}.
Additionally, in pure exploration problems, the goal becomes maximizing the entropy/or a KL-divergence of the state occupancy measure \cite{hazan2019provably,lee2019efficient, mutti2022importance}.
Recently, \cite{zhang2020variational} and \cite{zahavy2021reward} abstract such problems as Concave (Convex) Markov Decision Processes, which focus on finding a policy to maximize (minimize) a concave (convex) function of the induced state-action occupancy measure.
We refer the reader to \cite{zhang2020variational,zahavy2021reward, mutti2022challenging} and the references therein for more examples of concave MDPs.

However, the application of concave MDP frameworks in safety-critical domains---such as autonomous driving \cite{fisac2018general}, cyber-security \cite{zhang2019non}, and financial management \cite{abe2010optimizing}---introduces additional complexity.
To this end, we consider the Concave Constrained Markov Decision Process (Concave CMDP) problem where both the objective and constraints are defined as concave functions of the state-action occupancy measure.
This formulation strictly generalizes the classical safe RL and CMDPs \cite{altman1999constrained}, which assume that the objective and constraints are expressed by cumulative rewards, i.e., being linear in the state-action occupancy measure.

\begin{table}[!tb]
\centering
{\small
\renewcommand{\arraystretch}{1.5}
\begin{tabular}{c|c|c|c|c}
\hline
               \textbf{Setting}                & \textbf{Algorithm}    & \textbf{Objective} & \textbf{Optimality Gap} & \textbf{Constraint Violation} \\ \hline \hline
\multirow{2}{*}{Exact} & \multirow{2}{*}{PDPG} & Concave            & $\mc{O}\left(T^{-1/3}\right)$                    & $\mc{O}\left(T^{-1/3}\right)$                             \\ \cline{3-5} 
                               &                       & Strongly Concave   & $\mc{O}\left(T^{-1/2}\right)$                       & $\mc{O}\left(T^{-1/2}\right)$                             \\ \hline\hline
Sample-based           & VR-PDPG               & Concave            & $\mc{O}\left(T^{-1/4}\right)$                        & $\mc{O}\left(T^{-1/4}\right)$                             \\ \hline
\end{tabular}
 \caption{\small{We summarize our results for concave CMDPs, where the concavity is in terms of the occupancy measure and $T$ denotes the total number of iterations. 
 The number of samples required by VR-PDPG is $\mc{O}(\log T)$ in each iteration, thereby implying the overall sample complexity of $\widetilde{\mc{O}}(\epsilon^{-4})$ to achieve an $\mc{O}(\epsilon)$ optimality gap and constraint violation.
 For all cases in the table, a zero constraint violation can be achieved while maintaining the same order of convergence rate for the optimality gap.
 \label{table: comparison}}}
 }
\vspace{-1cm}
\end{table}

In this paper, we study concave CMDPs from the optimization perspective and aim to develop a theoretical foundation for direct policy search methods.
When moving beyond linear/additive structures in the objective and the constraints, we quickly face several technical challenges. 
Firstly, the problem has a nonconcave objective and nonconvex constraints with respect to the policy even under the simplest direct policy parameterization. 
Thus, the existing tools from the convex constrained optimization literature are not applicable.
Secondly, the gradient of the objective/constraint with respect to the occupancy measure becomes policy-dependent, thereby making the evaluation for the single-step improvement of the algorithm harder without knowing the occupancy measure.
Yet, accurately computing the occupancy measure for a given policy can be inefficient \cite{Tsybakov2008}.
Thirdly, the performance difference lemma introduced by \cite{kakade2002approximately}, which is key to the analysis of policy-based primal-dual methods for standard CMDPs \cite{ding2020natural,liu2021policy}, is no longer helpful for concave CMDPs. 

In view of the aforementioned challenges, our main contributions to the policy search of (discounted infinite-horizon) concave CMDPs are summarized in Table \ref{table: comparison} and provided below:
\begin{itemize}[leftmargin = *]
    \item Despite being nonconcave with respect to the policy and nonlinear with respect to the state-action occupancy measure, we prove that the strong duality still holds for concave CMDP problems under some mild conditions.
    \item In the exact setting, we propose a simple but effective algorithm---Primal-Dual Policy Gradient method (PDPG) that updates the primal variable via policy gradient ascent and the dual variable via projected sub-gradient descent.
    In the sample-based setting, we propose the Variance-Reduced PDPG method (VR-PDPG) by incorporating momentum-based recursive estimators for the occupancy measure and the policy gradient.
    Strong bounds on the optimality gap and the constraint violation are established for both algorithms, and the effectiveness of variance reduction is illustrated by numerical experiments.
    \item Inspired by the idea of ``optimistic pessimism in the face of uncertainty'', we further demonstrate that by including a diminishing pessimistic term to the constraint, a zero constraint violation can be achieved in both exact and sample-based settings while maintaining the same order of convergence rate for the optimality gap.
\end{itemize}

\subsection{Related work}
\vspace{3pt}
\paragraph{RL with General Utilities}
Motivated by emerging applications in RL that extend beyond cumulative rewards \cite{schaal1996learning,abbeel2004apprenticeship,ho2016generative,hazan2019provably,rosenberg2019online,lee2019efficient}, recent research has focused on developing general approaches for concave (or convex) MDPs.
Specifically, \cite{zhang2020variational} introduces a new policy gradient approach called variational policy gradient, and by exploiting the hidden concavity of the problem, it establishes the global convergence of the gradient ascent method in the exact setting. 
Following this, \cite{zhang2021convergence} studies a REINFORCE-based policy gradient method and its variance-reduced version, where the sample complexity of the algorithm is analyzed.
More recently, \cite{barakat2023reinforcement} achieves a similar sample complexity result with a new single-loop variance-reduced algorithm.
Additionally, \cite{zahavy2021reward} transforms the concave MDP problem to a saddle-point problem using Fenchel duality and proposes a meta-algorithm to solve the problem with standard RL techniques. \cite{geist2021concave} proves the equivalence between concave MDPs and mean-field games (MFGs) and shows that algorithms for MFGs can be used to solve concave MDPs. 
However, these studies only address unconstrained RL problems, which may lead to undesired policies in safety-critical applications. 
Therefore, additional effort is required to deal with the emerging safety constraints, and our work addresses this challenge.

\paragraph{CMDP}
Our work is also pertinent to policy-based CMDP algorithms \cite{altman1999constrained,borkar2005actor,achiam2017constrained,ding2022provably,chow2017risk,efroni2020exploration}.
Notably, \cite{ding2020natural} develops a natural policy gradient-based primal-dual algorithm, demonstrating an $\mc{O}(T^{-1/2})$ global convergence rate for both the optimality gap and the constraint violation under standard soft-max parameterization.
\cite{xu2021crpo} considers a primal-based approach, achieving a similar convergence rate.
More recently, the papers \cite{ying2021dual,liu2021policy,li2024faster} incorporate entropy regularization and obtain improved convergence rates with dual methods.
Nonetheless, these papers focus on cumulative rewards/utilities and do not directly generalize to a broader class of safe RL problems, such as safe imitation learning \cite{zhou2018safety} and safe exploration \cite{hazan2019provably}.
Beyond CMDPs with cumulative rewards/utilities, \cite{ying2024scalable, bai2023achieving} also study the concave CMDP problem. 
Specifically, \cite{ying2024scalable} considers a multi-agent setting and designs a scalable algorithm that converges to a stationary point of the problem.
The algorithm by \cite{bai2023achieving} is based on the randomized linear programming method proposed by \cite{wang2020randomized} and it achieves the sample complexity of $\widetilde{\mc{O}}(1/\epsilon^2)$.
However, as their approach works directly in the space of state-action occupancy measures, it is thus not applicable to more general problems where the state-action spaces are large and a function approximation is needed. 
In comparison, our work addresses this issue by focusing on the policy-based primal-dual method and adopting a general soft-max policy parameterization.

\subsection{Notations}
For a finite set $\mc{S}$, let $\Delta(\mathcal{S})$ denote the probability simplex over $\mathcal{S}$, and let $\left|\mathcal{S}\right|$ denote its cardinality.
For a number $x$, let $[x]_+:=\max\{x,0\}$.
For a random variable $x$ following distribution $\rho$, we write it as $x\sim \rho$.
Let $\mbb{E}[\cdot]$ (resp. $\mbb{P}[\cdot]$) and $\mbb{E}[\cdot\mid \cdot]$ (resp. $\mbb{P}[\cdot \mid \cdot]$) denote the expectation (resp. probability) and conditional expectation (resp. probability) of a random variable, respectively.
For brevity, we also use the shorthand notation $\mbb{E}\brac{x}_+:= \mbb{E}\brac{\brac{x}_+}$.
Let $\mbb{R}$ denote the set of real numbers.
For a vector $x$, we use $x^\top$ to denote the transpose of $x$ and use $\langle x,y\rangle$ to denote the inner product $x^\top y$.
We use the convention that $\|x\|_1 = \sum_i |x_i|$, $\|x\| = \sqrt{\sum_i x_i^2}$, and $\|x\|_\infty = \max_i |x_i|$.
For a set $X$ in the Euclidean space, let $\operatorname{cl}(X)$ denote the closure of $X$.
Let $\mc{P}_X$ denote the projection onto $X$, defined as $\mathcal{P}_{X}(y):=\arg \min _{x \in X}\|x-y\|$.
For a matrix $A$, let $\|A\|$ stand for the spectral norm, i.e., $\|A\| = \max_{\|x\|\neq 0}\left\{ \|Ax\|/\|x\|\right\}$.
For a function $f(x)$, let $\operatorname{arg}\min f(x)$ (resp. \hspace{-0.8mm}$\operatorname{arg}\max f(x)$) denote any global minimum (resp. global maximum) of $f(x)$ and let $\nabla_x f(x)$ denote its gradient with respect to $x$.

\subsection{Organizations}
The rest of the paper is organized as follows.
In Section \ref{sec:formulation}, we formulate the problem of concave CMDP and introduce the Lagrangian duality used in the paper.
Then, in Section \ref{sec:alg}, we introduce the PDPG and 
VR-PDPG algorithms and discuss our main analysis technique.
Next, in Section \ref{sec:convergence}, we provide the convergence proofs for our algorithms, both in the exact and sample-based settings.
In Section \ref{sec:zero}, we further show that a zero constraint violation can be achieved without sacrificing the convergence rate of the optimality gap.
Finally, we validate our algorithms through numerical experiments in Section \ref{sec: Numerical Experiment} and close the main paper with concluding remarks in Section \ref{sec: conclusion}.

%% file: file/2_formulation.tex
\section{Problem Formulation}\label{sec:formulation}
\vspace{3pt}
\paragraph{Standard CMDP}
Consider an infinite-horizon CMDP over a finite state space $\mc{S}$ and a finite action space $\mc{A}$ with a discount factor $\gamma\in[0,1)$. 
Let $\rho$ be the initial distribution.
The transition dynamics is given by $\mbb{P}: \mc{S}\times \mc{A} \rightarrow \Delta (\mc{S})$, where $\mbb{P}(s^\prime \vert s,a)$ is the probability of transition from state $s$ to state $s^\prime$ when action $a$ is taken.
A policy is a function $\pi: \mathcal{S} \rightarrow \Delta(\mathcal{A})$ that represents the decision rule the agent uses, i.e., the agent takes action $a$ with probability $\pi(a\vert s)$ in state $s$.
We denote the set of all stochastic policies as $\Pi$.
The goal of the agent is to find a policy that maximizes some long-term objective.
In standard CMDPs, the agent aims at maximizing the expected (discounted) cumulative reward for a given initial distribution $\rho$ while satisfying constraints on the expected (discounted) cumulative cost, i.e.,
\begin{equation}\label{eq:cmdp}
\begin{aligned}
\max_{\pi\in \Pi}\ &V^\pi(r) := \mbb{E}\left[\sum_{k=0}^{\infty} \gamma^{k} r\left(s_{k}, a_{k}\right) \bigg| \pi, s_{0}\sim \rho \right],\\
\text{s.t. } &V^\pi(c) :=   \mbb{E}\left[\sum_{k=0}^{\infty} \gamma^{k} c\left(s_{k}, a_{k}\right) \bigg| \pi, s_{0}\sim \rho \right]\leq 0,
\end{aligned}
\end{equation}
where the expectation is taken over all possible trajectories under policy $\pi$ with initial distribution $\rho$, and $r(\cdot,\cdot)$ and $c(\cdot,\cdot)$ denote the reward and cost functions, respectively. We note that the cost can be a vector-valued function, not limited to a single constraint.
For given $r(\cdot,\cdot)$, we define the state-action value function (Q-function) under policy $\pi$ as
\begin{equation}
Q^\pi(r;s,a) := \mbb{E}\left[\sum_{k=0}^{\infty} \gamma^{k} r\left(s_{k}, a_{k}\right) \bigg| \pi, s_{0}=s, a_0=a \right],
\end{equation}
which can be interpreted as the expected total reward with an initial state $s_0 = s$ and an initial action $a_0 = a$.
For each policy $\pi\in \Pi$ and state-action pair $(s,a)\in\mc{S}\times \mc{A}$, the unnormalized state-action occupancy measure is defined as 
\begin{equation}
\lambda^\pi(s,a)=\sum_{k=0}^{\infty} \gamma^{k} \mathbb{P}\left(s_{k}=s, a_k = a\vert \pi, s_0 \sim \rho \right).
\end{equation}
We use $\Lambda$ to denote the set of all possible state-action occupancy measures, which is a convex polytope (see its expression in (\ref{eq:Lambda})).
By using $\lambda$ as decision variables, we observe that the CMDP problem in (\ref{eq:cmdp}) can be re-parameterized as:
\begin{equation}\label{eq:cmdp-LP}
\max_{\lambda \in \Lambda} \ \langle r,\lambda\rangle,\quad \text{s.t. } \langle c,\lambda\rangle\leq 0,
\end{equation}
where we note that the inner product is taken while viewing $\lambda$, $r$, and $c$ are vectors of dimension $|\mc{S}||\mc{A}|$.
Problem \ref{eq:cmdp-LP} is known as the linear programming formulation of CMDP \cite{altman1999constrained}.
Once a solution $\lambda^\star$ is computed, the corresponding policy can be recovered through the one-to-one relation
$\pi(a\vert s) = \lambda^\pi(s,a)/\sum_{a^\prime \in \mc{A}}\lambda^\pi(s,a^\prime)$.
\vspace{3pt}
\paragraph{Concave CMDP}
In this work, we consider a more general problem where the agent's goal is to find a policy that maximizes a concave function of the state-action occupancy measure $\lambda$ subject to constraints defined by another concave function in $\lambda$, namely
\begin{align} \label{eq: main problem}
\underset{\lambda \in \Lambda}\max\ f(\lambda) \quad \text {s.t. } g(\lambda) \geq  {0},
\end{align}
where $f(\cdot)$ and $g(\cdot)$ are general concave functions, named as \textit{general utilities}.
We note that problem \eqref{eq: main problem} encompasses \eqref{eq:cmdp-LP} as a special case with $f(\lambda) = \inner{r}{\lambda}$ and $g(\lambda)= -\inner{c}{\lambda}$.
As problem \eqref{eq: main problem} is a concave program in $\lambda$, we refer to the problem as \textit{Concave CMDP}.
We emphasize that the method proposed in this paper directly generalizes to multiple constraints, and we present the single constraint setting only for brevity. Please refer to Remark \ref{rmk: multiple_constraints} for the detailed discussion on this generalization. 
Below, we provide several examples to illustrate the effectiveness of the general utility and the concave CMDP formulation.

\begin{example}[Safe learning] 
Motivated by the applications of CMDPs in safety-critical systems, \cite{ni2023interior} develops an approach that guarantees constraint satisfaction during the learning process, in contrast to ordinary CMDP approaches that only ensure policy feasibility upon convergence. They reformulate the standard CMDP \eqref{eq:cmdp} using the log barrier function, and the objective can be described as the maximization of a generality utility, defined as follows
\begin{equation}
\max_{\lambda \in \Lambda} f(\lambda) = \inner{r}{\lambda} + \eta\log(-\inner{c}{\lambda}),
\end{equation}
where $\eta$ is a positive penalty weight.
\end{example}

\begin{example}[Safe exploration]
Many learning problems originate from an unsupervised setting, in which no explicit tasks are assigned. Under these conditions, agents are typically inclined to efficiently explore the environment \cite{hazan2019provably}. Yet, such explorations can impose significant safety concerns. 
The recent work by \cite{yang2023cem} introduces a model where the learning agent aims to maximize the entropy under the premise of safety, as presented below
\begin{equation}
\max_{\lambda\in \Lambda} f(\lambda) = \operatorname{Entropy}(\lambda)
\quad \text {s.t. }  g(\lambda) = -\inner{c}{\lambda} \geq 0, 
\end{equation}
where the entropy function under policy $\pi$ is defined as $\operatorname{Entropy}(\lambda^\pi) := -\sum_{s\in \mc{S}} d^\pi(s)\cdot \log(d^\pi(s))$ with $d^\pi(s):=(1-\gamma)\sum_{a^\prime\in \mc{A}} \lambda^\pi(s,a^\prime)$ denoting the discounted state occupancy measure.
This formulation decouples the safety constraints from the reward to alleviate the challenge of designing a single reward signal that must carefully balance task performance with safety.
\end{example}

\begin{example}[Safety-aware apprenticeship learning (AL)]
In AL, instead of maximizing the long-term reward, the agent learns to mimic an expert's demonstrations.
When there are critical safety requirements, the learner will also strive to satisfy given constraints on the expected total cost \cite{zhou2018safety}.
This problem can be formulated as
\begin{align} \label{eq:AL}
\underset{\lambda \in \Lambda}\max\ f(\lambda) = -\operatorname{dist}(\lambda,\lambda_e)\ \ \text {s.t. } g(\lambda) =  -\inner{c}{\lambda}\geq {0},
\end{align}
where $\lambda_e$ corresponds to the expert demonstration, $c$ denotes the cost function, and $\operatorname{dist}(\cdot,\cdot)$ can be any distance function on $\Lambda$, e.g., $\ell^2$-distance or Kullback-Liebler (KL) divergence. 
\end{example}

\begin{example}[Feasibility constrained MDPs]\label{exp:feasibility}
As an extension to standard CMDPs, the designer may desire to control the MDP through more general constraints described by a convex feasibility region $C$ \cite{miryoosefi2019reinforcement} (e.g., a single point representing a known safe policy) such that the learned policy is not too far away from $C$.
In this case, the problem can be cast as 
\begin{align} \label{eq:feasibility_MDP}
\underset{\lambda \in \Lambda}\max\ f(\lambda) =  \langle r,\lambda \rangle \ \ \text {s.t. }  g(\lambda) =  -\operatorname{dist}(\lambda,C)\geq -d_0,
\end{align}
where $d_0\geq 0$ denotes the threshold of the allowable deviation.
\end{example}

\paragraph{Policy Parameterization}
Since recovering a policy from the associated state-action occupancy measure is straightforward, a natural approach to solving the concave CMDP problem is to optimize (\ref{eq: main problem}) directly (or equivalently (\ref{eq:cmdp-LP}) for standard CMDPs).
However, since the decision variable $\lambda$ has the size ${|\mc{S}||\mc{A}|}$, such approaches lack scalability and can converge extremely slowly for large state and action spaces.
In this work, we consider the direct policy search method, which can handle the curse of dimensionality via policy parameterization.
We assume that the policy $\pi = \pi_\theta$ is parameterized by a general soft-max function, meaning that
\begin{equation}\label{eq:policy_parameterization}
\pi_{\theta}(a \vert s)=\dfrac{\exp\big(\psi(\theta;s, a)\big)}{\sum_{a^{\prime}\in \mc{A}} \exp \big(\psi\left(\theta;s, a^{\prime}\right)\big)},\ \forall (s,a)\in \mc{S}\times\mc{A},
\end{equation}
where $\psi(\cdot\ ;s,a)$ is some smooth function and $\theta\in\mbb R^K$ is the unconstrained parameter vector.
We assume that $\theta$ over-parameterizes the set of all stochastic policies in the sense that $\operatorname{cl}\big(\big\{\pi_\theta\vert \theta\in \mbb{R}^K\big\}\big) = \Pi$ and $\operatorname{cl}\big(\big\{\lambda^{\pi_\theta}\vert \theta\in \mbb{R}^K\big\}\big) = \Lambda$.
Further assumptions on the parameterization will be formally stated in Section \ref{sec:convergence}.
In practice, the function $\psi$ can be chosen as a deep neural network, where $\theta$ is the parameter and the state-action pair $(s,a)$ is the input.
Under the parameterization (\ref{eq:policy_parameterization}), problem (\ref{eq: main problem}) can be rewritten as
\begin{equation}\label{eq:problem_theta}
\max_{\theta\in \mbb{R}^K}\ f(\lambda(\theta)) \  \quad \text {s.t.}\ \  g(\lambda(\theta)) \geq 0,
\end{equation}
where we introduce the shorthand notations $\lambda(\theta) := \lambda^{\pi_\theta}$ and $\lambda(\theta;s,a):=\lambda^{\pi_\theta}(s,a)$.
It is worth mentioning that, differently from \eqref{eq: main problem}, the problem \eqref{eq:problem_theta} is nonconcave due to its nonconcave objective function and nonconvex constraint with respect to $\theta$.

\subsection{Lagrangian Duality}
Consider the Lagrangian function associated with (\ref{eq:problem_theta}), defined as $L(\lambda(\theta),\mu) := f(\lambda(\theta))+\mu g(\lambda(\theta))$.
The dual function is $D(\mu):= \max_{\theta\in \mbb{R}^K} L(\lambda(\theta),\mu)$.
Let $\pi_{\theta^\star}$ be an optimal policy such that $\theta^\star$ is an optimal solution to (\ref{eq:problem_theta}), and $\mu^\star$ be an optimal dual variable.
In constrained optimization, strict feasibility can induce many desirable properties.
We assume that the following Slater's condition holds.
\begin{assumption}[Slater's condition]\label{assump:slater}
There exist  $\widetilde{\theta}$ and $\xi >{0}$ such that $g(\lambda({\widetilde{\theta}})) \geq  \xi $.
\end{assumption}

Slater's condition is a standard assumption in convex optimization, and it holds when the feasible region has a non-empty interior.
In practice, identifying an interior point $\widetilde\theta$ is often straightforward, utilizing prior knowledge of the problem. 
This interior point ensures that the optimization problem satisfies the necessary regularity conditions for the duality theory to apply effectively.
The following result is a direct consequence of Slater's condition \cite{altman1999constrained}.

\begin{lemma}[Strong duality and boundedness of $\mu^\star$]\label{lemma:duality}
Let Assumption \ref{assump:slater} hold, and suppose that $\operatorname{cl}\big(\big\{\lambda(\theta)\vert \theta\in \mbb{R}^K\big\}\big) = \Lambda$, where $\operatorname{cl}(\cdot)$ denotes the closure operation. Then, we have that
\begin{equation}
    f\left(\lambda(\theta^\star)\right) = D\left(\mu^\star\right) = L\left(\lambda(\theta^\star),\mu^\star\right); \quad 0 \leq \mu^\star \leq \big[f(\lambda(\theta^\star))-f(\lambda(\widetilde\theta))\big]\big/\xi.
\end{equation}
Hence, it suffices to search the optimal dual variable within the closed interval $U:=[0, C_0]$, where $C_0 := 1+\big(M-f(\lambda(\widetilde \theta))\big)/{\xi}$ and $M$ is some upper bound on $f(\cdot)$.
\end{lemma}

For completeness, we provide a proof for Lemma \ref{lemma:duality} in Appendix \ref{app:2}.
We remark that in the definition of $C_0$, the constant $M$ will be formalized in Lemma \ref{lemma:boundedness_smoothness}, and the constant number $1$ plays as a slackness to facilitate the convergence analysis.
The strong duality implies that (\ref{eq:problem_theta}) is equivalent to the following saddle point problem:
\begin{equation}\label{eq:saddle}
\max_{\theta\in \mbb{R}^K }\min_{\mu \in U} L(\lambda(\theta),\mu) = \min_{\mu\in U} \max_{\theta\in \mbb{R}^K} L(\lambda(\theta),\mu).
\end{equation}
Motivated by this equivalence, we seek to develop a primal-dual algorithm to solve the problem.

%% file: file/3_algorithm.tex
\section{Safe Policy Search Beyond Cumulative Rewards/Utilities}\label{sec:alg}
To solve (\ref{eq:saddle}), we propose the following Primal-Dual Policy Gradient Algorithm (PDPG):
\begin{equation}\label{eq: algorithm_upate}
\left\{\begin{aligned}
\theta_{t+1} &= \theta_t +\eta_{\theta,t}\nabla_\theta L(\lambda(\theta_t),\mu_t)\\
\mu_{t+1} &= \mc{P}_U\left(\mu_t - \eta_\mu\nabla_\mu L(\lambda(\theta_t),\mu_t) \right)
\end{aligned}\right.
,\quad \forall t=0,1,\dots
\end{equation}
where $\eta_{\theta,t} >0$, $\eta_\mu >0$ are step-sizes, and the dual feasible region $U= [0,C_0]$ is introduced in Lemma \ref{lemma:duality}. The method \eqref{eq: algorithm_upate} adopts an alternating update scheme: the primal step performs the gradient ascent in the policy space, whereas the dual step updates the multiplier with projected sub-gradient descent such that $\mu_{t+1}$ is obtained by adding a multiple of the constraint violation to $\mu_t$. The values of $\eta_{\theta,t}$ and $\eta_{\mu}$ will be specified later in the paper. 

In an MDP with the reward $r$, the \textit{Policy gradient} theorem \cite{sutton1999policy} suggests that the gradient $\nabla_\theta V^{\pt}(r)$ can be estimated via the following stochastic estimator (see Lemma \ref{lemma:policy gradient_general}):
\begin{equation}\label{eq:pg_estimator}
\widehat d(\tau,\theta,r) = \sum_{h=0}^{H-1} \nabla \log \pi_\theta (a_h\vert s_h) \left[ \sum_{k=h}^{H-1} \gamma^k r(s_k,a_k)\right],
\end{equation}
where $\tau = \{(s_k,a_k)\}_{k=0}^{H-1}$ is a trajectory of length $H$ under the policy $\pi_\theta$ and initial distribution $\rho$.
In the standard CMDP \eqref{eq:cmdp}, where the value function still admits an additive structure and is linear in the occupancy measure (see \eqref{eq:cmdp-LP}), the Lagrangian can be viewed as a value function with the reward $r-\mu c$, i.e., $L(\lambda(\theta),\mu) = V^{\pi_{\theta}}\left(r-\mu c\right)$.
Hence, it is natural to estimate $\nabla_\theta L(\lambda(\theta),\mu)$ via $\widehat d(\tau,\theta,r-\mu c)$.
However, this favorable result is not applicable when $f(\cdot)$ and $g(\cdot)$ are general concave functions, making the gradient evaluation for the Lagrangian in the concave CMDP far more challenging.
To address this challenge, one approach is to utilize the variational policy gradient technique proposed by \cite{zhang2020variational}, which leverages Fenchel duality to compute the gradient by solving a stochastic saddle point problem.
Another more straightforward method, the REINFORCE-based Policy Gradient introduced by \cite{zhang2021convergence}, reveals that $\nabla_\theta L(\lambda(\theta),\mu)$ can be expressed as the policy gradient of a standard value function with the reward $\nabla_\lambda L(\lambda(\theta),\mu)$, i.e.,
\begin{equation}\label{eq:grad_policy_grad_view}
\begin{aligned}
\nabla_\theta L(\lambda(\theta),\mu) &=\left[ \nabla_{\theta}\lambda(\theta)\right]^\top \nabla_\lambda L(\lambda(\theta),\mu) =\nabla_\theta V^{\pi_\theta}(\nabla_\lambda L(\lambda(\theta),\mu)),
\end{aligned}
\end{equation}
where the first equality follows from the chain rule, and the second equality uses the fact that $\nabla_\theta V^\pt(r) = \left[ \nabla_{\theta}\lambda(\theta)\right]^\top r$. 
To distinguish the policy-dependent reward in \eqref{eq:grad_policy_grad_view} from the standard reward function, we introduce the concept of \textit{shadow reward}, defined as $\nabla_\lambda L(\lambda(\theta),\mu) = \nabla_\lambda f(\lambda(\theta),\mu) + \mu \nabla_\lambda g(\lambda(\theta),\mu)$.
Equation \eqref{eq:grad_policy_grad_view} suggests the following scheme for computing $\nabla_\theta L(\lambda(\theta),\mu)$. 
Firstly, we collect a trajectory $\tau = \{(s_k,a_k)\}_{k=0}^{H-1}$ under $\pt$ and approximate the occupancy measure $\lambda(\theta)$ through the Monte Carlo estimator:
\begin{equation}\label{eq: vanilla_OM_estimator}
\widehat{\lambda}(\tau) = \sum_{k=0}^{H-1} \gamma^k\cdot \mathbbm{1}(s_k,a_k),
\end{equation}
where $\mathbbm{1}(s,a)\in \mbb{R}^{|\mc{S}|\times |\mc{A}|}$ is the indicator vector that takes one in the $(s,a)$-th entry and zero elsewhere.
It is clear that $\mbb{E}_{\tau \sim \pt,\rho}[\widehat{\lambda}(\tau)] \hspace{-0.5mm}=\hspace{-0.5mm} \sum_{k=0}^{H-1}\gamma^{k} \mathbb{P}\left(s_{k}\hspace{-0.5mm}=\hspace{-0.5mm} s, a_k = a\vert \pt, s_0 \sim \rho \right)$ $=: \lambda_H(\theta)$, i.e., $\widehat{\lambda}(\tau)$ is an unbiased estimator for the truncated occupancy measure $\lambda_H(\theta)$.
Secondly, we estimate the true shadow reward $\nabla_\lambda L(\lambda(\theta),\mu)$ by $\nabla_\lambda L(\widehat{\lambda}(\tau),\mu)$.
Finally, we substitute the shadow reward into the stochastic estimator \eqref{eq:pg_estimator}, and approximate the policy gradient $\nabla_\theta L(\lambda(\theta),\mu)$ by $\widehat d(\tau,\theta, \nabla_\lambda L(\widehat{\lambda}(\tau),\mu))$.
Nevertheless, without additional techniques, the convergence for this vanilla REINFORCE-based approach can be slow, especially with large state and action spaces \cite{Tsybakov2008,zhang2020variational}.

\subsection{Variance-Reduced Primal-Dual Policy Gradient Algorithm}
In this section, we present the sample-based version of algorithm \eqref{eq: algorithm_upate}, where we incorporate variance reduction in the gradient evaluation of the Lagrangian function.
Specifically, we adopt a recursive variance reduction scheme similar to the STORM algorithm in stochastic optimization \cite{cutkosky2019momentum, barakat2023reinforcement}, which only requires a single policy rollout in each iteration.
Compared to the vanilla Monte Carlo estimators in \eqref{eq:pg_estimator} and \eqref{eq: vanilla_OM_estimator}, the variance-reduced estimators can achieve a more accurate and sample-efficient approximation.

We begin by introducing the importance sampling (IS), which plays a key role in variance-reduced policy gradient methods \cite{yuan2020stochastic,zhang2021convergence,ding2022global,barakat2023reinforcement}.
Given any $\theta\in \mbb{R}^K$, let $\tau = \{(s_k,a_k)\}_{k=0}^{H-1}$ be a trajectory generated under $\pt$.
Then, for any target policy $\theta^\prime$, the IS weight is defined as
\begin{equation}\label{eq:IS_def}
w(\tau \vert \theta^\prime,\theta) := \prod_{k=0}^{H-1} \dfrac{\pi_{\theta^\prime}(a_k\vert s_k)}{\pi_{\theta}(a_k\vert s_k)}.
\end{equation}
One major effect of $w(\tau \vert \theta^\prime,\theta)$ is to mitigate the distribution shift brought by the policy update \cite{cutkosky2019momentum,zhang2021convergence,barakat2023reinforcement}.
For example, for any reward $r$, it holds that
\begin{equation}\label{eq: effect_of_ISweight}
\mbb{E}_{\tau\sim \pt,\rho}\brac{w(\tau\vert \theta^\prime, \theta) \widehat{\lambda}(\tau)} = \lambda_H(\theta^\prime),\quad \mbb{E}_{\tau \sim \pt,\rho}\brac{w(\tau\vert \theta^\prime, \theta)\widehat d(\tau,\theta^\prime,r)} = \left[ \nabla_{\theta}\lambda_H(\theta^\prime)\right]^\top r,
\end{equation}
where we recall that $\lambda_H(\theta) = \sum_{k=0}^{H-1}\gamma^{k} \mathbb{P}\left(s_{k}=s, a_k = a\vert \pt, s_0 \sim \rho \right)$ is the truncated occupancy measure.
Consequently, even when $\tau$ is generated under policy $\pi_{\theta_t}$ in period $t$, we can still construct an unbiased estimator for $\lambda_H(\theta_{t-1})$ by scaling $\widehat{\lambda}(\tau)$ with the IS weight $w(\tau\vert \theta_{t-1},\theta_t)$.
This idea forms the basis for using momentum-based variance reduction methods, which typically require simulating stochastic estimators for preceding variables using the data from the current period.
The same approach also applies to the estimation of the policy gradient.
In Appendix \ref{app: ISweight}, we provide a proof for \eqref{eq: effect_of_ISweight}.

\renewcommand{\thealgorithm}{\arabic{algorithm}}
\setcounter{algorithm}{0}
\begin{algorithm}[tb]
  \caption{Variance-Reduced Primal-Dual Policy Gradient Algorithm (VR-PDPG)\label{alg:pdpg}}
\begin{algorithmic}[1]
  \STATE {\bfseries Input:} Iteration number $T$; initial policy $\theta_0$; initial dual variable $\mu_0$; step-sizes $\eta_\theta$, $\eta_\mu$, and $\{\alpha_t\}_{t\geq 1}$; trajectory length $H$; dual feasible region $U = [0,C_0]$.
  \STATE Sample trajectory $\tau_0$ with length $H$ under policy $\pi_{\theta_0}$.
  \STATE Compute occupancy measure: $\lambda_0 = \widehat{\lambda}(\tau_0)$. 
  \STATE Compute shadow rewards $r_{f,0} = \nabla_\lambda f(\lambda_0)$ and $r_{g,0} = \nabla_\lambda g(\lambda_0)$. Let $r_{f,-1} = r_{f,0}$ and $r_{g,-1} = r_{g,0}$.
  \STATE Compute gradients: $d_{\diamond,0} = \widehat d(\tau_0, \theta_0, r_{\diamond, 0})$, for $\diamond \in \{f,g\}$.
  \STATE Update policy: $\theta_{1} = \theta_0 + \eta_{\theta,0} d_{L,0}$, where $d_{L,0} = d_{f,0}+\mu_0d_{g,0}$ and $\eta_{\theta,0} = \eta_\theta/\norm{d_{L,0}}$.
  \STATE Update dual variable: $\mu_1 = \mc{P}_U\big(\mu_0-\eta_\mu \cdot g(\lambda_0)\big)$.
  \FOR{iteration $t=1,2,\dots,T-1$}
  \STATE Sample trajectory $\tau_t$ with length $H$ under policy $\pi_{\theta_t}$.
  \STATE Compute occupancy measure: \begin{equation}\label{eq:update_lambda}
\lambda_t = \widehat{\lambda}(\tau_t) + (1-\alpha_t)\big[\lambda_{t-1} - w(\tau_t\vert \theta_{t-1},\theta_t)\widehat{\lambda}(\tau_t)\big].
\end{equation}
\vspace{-1em}
  \STATE Compute shadow rewards $r_{f,t} = \nabla_\lambda f(\lambda_t)$ and $r_{g,t} = \nabla_\lambda g(\lambda_t)$.
 \STATE Compute gradients: 
 \begin{equation}
 d_\dt = \widehat d(\tau_t,\theta_t,r_\dtmone) + (1-\alpha_t)\big[d_\dtmone - w(\tau_t\vert \theta_{t-1},\theta_t)\widehat d(\tau_t,\theta_{t-1},r_{\diamond,t-2}) \big],    
 \end{equation}
 for $\diamond \in \{f,g\}$.
  \STATE Update policy: $\theta_{t+1} = \theta_t + \eta_{\theta,t} d_{L,t}$, where $d_{L,t} = d_{f,t}+\mu_td_{g,t}$ and $\eta_{\theta,t} = \eta_\theta/\norm{d_{L,t}}$.
  \STATE Update dual variable: $\mu_{t+1} = \mc{P}_U\big(\mu_t-\eta_\mu \cdot g(\lambda_t)\big)$.
  \ENDFOR
\end{algorithmic}
\end{algorithm}

Now, we present the Variance-Reduced Primal-Dual Policy Gradient Algorithm (VR-PDPG), as outlined in Algorithm \ref{alg:pdpg}.
During each iteration, VR-PDPG consists of four stages.

\vspace{5pt}\noindent \textbf{Stage 1} (lines 9-11): We collect a single trajectory $\tau_t$ and construct a vanilla occupancy measure estimator through \eqref{eq: vanilla_OM_estimator}, which is then used to refine the variance-reduced estimator $\lambda_t$ in \eqref{eq:update_lambda}.
Specifically, $\lambda_t$ is updated in a recursive way akin to the STORM algorithm \cite{cutkosky2019momentum, barakat2023reinforcement}.
Next, we substitute $\lambda_t$ into $\nabla_\lambda f(\cdot)$ and $\nabla_\lambda g(\cdot)$ to obtain empirical shadow rewards.

\vspace{5pt}\noindent \textbf{Stage 2} (lines 12): We use a similar momentum-based scheme to estimate the policy gradients $\nabla_\theta f(\lambda(\theta_t))$ and $\nabla_\theta g(\lambda(\theta_t))$. 
Note that the shadow rewards from the previous two periods, i.e., $r_\dtmone$ and $r_{\diamond,t-2}$, are intentionally used to avoid the independence issues in the analysis.

\vspace{5pt}\noindent \textbf{Stage 3} (lines 13): We update the policy parameter through a gradient ascent.
Motivated by \cite{barakat2023reinforcement}, we use an adaptively normalized step-size such that $\eta_{\theta, t} = \eta_\theta/\norm{d_{L,t}}$ for each $t\geq 0$, where $d_{L,t}$ is the approximate gradient direction obtained in stage 2.
This normalization ensures the boundedness of the IS weight (see Appendix \ref{app: ISweight} for further discussions).

\vspace{5pt}\noindent \textbf{Stage 4} (lines 14): Finally, we adjust the dual variable by a projected sub-gradient descent, where the dual gradient $\nabla_\mu L(\lambda(\theta_t),\mu_t)$ is approximated by $g(\lambda_t)$.
Recall that Lemma \ref{lemma:duality} ensures that we can limit the dual variable search within the closed interval $U = [0,C_0]$.

\subsection{Exploiting the Hidden Concavity}\label{subsec:hidden convexity}
By itself, problem \eqref{eq:saddle} is a nonconcave-linear maximin problem. 
The existing results \cite{lin2020gradient} for the analysis of the gradient ascent descent method for such problems can only guarantee the convergence to a stationary point in the rate of $\mc{O}(T^{-1/6})$.
Thus, new analyses are required to obtain an improved convergence rate and achieve global optimality.
However, we comment that those analyses based on the performance difference lemma do not apply to concave CMDPs \cite{agarwal2021theory,ding2020natural}.
A key insight is that, due to the loss of linearity in $\lambda$, the performance difference lemma (see Lemma \ref{lemma:performance difference}) can only provide an upper bound for the single-step improvement with the gradient information at the current step, i.e.,
\begin{equation}
L\left(\lambda(\theta_{t+1}), \mu_t\right)-L\left(\lambda(\theta_t), \mu_t\right) 
\leq \dfrac{1}{1-\gamma}\mbb{E}_{s\sim d^{\pi_{\theta_{t+1}}}}\mbb{E}_{a\sim \pi_{\theta_{t+1}}}\brac{A^\ptt\prth{\nabla_\lambda L\left(\lambda\left(\theta_t\right), \mu_t\right);s,a}},
\end{equation}
where $d^\pi(\cdot)$ is the state occupancy measure and $A^\pi(r;s,a):= Q^\pi(r;s,a) - \mbb{E}_{a\sim \pi(\cdot\vert s)}\brac{Q^\pi(r;s,a)}$ is the advantage function with the reward $r$.
Below, we elaborate on how the ``hidden concavity'' of \eqref{eq:problem_theta} with respect to $\lambda$ can help us relate the policy update to the sub-optimality gap.

Assume, for illustration purposes only, that the inverse mapping $\lambda^{-1}:\Lambda \rightarrow \mbb{R}^K$ is globally well-defined such that $\lambda^{-1}(\lambda(\theta)) = \theta$ for all $\theta\in \mbb{R}^K$.
Then, for every period $t$, we can associate $\theta_t$ with an occupancy measure construncted as $(1-\varepsilon)\lambda(\theta_t) + \varepsilon \lambda(\theta^\star) \in \Lambda$, which admits the inverse $\theta_{\varepsilon,t}:= \lambda^{-1}\big((1-\varepsilon)\lambda(\theta_t) + \varepsilon \lambda(\theta^\star)\big)\in \mbb{R}^K$.
Since $L(\cdot,\mu)$ is concave, it follows that
\begin{equation}\label{eq: use_hidden_concavity}
\begin{aligned}
&L(\lambda(\theta_{\varepsilon,t}),\mu_t) = L\big((1-\varepsilon)\lambda(\theta_t)+\varepsilon\lambda(\theta^\star),\mu_t\big) \geq (1-\varepsilon)L(\lambda(\theta_t),\mu_t) + \varepsilon L(\lambda(\theta^\star),\mu_t)\\[5pt]
&\Longleftrightarrow \quad L(\lambda(\theta^\star),\mu_t) - L(\lambda(\theta_{\varepsilon,t}),\mu_t) \leq (1-\varepsilon)\big[L(\lambda(\theta^\star),\mu_t) - L(\lambda(\theta_t),\mu_t)\big].
\end{aligned}
\end{equation}
Hence, if we can demonstrate that $L(\lambda(\theta_{t+1}),\mu_t) \approx L(\lambda(\theta_{\varepsilon,t}),\mu_t)$, inequality \eqref{eq: use_hidden_concavity} implies a nearly $(1-\varepsilon)$-contraction of the Lagrangian from $(\theta_t,\mu_t)$ to $(\theta_{t+1},\mu_t)$.
When the dual step-size is properly chosen, we can further obtain a recursive relation between the sub-optimality gaps $L(\lambda(\theta^\star),\mu_{t+1}) - L(\lambda(\theta_{t+1}),\mu_{t+1})$ and $L(\lambda(\theta^\star),\mu_t) - L(\lambda(\theta_t),\mu_t)$.
This relation paves the way for controlling the average performance in terms of Lagrangian, i.e., $\frac{1}{T}\sum_{t=0}^{T-1}\big[L(\lambda(\theta^\star),\mu_{t})-L(\lambda(\theta_{t}), \mu_{t})\big]$, which is the key quantity in our convergence analysis.

The argument presented above offers an initial insight for the convergence analysis.
However, it is crucial to comment that besides direct parameterization, the global inverse $\lambda^{-1}(\cdot)$ typically does not exist.
This limitation is particularly evident for soft-max type parameterizations such as \eqref{eq:policy_parameterization}, where the parameter space is unconstrained (see Appendix \ref{app:3.3} for further discussions).
Therefore, motivated by \cite{zhang2021convergence}, we employ a more flexible condition that only requires the local invertibility of the occupancy measure mapping, which will be formalized in Assumption \ref{assump:parameterization}.
Finally, it is worth noting that although the hidden concavity/convexity has been studied in unconstrained (RL) problems \cite{zhang2020variational,zhang2021convergence,barakat2023reinforcement,fatkhullin2023stochastic}, the analysis diverges remarkably in the presence of safety constraints.

%% file: file/4_convergence.tex
\section{Convergence Analysis}\label{sec:convergence}
By exploiting the hidden concavity of \eqref{eq:problem_theta} with respect to $\lambda$, we establish the global convergence for both the PDPG and VR-PDPG algorithms in the time-average sense. Particularly, we are interested in quantifying the average performance of the optimality gap, i.e., $\frac{1}{T} \sum_{t=0}^{T-1} \big[f(\lambda(\theta^\star)) -f(\lambda(\theta_t))\big]$, and constraint violation, i.e., $\frac{1}{T}\left[\sum_{t=0}^{T-1}-g(\lambda(\theta_t))\right]_+$, where we recall that $[x]_+:=\max\{x,0\}$.
By definition, when the average performance bound holds, an iteration selected uniformly at random from the sequence $\left\{\left(\theta_t, \mu_t\right)\right\}_{t=0}^{T-1}$ would satisfy the same performance bound in expectation.

First, we formally state our assumption about the general soft-max parameterization (\ref{eq:policy_parameterization}). To avoid introducing additional biases, it is natural to assume that the parameterization has enough expressibility to represent any policy, i.e., $\forall \pi\in \Pi$, $\exists \theta\in \mbb R^K$ such that $\pi=\pi_\theta$.
However, assuming a one-to-one correspondence between $\pi$ and $\theta$ is too restrictive, and it is not satisfied even by the standard soft-max policy.
In practice, using a deep neural network to represent the policy can often arrive at an over-parameterization.
Therefore, following \cite{zhang2021convergence}, we assume that $\pi_\theta$ is defined such that it can represent any policy and that $\lambda(\cdot): \mbb{R}^K \rightarrow \Lambda$ is locally continuously invertible.
As briefly introduced in Section \ref{subsec:hidden convexity}, this (local) invertibility is important in the exploitation of the hidden concavity.
A more detailed discussion can be found in Appendix \ref{app:3.3}.

\begin{assumption}[Parameterization]\label{assump:parameterization}
\hspace{-2mm}Under general soft-max parameterization \eqref{eq:policy_parameterization}, the function $\psi(\cdot;s,a)$ is twice differentiable for all $(s,a)\in \mc{S}\times \mc{A}$ and there exist $M_{\pi,1}, M_{\pi,2} >0$ such that
\begin{equation}\label{eq:assum_para}
\max _{(s,a) \in \mathcal{S}\times \mathcal{A}} \sup _{\theta\in \mbb{R}^K}\left\|\nabla_{\theta} \psi( \theta;s, a )\right\| \leq M_{\pi,1},\quad  \max _{(s,a) \in \mathcal{S}\times \mathcal{A}} \sup _{\theta\in \mbb{R}^K}\left\|\nabla_{\theta}^{2} \psi( \theta;s, a)\right\| \leq M_{\pi,2}.
\end{equation}
Furthermore, $\pi_\theta$ over-parameterizes the set of all stochastic policies in the following senses:
\begin{enumerate}
    \item[\textbf{\textit{(I)}}] For every $\theta\in \mbb{R}^K$, there exists a neighborhood $\mc{U}_\theta \ni \theta$ such that the restriction of $\lambda(\cdot)$ to $\mc{U}_\theta$ is a bijection between $\mc{U}_\theta$ and $\mc{V}_{\lambda(\theta)}:=\lambda(\mc{U}_\theta)$, where we
    denote $\lambda^{-1}_{\mc{V}_{\lambda(\theta)}}: \mc{V}_{\lambda(\theta)}\rightarrow \mc{U}_\theta$ as the local inverse of $\lambda(\cdot)$, i.e., $\lambda^{-1}_{\mc{V}_{\lambda(\theta)}}(\lambda(\theta_0))=\theta_0,\ \forall \theta_0\in \mc{U}_\theta$.
    \item[\textbf{\textit{(II)}}] 
    There exists a universal constant $L_\lambda$ such that $\lambda^{-1}_{\mc{V}_{\lambda(\theta)}}(\cdot)$ is $L_\lambda$-Lipschitz continuous for all $\theta\in \mbb{R}^K$;
    \item[\textbf{\textit{(III)}}] There exists $\bar\varepsilon >0$ such that $(1-\varepsilon)\lambda(\theta)+\varepsilon\lambda(\theta^\star)\in \mc{V}_{\lambda(\theta)},\ \forall \varepsilon\leq \bar\varepsilon,\ \forall \theta\in \mbb{R}^K$.
\end{enumerate}
\end{assumption}

Under \eqref{eq:assum_para} of Assumption \ref{assump:parameterization}, the score function $\nabla_\theta\log \pi_\theta (a\vert s)$ is guaranteed to be upper-bounded (see Lemma \ref{lemma:parameterization}), thereby ensuring the boundedness of the policy gradient estimator in \eqref{eq:pg_estimator}.
Next, we make the assumptions regarding the Lipschitz continuity and smoothness of the objective and constraint functions.

\begin{assumption}[Utility function]\label{assump:lipschit_gradient}
There exist $M_\lambda, \ell_\lambda >0$ such that
\begin{enumerate}
    \item[\textbf{\textit{(I)}}] $f(\cdot)$ and $g(\cdot)$ are $M_\lambda$-Lipschitz continuous, i.e., $\norm{\nabla_\lambda f(\lambda)} \leq M_\lambda$, $\norm{\nabla_\lambda g(\lambda)} \leq M_\lambda$ for all $\lambda$.
    \item[\textbf{\textit{(II)}}] $f(\cdot)$ and $g(\cdot)$ are $\ell_\lambda$-smooth in the sense that for all $\lambda_1, \lambda_2$, it holds that
    \begin{equation*}
    \norm{\nabla_\lambda f(\lambda_1) - \nabla_\lambda f(\lambda_2)}_\infty \leq \ell_\lambda \norm{\lambda_1 - \lambda_2},\ \norm{\nabla_\lambda g(\lambda_1) - \nabla_\lambda g(\lambda_2)}_\infty \leq \ell_\lambda \norm{\lambda_1 - \lambda_2}.
    \end{equation*}
\end{enumerate}
\end{assumption}

We note that Assumptions \ref{assump:parameterization} and \ref{assump:lipschit_gradient} are standard in the literature of RL with general utility \cite{zhang2021convergence, zhang2022multi, barakat2023reinforcement, ying2024scalable}. The following function properties are derived from Assumptions \ref{assump:parameterization} and \ref{assump:lipschit_gradient}. 

\begin{lemma}\label{lemma:boundedness_smoothness}
Let Assumptions \ref{assump:parameterization} and \ref{assump:lipschit_gradient} hold. There exist $M,\ell_\theta,L_{\theta,\lambda} >0$ such that
\begin{enumerate}
    \item[\textbf{\textit{(I)}}]  $f(\cdot)$ and $g(\cdot)$ are bounded by $M$ on $\Lambda$, i.e., $|f(\lambda)|\leq M$, $|g(\lambda)|\leq M$, for all $\lambda\in \Lambda$.
    \item[\textbf{\textit{(II)}}] $f(\lambda(\theta))$ and $g(\lambda(\theta))$ are $\ell_\theta$-smooth w.r.t. $\theta$, i.e., for all $\theta_1, \theta_2 \in \mbb{R}^K$, it holds that
    \begin{equation*}
    \norm{\nabla_\theta f(\lambda(\theta_1)) \hspace{-2pt}-\hspace{-2pt} \nabla_\theta f(\lambda(\theta_2))} \leq \ell_\theta \norm{\theta_1 - \theta_2},\ \norm{\nabla_\theta g(\lambda(\theta_1)) \hspace{-2pt}-\hspace{-2pt} \nabla_\theta g(\lambda(\theta_2))} \leq \ell_\theta \norm{\theta_1 - \theta_2}.
    \end{equation*}
    \item[\textbf{\textit{(III)}}] For every $H > 0$ and $\theta \in \mbb{R}^K$, it holds that
    \begin{equation*}
    \norm{\nabla_\theta f(\lambda(\theta)) - \nabla_\theta f(\lambda_H(\theta))} \leq L_{\theta,\lambda}\gamma^H,\ \norm{\nabla_\theta g(\lambda(\theta)) - \nabla_\theta g(\lambda_H(\theta))} \leq L_{\theta,\lambda}\gamma^H,
    \end{equation*}
    where $\lambda_H(\theta) = \sum_{k=0}^{H-1}\gamma^{k} \mathbb{P}\left(s_{k}=s, a_k = a\vert \pt, s_0 \sim \rho \right)$.
    \item[\textbf{\textit{(IV)}}] For any $\mu \in U = [0,C_0]$, the Lagrangian function $L(\cdot,\mu)$ is bounded by $(1+C_0)M$ on $\Lambda$; $L(\lambda(\theta),\mu)$ is $(1+C_0)\ell_\theta$-smooth w.r.t. $\theta$; for every $H > 0$ and $\theta \in \mbb{R}^K$, it holds that
\begin{equation}\label{eq: L_smooth}
\norm{\nabla_\theta L(\lambda(\theta),\mu) - \nabla_\theta L(\lambda_H(\theta),\mu)} \leq (1+C_0)L_{\theta,\lambda}\gamma^H.
\end{equation}
\end{enumerate}
\end{lemma}

It is worth mentioning that the property \textbf{\textit{(III)}} is similar to the Lipschitz continuity of $\nabla_\theta f(\lambda)$ with respect to $\lambda$, where we use the fact that $\norm{\lambda(\theta)-\lambda_H(\theta)}$ is bounded by $\gamma^H$. 
This boundedness plays a crucial role in the convergence proof for the sample-based scenario since $\lambda_t$ in Algorithm \ref{alg:pdpg} is indeed an unbiased estimator for the truncated occupancy measure $\lambda_H(\theta)$, instead of $\lambda(\theta)$ directly.
In addition, we note that the properties in \textbf{\textit{(IV)}} are a direct consequence of \textbf{\textit{(I)}} to \textbf{\textit{(III)}} since the Lagrangian is the simple sum of $f(\lambda(\theta))$ and $\mu g(\lambda(\theta))$.

The rest of this section is divided into two cases: the exact setting (PDPG in \eqref{eq: algorithm_upate}) and the sample-based setting (VR-PDPG in Algorithm \ref{alg:pdpg}). We unfold the convergence results by first presenting a key finding regarding the average performance of the Lagrangian, which lays the foundation for deriving both the optimality gap and constraint violation. 

\subsection{Exact setting}\label{subse:exact}
This section is dedicated to the convergence analysis of the PDPG algorithm in \eqref{eq: algorithm_upate}. 
We first characterize the average performance bounds of the Lagrangian for both general concave and $\sigma$-strongly concave frameworks in the following lemma. 

\begin{lemma}\label{lemma: avg_L}
Let Assumptions \ref{assump:parameterization} and \ref{assump:lipschit_gradient} hold.
Then, for every $T> 0$ and $\varepsilon \leq \bar\varepsilon$, the sequence $\left\{\left(\theta_t, \mu_t\right)\right\}_{t=0}^{T-1}$ generated by PDPG in \eqref{eq: algorithm_upate} with $\eta_{\theta,t} \equiv  \frac{1}{(1+C_0) \ell_\theta}$ satisfies that
\begin{equation}\label{eq:prop_convergence}
\dfrac{1}{T}\sum_{t=0}^{T-1}\big[L(\lambda(\theta^\star),\mu_{t})-L(\lambda(\theta_{t}), \mu_{t})\big]\leq \dfrac{L(\lambda(\theta^\star),\mu_{0})-L(\lambda(\theta_{0}), \mu_{0})}{\varepsilon T} + \dfrac{2\varepsilon (1+C_0) \ell_\theta L_\lambda^2}{(1-\gamma)^2} + \dfrac{2 \eta_\mu M^2}{\varepsilon}.
\end{equation}
If $f(\cdot)$ is $\sigma$-strongly concave with respect to $\lambda$, the sequence $\left\{\left(\theta_t, \mu_t\right)\right\}_{t=0}^{T-1}$ further satisfies that
\begin{equation}\label{eq:prop_convergence_sc}
\dfrac{1}{T}\sum_{t=0}^{T-1}\big[L(\lambda(\theta^\star),\mu_{t})-L(\lambda(\theta_{t}), \mu_{t})\big]\leq \dfrac{L(\lambda(\theta_T),\mu_{T-1})-L(\lambda(\theta_{0}), \mu_{0})}{\tilde\varepsilon T} + \dfrac{\eta_\mu M^2}{\tilde \varepsilon},
\end{equation}
where $\tilde{\varepsilon} = \min\left\{\bar\varepsilon, \frac{\sigma}{\sigma+2(1+C_0) \ell_\theta L_\lambda^2 }\right\}$.
\end{lemma}

The core idea in proving Lemma \ref{lemma: avg_L} is that one can relate the primal update to the sub-optimality gap $L(\lambda(\theta^\star),\mu_{t})- L(\lambda(\theta_{t}), \mu_{t})$ by leveraging the hidden concavity of (\ref{eq:problem_theta}) with respect to $\lambda$.
Then, as $|\mu_{t+1}-\mu_t| =\mc{O}(\eta_\mu)$, we are able to draw a recursion between the sub-optimality gaps for two consecutive periods (see (\ref{eq:recursion_convex})).
The average performance in terms of the Lagrangian can be decomposed into the summation of the average optimality gap and the weighted average of the constraint violation, i.e.,
\begin{equation*}
\dfrac{1}{T}\sum_{t=0}^{T-1} \hspace{-2pt} \big[L(\lambda(\theta^\star),\mu_{t})-L(\lambda(\theta_{t}), \mu_{t})\big] \hspace{-2pt}=\hspace{-2pt} \dfrac{1}{T}\sum_{t=0}^{T-1} \big[f(\lambda(\theta^\star)) - f(\lambda(\theta_t))\big] + \dfrac{1}{T}\sum_{t=0}^{T-1}\mu_t\big[g(\lambda(\theta^\star))-g(\lambda(\theta_t))\big].
\end{equation*}
Since $\theta^\star$ must be a feasible solution, the difference $g(\lambda(\theta^\star))-g(\lambda(\theta_t))$ can be interpreted as an approximate of the constraint violation.
Next, in Lemma \ref{lemma:obtain_opt_const_exact}, we decouple the Lagrangian bound to obtain the optimality gap and the true constraint violation.
\begin{lemma}\label{lemma:obtain_opt_const_exact}
Let Assumptions \ref{assump:slater} and \ref{assump:lipschit_gradient} hold, and the sequence $\left\{\left(\theta_t, \mu_t\right)\right\}_{t=0}^{T-1}$ generated by PDPG in \eqref{eq: algorithm_upate} satisfy that 
$$\frac{1}{T}\sum_{t= 0}^{T-1}\big[ L(\lambda(\theta^\star),\mu_t) - L(\lambda(\theta_t),\mu_t)\big] \leq C.$$
Then, the optimality gap and constraint violation are upper-bounded as follows
\begin{subequations}
\begin{align}
&\dfrac{1}{T} \sum_{t=0}^{T-1} \big[f(\lambda(\theta^\star)) - f(\lambda(\theta_t))\big] \leq C+\dfrac{\mu_0^2-\mu_T^2}{2T\eta_\mu} + \dfrac{\eta_\mu M^2}{2},\label{eq:optimality_gap_lemma_exact}\\
&\dfrac{1}{T}\left[\sum_{t=0}^{T-1}-g(\lambda(\theta_t))\right]_+\leq C+\dfrac{\max_{\mu\in U}\left\{(\mu_0-\mu)^2 - (\mu_T-\mu)^2\right\}}{2T\eta_\mu} + \dfrac{\eta_\mu M^2}{2} \label{eq:constraint_violation_lemma_exact}.
\end{align}
\end{subequations}
\end{lemma}

Recall that the dual feasible region $U = [0,C_0]$ is set by taking $C_0 = 1+\big(M_F-f(\lambda(\widetilde \theta))\big)/{\xi}$ in Lemma \ref{lemma:duality}. 
Since $\mu^\star\leq \big[f(\lambda(\theta^\star))-f(\lambda(\widetilde\theta))\big]/\xi\leq C_0-1$, it implies that $\widehat\mu :=  \mu^\star+1 \in U$.
This ``slackness'' plays an important role when bounding the constraint violation, as we can write $\big[\sum_{t=0}^{T-1} -g(\lambda(\theta_t))\big]_+ = \big[(\mu^\star - \widehat{\mu})\sum_{t=0}^{T-1} \nabla_\mu L(\lambda(\theta_t), \mu_t)\big]_+$, where the latter term can be related to the first-order expansion of $L(\lambda(\theta),\cdot)$ and bounded via the use of telescoping sums.

Combining Lemmas \ref{lemma: avg_L} and \ref{lemma:obtain_opt_const_exact}, we optimize the step-size choice and establish the global convergence rate for algorithm \eqref{eq: algorithm_upate} in the following theorem.

\begin{theorem}[Exact setting]\label{thm:exact}
Let Assumptions \ref{assump:slater}, \ref{assump:parameterization}, and \ref{assump:lipschit_gradient} hold.
For every $T\geq ({\bar\varepsilon})^{-3}$, we choose $\mu_0=0$, $\eta_{\theta,t} =  \frac{1}{(1+C_0)\ell_\theta}$, and $\eta_\mu = T^{-2/3}$. 
Then, the sequence $\left\{\left(\theta_t, \mu_t\right)\right\}_{t=0}^{T-1}$ generated by PDPG in \eqref{eq: algorithm_upate} satisfies that
\begin{subequations}\label{eq:rate_exact_generalconcave}
\begin{align}
&\dfrac{1}{T} \sum_{t=0}^{T-1} \big[f(\lambda(\theta^\star)) - f(\lambda(\theta_t))\big] \leq \dfrac{2M+M^2/2}{T^{2/3}} + \dfrac{2(1+C_0)\ell_\theta L_\lambda^2}{(1-\gamma)^2T^{1/3}} + \dfrac{2M^2}{T^{1/3}},\label{eq:optimality_gap}\\
&\dfrac{1}{T}\left[\sum_{t=0}^{T-1}-g(\lambda(\theta_t))\right]_+ \leq \dfrac{2M+M^2/2}{T^{2/3}} + \dfrac{2(1+C_0)\ell_\theta L_\lambda^2}{(1-\gamma)^2T^{1/3}}+ \dfrac{2M^2+C_0^2/2}{T^{1/3}}.\label{eq:constraint_violation}
\end{align}
\end{subequations}
Suppose further that $f(\cdot)$ is $\sigma$-strongly concave with respect to $\lambda$. For every $T > 0$, we choose  $\mu_0=0$, $\eta_{\theta,t} \equiv \frac{1}{(1+C_0)\ell_\theta}$, and $\eta_\mu = T^{-1/2}$. Then, the sequence $\left\{\left(\theta_t, \mu_t\right)\right\}_{t=0}^{T-1}$ satisfies that
\begin{subequations}
\begin{align}
&\dfrac{1}{T} \sum_{t=0}^{T-1} \big[f(\lambda(\theta^\star)) - f(\lambda(\theta_t))\big]  \leq \dfrac{(2+C_0)M}{\tilde\varepsilon T} +  \left(\dfrac{M^2}{\tilde\varepsilon}+ \dfrac{M^2}{2}\right)\dfrac{1}{\sqrt{T}},\label{eq:optimality_gap_sc}
\\
&\dfrac{1}{T}\left[\sum_{t=0}^{T-1}-g(\lambda(\theta_t))\right]_+ \leq \dfrac{(2+C_0)M}{\tilde\varepsilon T} +  \left(\dfrac{M^2}{\tilde\varepsilon}+ \dfrac{M^2+C_0^2}{2}\right)\dfrac{1}{\sqrt{T}},\label{eq:constraint_violation_sc}
\end{align}
\end{subequations}
where $\tilde{\varepsilon} = \min\left\{\bar\varepsilon, \frac{\sigma}{\sigma+2(1+C_0)\ell_\theta L_\lambda^2 }\right\}$.
\end{theorem}
\vspace{5pt}

In the case of general concave CMDPs, Theorem \ref{thm:exact} shows that PDPG \eqref{eq: algorithm_upate} achieves a global convergence in the time-average sense such that the optimality gap and constraint violation decay to zero with the rate of $\mc{O}(T^{-1/3})$. 
In other words, to obtain an $\mc{O}(\epsilon)$-accurate solution, the iteration complexity is $\mc{O}(\epsilon^{-3})$.
When $f(\cdot)$ and $g(\cdot)$ are linear functions as in standard CMDPs, i.e., problem \eqref{eq:cmdp}, Theorem \ref{thm:exact} shows a better convergence rate than the existing $\mc{O}(T^{-1/4})$ rate for the policy gradient primal-dual methods (see \cite{ding2022convergence} or \cite[Theorem 6]{ding2022convergencejournal}).
When the objective function $f(\lambda(\theta))$ is strongly concave with respect to $\lambda$, we can further improve the convergence rate to $\mc{O}\left(T^{-1/2}\right)$ by taking the dual step-size $\eta_\mu = \mc{O}\left(T^{-1/2}\right)$.
This rate improvement mainly results from the better average performance of the Lagrangian in the strongly concave scenario (see \eqref{eq:prop_convergence_sc}), which does not contain an $\mc{O}(\tilde\varepsilon)$ error term.
Hence, unlike the general cacave setting where we need to choose $\epsilon = \mc{O}(T^{-1/3})$ in \eqref{eq:prop_convergence}, we can choose $\varepsilon = \mc{O}(1)$ to obtain a sharper convergence rate in the strongly concave scenario.

\subsection{Sample-based setting}
In this section, we analyze the convergence of VR-PDPG presented in Algorithm \ref{alg:pdpg}, with a focus on the general concave scenario. 
Differently from the exact setting in Section \ref{subse:exact} where the constant step-size is employed, the sample-based algorithm adopts the normalized policy update rule, i.e., $\eta_{\theta,t} = \eta_\theta/\norm{d_{L,t}}$, which inevitably requires the use of sufficiently small primal step-sizes. As a result, the strong concavity cannot provide enough curvature to offset the approximation errors, thereby not improving the rate of convergence over the general concavity in the sample-based setting. 

This section is structured as follows.
We start by providing an average performance bound for 
the Lagrangian function in the sample-based scenario (see Lemma \ref{lemma:avg_L_sample}), which is then decoupled into optimality gap and constraint violation (see Lemma \ref{lemma:obtain_opt_const_sample}). These steps pave the way for the main convergence result in Theorem \ref{thm:sample}.

\begin{lemma}\label{lemma:avg_L_sample}
\hspace{-2.5pt}Let Assumptions \ref{assump:parameterization} and \ref{assump:lipschit_gradient} hold.\hspace{-2pt}
Then, for every $T > 0$ and $\varepsilon \leq \min\Big\{\bar\varepsilon,\frac{(1-\gamma)\eta_\theta}{\sqrt{2}L_\lambda}\Big\}$, the sequence $\left\{\left(\theta_t, \mu_t\right)\right\}_{t=0}^{T-1}$ generated by VR-PDPG in Algorithm \ref{alg:pdpg} satisfies that
{\setlength{\abovedisplayskip}{20pt}
\setlength{\belowdisplayskip}{20pt}
\begin{equation}\label{eq:avg_L_sample}
\begin{aligned}
\dfrac{1}{T}\sum_{t=0}^{T-1}\big[L(\lambda(\theta^\star),\mu_{t})-L(\lambda(\theta_{t}), \mu_{t})\big]
\leq& \dfrac{L(\lambda(\theta^\star),\mu_{0})-L(\lambda(\theta_{0}), \mu_{0})}{\varepsilon T} + \dfrac{2\eta_\theta }{\varepsilon T}\sum_{t=0}^{T-1}\norm{e_{L,t}}\\
& \hspace{-3cm} + \dfrac{\varepsilon(1+C_0) \ell_\theta L_\lambda^2}{(1-\gamma)^2} + \dfrac{1}{\varepsilon}\left[(1+C_0)\left(\dfrac{ \ell_\theta \eta_\theta^2}{2} + 2L_{\theta,\lambda} \eta_\theta \gamma^H\right) + 2\eta_\mu M^2 \right],
\end{aligned}
\end{equation}
}
where $e_{L,t}:= d_{L,t} - \nabla_\theta L(\lambda_H(\theta_t),\mu_t)$.
\end{lemma}

Compared to its counterpart in the exact setting (see Lemma \ref{lemma: avg_L}), Lemma \ref{lemma:avg_L_sample} yields a different bound, which is due to the choice of the normalized update rule and the sample-based estimations.
In particular, the extra term $e_{L,t}$ quantifies the approximation error of the policy gradient estimator $d_{L,t}$ with respect to $\nabla_\theta L(\lambda_H(\theta_t),\mu_t)$, and the term that contains $\gamma^H$ measures the truncation error $\nabla_\theta L(\lambda_H(\theta_t),\mu_t) - \nabla_\theta L(\lambda(\theta_t),\mu_t)$.
Next, we decompose the boundedness on the Lagrangian into the optimality gap and constraint violation.

\begin{lemma}\label{lemma:obtain_opt_const_sample}
Let Assumptions \ref{assump:slater} and \ref{assump:lipschit_gradient} hold, and the sequence $\left\{\left(\theta_t, \mu_t\right)\right\}_{t=0}^{T-1}$ generated by VR-PDPG in Algorithm \ref{alg:pdpg} satisfies that 
$$\frac{1}{T}\sum_{t= 0}^{T-1}\big[ L(\lambda(\theta^\star),\mu_t) - L(\lambda(\theta_t),\mu_t)\big] \leq C.$$
Then, the optimality gap and constraint violation can be respectively upper-bounded as follows
\begin{subequations}
\begin{align}
&\dfrac{1}{T} \sum_{t=0}^{T-1}\big[f(\lambda(\theta^\star)) - f(\lambda(\theta_t))\big] \leq C+\dfrac{\mu_0^2-\mu_T^2}{2T\eta_\mu} + \eta_\mu M^2 + C_0M_\lambda \gamma^H  + 2\eta_\mu M_\lambda^2 \gamma^{2H} \label{eq:optimality_gap_lemma}\\
&\hspace{138pt}+ \dfrac{C_0M_\lambda}{T}\sum_{t=0}^{T-1} \norm{e_{\lambda,t}}  + \dfrac{2\eta_\mu M_\lambda^2}{T}\sum_{t=0}^{T-1} \norm{e_{\lambda,t}}^2,\notag\\
&\dfrac{1}{T}\left[\sum_{t=0}^{T-1}-g(\lambda(\theta_t))\right]_+ \leq C+\dfrac{\max_{\mu\in U}\left\{(\mu_0-\mu)^2 - (\mu_T-\mu)^2\right\}}{2T\eta_\mu}+ \eta_\mu M^2 + C_0M_\lambda \gamma^H  \label{eq:constraint_violation_lemma}\\
&\hspace{110pt} + 2\eta_\mu M_\lambda^2 \gamma^{2H}+ \dfrac{C_0M_\lambda}{T}\sum_{t=0}^{T-1} \norm{e_{\lambda,t}}+ \dfrac{2\eta_\mu M_\lambda^2}{T}\sum_{t=0}^{T-1} \norm{e_{\lambda,t}}^2,\notag
\end{align}
\end{subequations}
where $e_{\lambda,t} := \lambda_H(\theta_t) - \lambda_t$, for all $t\geq 0$.
\end{lemma}

The upper bounds in \eqref{eq:optimality_gap_lemma} and \eqref{eq:constraint_violation_lemma}
contain the approximation error term $e_{\lambda,t}$, which stems from the difference between the estimated occupancy measure $\lambda_t$ and the true truncated occupancy measure $\lambda_H(\theta_t)$. 
We control the magnitude of this error as documented in Appendix \ref{subsec:control_e} and then fine-tune the step-sizes to arrive at our main convergence result.
For the brevity of the presentation, we use the symbol ``$\lesssim$'' to hide numerical constants and higher-order terms. 

\begin{theorem}[Sample-based setting]\label{thm:sample}
\hspace{-1mm}Let Assumptions \ref{assump:slater}, \ref{assump:parameterization}, and \ref{assump:lipschit_gradient} hold. For every $T \geq \frac{1}{2}\Big(\frac{1-\gamma}{L_\lambda \bar\varepsilon}\Big)^2$, we choose $\mu_0 = 0$, $\eta_\theta = T^{-1/2}$, $\eta_\mu = T^{-3/4}$, $\alpha_t = (t+1)^{-1}$, and $H = \frac{1}{2}\log_{1/\gamma}T$.
Then, the sequence $\left\{\left(\theta_t, \mu_t\right)\right\}_{t=0}^{T-1}$ generated by VR-PDPG in Algorithm \ref{alg:pdpg} satisfies that
\begin{subequations}
\begin{align}
\mbb{E}& \left[\dfrac{1}{T} \sum_{t=0}^{T-1} f(\lambda(\theta^\star)) - f(\lambda(\theta_t))\right]  
\lesssim \left[C_0M_\lambda \sqrt{C_\lambda} + \dfrac{C_0L_\lambda C_e  + L_\lambda M^2}{1 - \gamma}\right]\dfrac{1}{T^{1/4}},\label{eq:optimality_gap_sample}
\\
\mbb{E}&\left[\dfrac{1}{T} \sum_{t=0}^{T-1} -g(\lambda(\theta_t))\right]_+ \hspace{-8pt}\lesssim \left[C_0^2+C_0M_\lambda \sqrt{C_\lambda} + \dfrac{C_0L_\lambda C_e + L_\lambda M^2}{1 -\gamma}\right]\dfrac{1}{T^{1/4}},\label{eq:constraint_violation_sample}
\end{align}
\end{subequations}
where $\mbb{E}[x]_+ := \mbb{E}[\max\{x,0\}]$, and the constants $C_\lambda, C_e$ are defined as follows
\begin{subequations}
\begin{align}
&C_\lambda := \dfrac{1+C_w}{(1-\gamma)^2},\label{eq:C_lambda}\\[3pt]
&C_e := \dfrac{\sqrt{M_\lambda^2\left[(1+C_w)M_{\pi,1}^2 + (M_{\pi,1}^2+M_{\pi,2})^2 \right] +(M_{\pi,1} \ell_\lambda)^2 C_\lambda} + M_{\pi,1}\ell_\lambda \sqrt{C_\lambda}}{(1-\gamma)^2}, \label{eq:C_e}\\[5pt]
&C_w := H(W+1)\left[(8H+2)M_{\pi,1}^2 + 2M_{\pi,2}\right].\label{eq:C_w}
\end{align}
\end{subequations}
\end{theorem}

Since each $\theta_t$ is a random variable in the sample-based setting, the convergence established in Theorem \ref{thm:sample} is in the expectation sense.
In each iteration of Algorithm \ref{alg:pdpg}, $ H = \mc{O}(\log T)$ new samples are collected.
Thus, the $\mc{O}(T^{-1/4})$ convergence rate implies that a total of $\widetilde{\mc{O}}(\epsilon^{-4})$ samples are required to achieve an $\mc{O}(\epsilon)$ optimality gap and constraint violation.
In the special case of standard CMDPs, i.e., problem \eqref{eq:cmdp}, Theorem \ref{thm:sample} matches the sample complexity of natural policy gradient-based primal-dual methods by \cite{ding2022convergencejournal,bai2023achieving}.

\begin{remark}[Extension to multiple constraints]\label{rmk: multiple_constraints}
Although we present the single constraint setting only, it is straightforward to extend all of our results, including the zero constraint violation result in Section \ref{sec:zero}, to the general $n$-constraint setting, i.e., 
\begin{equation}
\max_{\theta\in \mbb{R}^K}\ f(\lambda(\theta)) \ \quad \text {s.t.}\ \  g_i(\lambda(\theta)) \geq 0,\quad \forall i= 1,\dots, n.
\end{equation}
Firstly, under the Slater's condition, we can generalize the strong duality in Lemma \ref{lemma:duality} and show that the optimal dual variable $\mu^\star \in \mbb{R}^n$ satisfies that $0\leq \mu^\star_i \leq \big[f(\lambda\left(\theta^{\star}\right))-f(\lambda(\tilde{\theta}))\big] / \xi_i$, where $\xi_i$ is the slackness for the $i$-th constraint.
Hence, we can replace the dual feasible region $U = [0,C_0] = [0,1+(M_F-f(\lambda(\widetilde{\theta}))) / \xi]$ with the hypercube $\big\{\mu\in \mbb{R}^n\vert 0\leq \mu_i\leq (M_F-f(\lambda(\widetilde{\theta})))/\xi_i \big\}$.
Secondly, in Algorithm \ref{alg:pdpg}, we shall create separate variance-reduced estimators for each constraint (similar to $d_{g,t}$ and $g(\lambda_t)$).
Finally, in the convergence proof, we replace the simple square of difference (e.g., $(\mu_{t+1}-\mu)^2$ in \eqref{eq: u-u2}) with the squared norm and replace the scalar multiplication (e.g., $\mu_tg(\lambda_t)$ in \eqref{eq: mu_tg(lambda_t)}) with the inner product.
Then, it is easy to verify that the same convergence results hold for the optimality gap and constraint violation.
Note that the constraint violation in the $n$-constraint setting is defined as $\frac{1}{T}\big\|\big[\sum_{t=0}^{T-1}-g(\lambda(\theta_t))\big]_+\big\|$, where function $g(\cdot)$ represents the vector of $n$ constraints and the operator $[\cdot]_+$ applies entrywisely.
\end{remark}

%% file: file/5_zero_violation.tex
\section{Zero Constraint Violation}\label{sec:zero}
In safety-critical systems where violating the constraint may induce an unexpected cost, having a zero constraint violation is of great importance.
Motivated by the recent works \cite{liu2021policy,liu2021learning}, we will show that a zero constraint violation can be achieved while maintaining the same order of convergence rate for the optimality gap.
We remark that the proof ideas are the same for both the exact setting and the sample-based setting, with the constant of the latter one being more involved.
For the ease of presentation, we use the exacting setting as the showcase.
Consider the pessimistic counterpart of (\ref{eq:problem_theta}):
\begin{equation}\label{eq:prob_pessimistic_new}
\max_{\theta\in \mbb{R}^K} f(\lambda(\theta))  \quad \text {s.t.}\ \ g_\delta(\lambda(\theta)):= g(\lambda(\theta))-\delta \geq 0,
\end{equation}
where $\delta >0$ is the pessimistic term to be determined.
Suppose that $\delta <\xi$, i.e., the pessimistic term is smaller than the strict feasibility of the Slater point, so that the Slater's condition still holds for problem \eqref{eq:prob_pessimistic_new}.
Consider applying algorithm \eqref{eq: algorithm_upate} to problem \eqref{eq:prob_pessimistic_new}, i.e.,
\begin{equation}\label{eq:algorithm_update_zero}
\left\{\begin{aligned}
\theta_{t+1} &= \theta_t +\eta_{\theta,t}\nabla_\theta L_\delta(\lambda(\theta_t),\mu_t)\\
\mu_{t+1} &= \mc{P}_U\left(\mu_t - \eta_\mu\nabla_\mu L_\delta(\lambda(\theta_t),\mu_t) \right)
\end{aligned}\right.
,\ t=0,1,\dots,
\end{equation}
where $L_\delta(\lambda(\theta),\mu):= f(\lambda(\theta))+\mu g_\delta(\lambda(\theta))$ is the Lagrangian function for problem \eqref{eq:prob_pessimistic_new}.
The following theorem states that, with a carefully chosen $\delta$, the constraint violation of the iterates generated by algorithm \eqref{eq:algorithm_update_zero} will be zero for the original problem (\ref{eq:problem_theta}) when $T$ is reasonably large, and the optimality gap still remains the same order as before.
We direct the reader to Appendix \ref{app:aux} for the proof of Theorem \ref{thm:zero}.

\begin{theorem}\label{thm:zero}
Let Assumptions \ref{assump:slater}, \ref{assump:parameterization}, and \ref{assump:lipschit_gradient} hold.
For fixed $T>0$, let $\delta=\mc{O}(T^{-1/3})$ be the solution to the equation
\begin{equation*}
\dfrac{2M+(M+\xi)^2/2}{T^{2/3}} + \dfrac{2(1+C_{0,\delta})\ell_\theta L_\lambda^2}{(1-\gamma)^2T^{1/3}}+ \dfrac{2 (M+\xi)^2 + C_{0, \delta}^2/2}{T^{1/3}}-\delta = 0,
\end{equation*}
where $C_{0,\delta}=1+\left(M-f(\lambda(\widetilde \theta))\right)/{(\xi-\delta)}$.
For $T>0$ such that $\delta <\xi$, we choose $\mu_0=0$, $\eta_{\theta,t} =\frac{1}{(1+C_0)\ell_\theta}$, and $\eta_\mu = T^{-2/3}$.
Then, the sequence $\left\{\left(\theta_t, \mu_t\right)\right\}_{t=0}^{T-1}$ generated by algorithm \eqref{eq:algorithm_update_zero} with $U = [0, C_{0,\delta}]$ satisfies that
\begin{align}
\dfrac{1}{T} \sum_{t=0}^{T-1} \Big[f(\lambda(\theta^\star)) \hspace{-2pt}-\hspace{-2pt} f(\lambda(\theta_t))\Big] &\leq \dfrac{2\delta M}{\xi} + \dfrac{2M \hspace{-2pt}+\hspace{-2pt} (M \hspace{-2pt}+\hspace{-2pt} \xi)^2/2}{T^{2/3}} + \dfrac{2 (1 \hspace{-2pt}+\hspace{-2pt} C_{0,\delta})\ell_\theta L_\lambda^2}{(1-\gamma)^2T^{1/3}}+\dfrac{2(M \hspace{-2pt}+\hspace{-2pt} \xi)^2}{T^{1/3}},\\
\dfrac{1}{T}\left[\sum_{t=0}^{T-1}-g(\lambda(\theta_t))\right]_+ &= 0.
\end{align}
Suppose further that $f(\cdot)$ is $\sigma$-strongly concave w.r.t. $\lambda$ on $\Lambda$. 
For fixed $T>0$, let $\delta=\mc{O}(T^{-1/2})$ be the solution to the equation
\begin{equation*}
\dfrac{2M+C_{0,\delta}(M+\xi)}{\tilde\varepsilon T} +  \left(\dfrac{(M+\xi)^2}{\tilde\varepsilon}+ \dfrac{(M+\xi)^2+C_{0,\delta}^2}{2}\right)\dfrac{1}{\sqrt{T}} -\delta = 0,
\end{equation*}
For $T>0$ such that $\delta <\xi$, we choose $\mu_0=0$, $\eta_{\theta,t} =\frac{1}{(1+C_0)\ell_\theta}$, and $\eta_\mu = T^{-1/2}$.
Then, the sequence $\left\{\left(\theta_t, \mu_t\right)\right\}_{t=0}^{T-1}$ generated by algorithm \eqref{eq:algorithm_update_zero} with $U = [0, C_{0,\delta}]$ satisfies that
\begin{align}
\dfrac{1}{T} \sum_{t=0}^{T-1} \Big[f(\lambda(\theta^\star)) \hspace{-2pt}-\hspace{-2pt} f(\lambda(\theta_t))\Big] &\leq \dfrac{2\delta M}{\xi} \hspace{-2pt}+\hspace{-2pt} \dfrac{2M \hspace{-2pt}+\hspace{-2pt} C_{0,\delta}(M+\xi)}{\tilde\varepsilon T} \hspace{-2pt}+\hspace{-2pt}  \left(\dfrac{(M \hspace{-2pt}+\hspace{-2pt} \xi)^2}{\tilde\varepsilon} \hspace{-2pt}+\hspace{-2pt} \dfrac{(M \hspace{-2pt}+\hspace{-2pt} \xi)^2}{2}\right)\dfrac{1}{\sqrt{T}},\\
\dfrac{1}{T} \left[\sum_{t=0}^{T-1} -g(\lambda(\theta_t))\right]_+ &= 0.
\end{align}
\end{theorem}

The zero constraint violation presented in Theorem \ref{thm:zero} applies to the exact setting, which achieves the same order of convergence for the optimality gap as in Theorem \ref{thm:exact}.
We briefly introduce the ideas behind Theorem \ref{thm:zero} here. 
Adding the pessimistic term $\delta$ would shift the optimal solution from $\theta^\star$ to another point $\theta_\delta^\star$.
By leveraging the Slater's condition, we can upper-bound the sub-optimality gap $|f(\lambda(\theta^\star))-f(\lambda(\theta_\delta^\star))|$ by $\mc{O}(\delta)$.
Since the orders of convergence rates are the same for optimality gap and constraint violation in Theorem \ref{thm:exact}, we can choose $\delta$ to have the same order and then offset the constraint violation for the pessimistic problem \eqref{eq:prob_pessimistic_new}.
As a result, the constraint violation becomes zero for the original problem \eqref{eq:problem_theta}, and the optimality gap preserves its previous order.

%% file: file/6_experiment2.tex
\section{Numerical Experiment}
\label{sec: Numerical Experiment}
\begin{figure*}[tb]
    \centering
    \begin{subfigure}[b]{0.285\textwidth}
      \includegraphics[width=\textwidth]{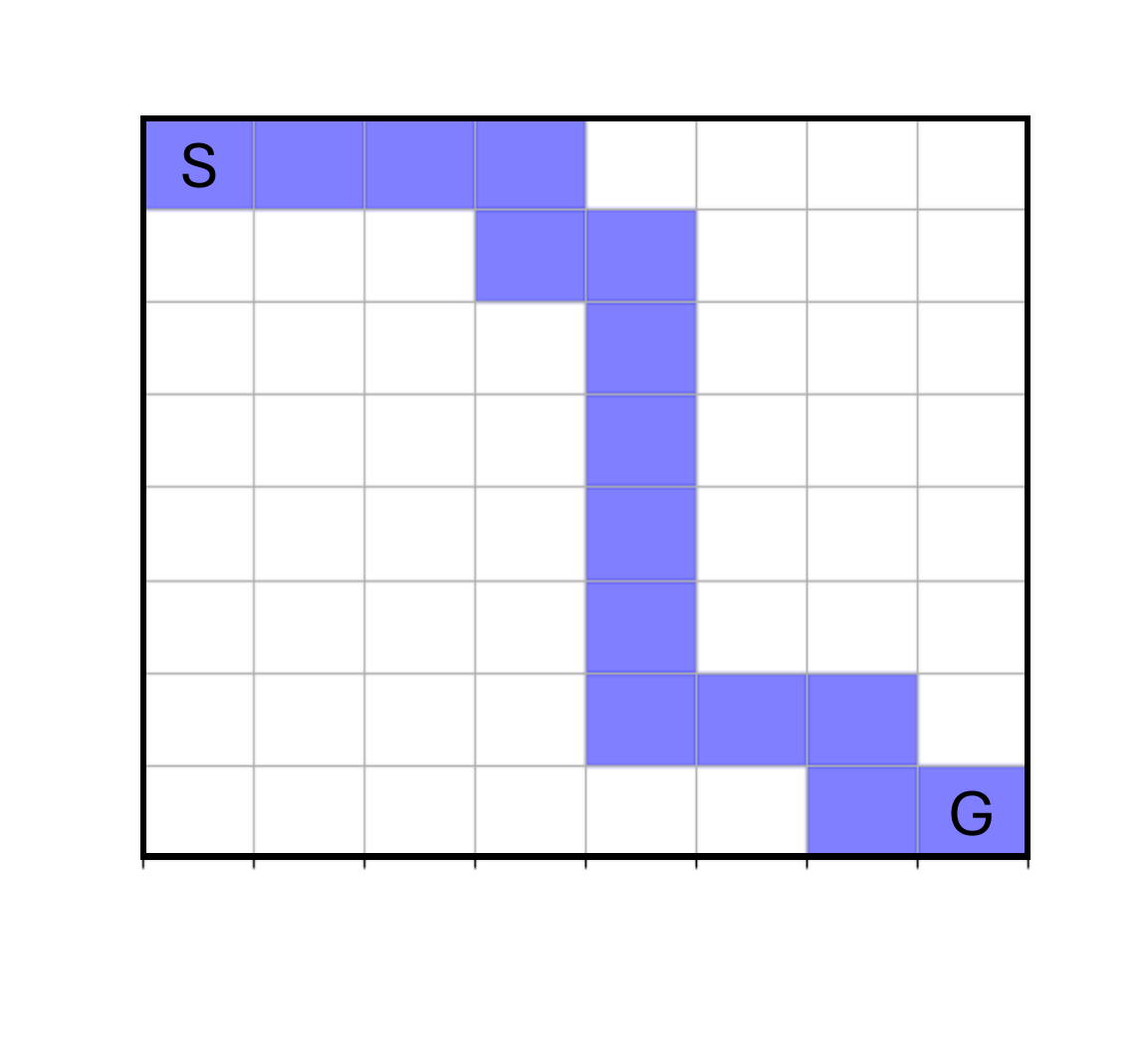}
      \caption{Gridworld\label{fig:gridworld8x8_env}}
    \end{subfigure}%
    \begin{subfigure}[b]{0.35\textwidth}
      \includegraphics[width=\textwidth]{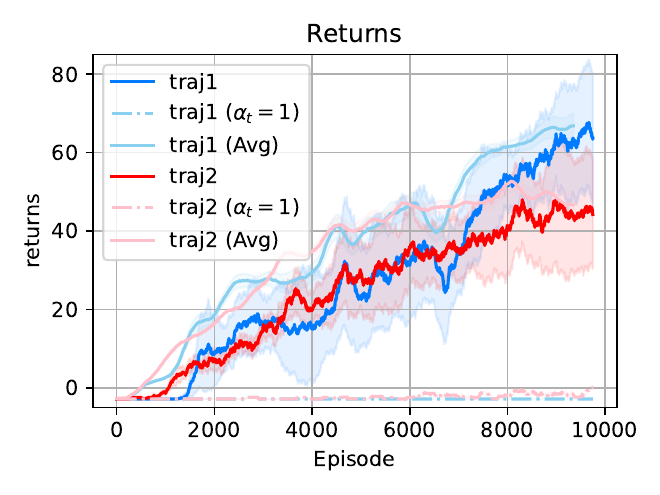}
      \caption{Returns\label{fig:gridworld8x8_returns}}
    \end{subfigure}%
    \hspace{0mm}
    \begin{subfigure}[b]{0.35\textwidth}
      \includegraphics[width=\textwidth]{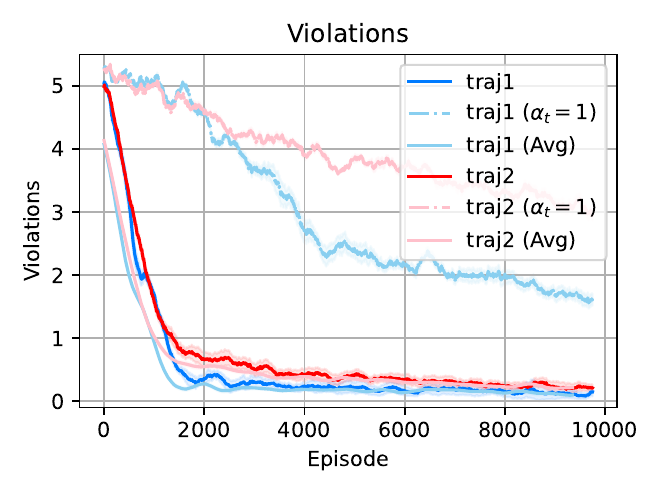}
      \caption{Violations\label{fig:gridworld8x8_violations}}
    \end{subfigure}%
    \caption{Two instances of $8 \times 8$ gridworld experiments under different reference trajectories.  
    For $\texttt{k}\in \texttt{\{1,2\}}$, \texttt{trajk}, 
    \texttt{trajk($\alpha_t=1$)}, and \texttt{trajk(Avg)} correspond to VR-PDPG, VR-PDPG with $\alpha_t=1$, and VR-PDPG with $\alpha_t=1$ and multi-trajectory estimation, respectively.
    }
    \label{fig:gridworld8x8}
\end{figure*}

\begin{figure*}[tb]
    \centering
    \begin{subfigure}[b]{0.35\textwidth}
      \includegraphics[width=\textwidth]{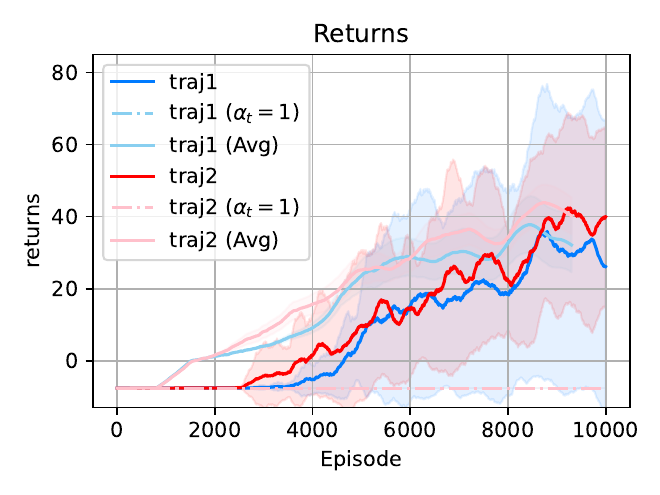}
      \caption{Returns\label{fig:gridworld20x20_returns}}
    \end{subfigure}%
    \hspace{1.8mm}
    \begin{subfigure}[b]{0.35\textwidth}
      \includegraphics[width=\textwidth]{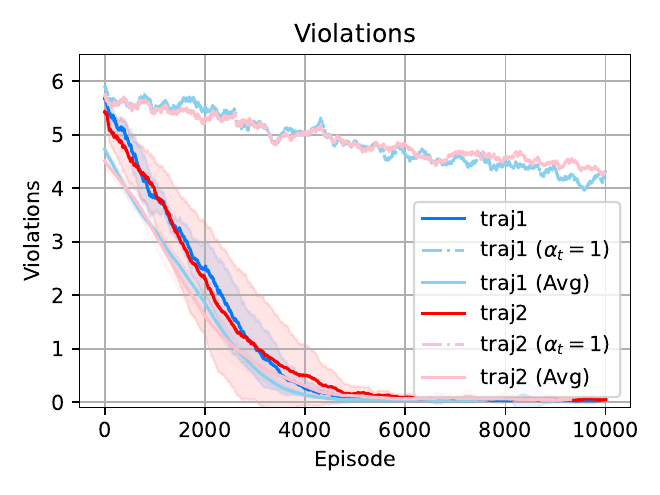}
      \caption{Violations\label{fig:gridworld20x20_violations}}
    \end{subfigure}%
    \caption{Two instances of $20 \times 20$ gridworld experiments under different reference trajectories.}
    \label{fig:gridworld20x20}
\end{figure*}

We validate Algorithm \ref{alg:pdpg} within a feasibility-constrained MDP problem (see Example \ref{exp:feasibility}), with the code is available at: \href{https://github.com/hyunin-lee/VR-PDPG}{https://github.com/hyunin-lee/VR-PDPG}.
The experiment involves tracking a reference trajectory, i.e., expert demonstration, which is a reformulation of Example \ref{exp:feasibility}, as follows:
\begin{equation}
\label{eq:exp_tracking}
\underset{\lambda \in \Lambda}\max\ f(\lambda) = \langle r,\lambda \rangle \ \ \text{s.t.} \ g(\lambda) = \norm{ \lambda - \lambda_e} \leq d_0,
\end{equation}
where $\lambda_e$ is the occupancy measure of the reference trajectory, computed prior to conducting the experiments. The experiments are conducted in two gridworld environments: $8 \times 8$ and $20 \times 20$, where the agent starts at the upper left corner (marked as `\texttt{S}' in Figure \ref{fig:gridworld8x8_env}) and aims to reach a goal located at the right bottom corner (marked as `\texttt{G}' in Figure \ref{fig:gridworld8x8_env}). For each grid size, we conduct two different experiments with varying reference trajectories, denoted as \texttt{traj1} and \texttt{traj2} in Figures \ref{fig:gridworld8x8_returns} and \ref{fig:gridworld8x8_violations}. To demonstrate the superiority of our VR-PDPG algorithm, we present the episode returns (Figures \ref{fig:gridworld8x8_returns} and \ref{fig:gridworld20x20_returns}) and constraint violations (Figures \ref{fig:gridworld8x8_violations} and \ref{fig:gridworld20x20_violations}). In Figures \ref{fig:gridworld8x8} and \ref{fig:gridworld20x20}, the solid lines represent the mean, and the shaded areas indicate the 95\% confidence level among different hyperparameters. For $8 \times 8$ grid world experiment (Figure \ref{fig:gridworld8x8}), we use $12$ different combinations of the hyperparameters, including the constraint violation $d_0 \in \{0.001, 0.005, 0.01, 0.05\}$ and the initial step size for primal parameter $\eta_\theta \in \{0.23, 0.24, 0.25\}$, and take the average performance. For $20 \times 20$ grid world experiment (Figure \ref{fig:gridworld20x20}), we use 60 different combinations of the hyperparameters, including the initial step size for the dual parameter $\eta_\mu = \{ 0.01,0.02,0.03,0.04,0.05\}$, the momentum update parameter $\alpha \in \{ 0.03, 0.05, 0.07, 0.09\}$, and the constant violation $d_0 \in \{ 0.001,0.005,0.01\}$. We then select the 12 best results out of 60 to compute the mean and variance (see Algorithm \ref{alg:pdpg} for hyperparameters). We set the step-sizes for the primal parameter $\eta_{\theta,t}$ and the momentum parameter $\alpha_t$ to be time-invariant, i.e., $\eta_{\theta} = \eta_{\theta,t}$ and $\alpha=\alpha_t$ for all $t \geq 0$.

To demonstrate the superiority of VR-PDPG, which incorporates momentum terms, we compare it with the following two baselines: (i) VR-PDPG with $\alpha_t=1, \forall t \geq 0$ (referred to as \texttt{traj1($\alpha_t=1$)} and \texttt{traj2($\alpha_t=1$)}), which do not utilize the momentum acceleration for convergence. (ii) VR-PDPG with $\alpha_t=1, \forall t \geq 0$ while employing multi-trajectory estimation (referred to as \texttt{traj1(Avg)} and \texttt{traj2(Avg)}), which uses the mean of $100$ trajectories with the same policy to estimate the gradient in \eqref{eq:pg_estimator} and occupancy measure in \eqref{eq: vanilla_OM_estimator}.

The results shown in Figures \ref{fig:gridworld8x8} and \ref{fig:gridworld20x20} demonstrate that the significant effectiveness of VR-PDPG. Specifically, comparing that trajectories $\texttt{traj1}$ (or $\texttt{traj2}$) and \texttt{traj1($\alpha_t=1$)} (or \texttt{traj2($\alpha_t=1$)}) reveals that VR-PDPG guarantees a faster convergence and higher returns within a fixed number of episodes. Additionally, the high standard deviation observed when estimating from a single trajectory suggests that improvements may be achieved through multi-trajectory estimation since it helps reduce the variance.
When comparing $\texttt{traj1}$ (or $\texttt{traj2}$) and $\texttt{traj1(Avg)}$ (or $\texttt{traj2(Avg)}$), we observe that averaging multiple trajectories does help improve the performance; however, the extent of this improvement is marginal.
It is also worth noting that although the multi-trajectory estimation demonstrates some performance, it consumes $100$ times more samples than our VR-PDPG algorithm, making it less attractive for practice purposes. 

%% file: file/7_conclusion.tex
\section{Conclusion}\label{sec: conclusion}
In this work, we propose primal-dual policy gradient (PDPG) algorithms to solve concave CMDP problems under the general soft-max policy with an over-parameterization assumption.
We first consider the exact setting and show that the PDPG algorithm achieves an $\widetilde{\mc{O}}(T^{-1/3})$ rate of convergence in terms of the average optimality gap and constraint violation. 
This rate can be further improved to $\widetilde{\mc{O}}(T^{-1/2})$ when the objective is strongly concave in the state-action visitation distribution.
Subsequently, in the sample-based setting, we propose the VR-PDPG algorithm and demonstrate that it can achieve an $\epsilon$-global optimality using a total of $\widetilde{\mc{O}}(\epsilon^{-4})$ samples.
In addition, by incorporating a diminishing pessimistic term into
the original constraint, we also prove that PDPG and VR-PDPG can attain a zero constraint violation while maintaining the same convergence rate for the optimality gap.

One important direction of future work lies in establishing a lower bound for concave CMDP problems under a general soft-max parameterization to verify the optimality of our upper bounds.
Furthermore, it is interesting to study whether geometric structures, such as entropy regularization  \cite{ying2021dual,liu2021policy,li2024faster} or policy mirror descent \cite{xiao2022convergence}, can be exploited to accelerate the convergence. 
Finally, as opposed to the average-iterate convergence results in this paper, an alternative direction is to find last-iterate convergent algorithms for concave CMDPs \cite{ding2024last}, as such algorithms may exhibit a higher degree of stability and could directly output the final solution.

%% file: app/appendix_1.tex
\vspace{10pt}
\begin{center}
\section*{SUPPLEMENTARY MATERIALS: Policy-based Methods for Concave CMDP}
\end{center}
\vspace{8pt}
The supplementary materials are organized as follows. In Appendix \ref{app:3.3}, we present a discussion on Assumption \ref{assump:parameterization} regarding the parameterization.
In Appendix \ref{app:2}, we document the supplementary materials corresponding to Sections \ref{sec:formulation}
and \ref{sec:alg}, including the proofs of Lemma \ref{lemma:duality} and auxiliary lemmas regarding the importance sampling weight.
Then, Appendix \ref{app:4} contains all proofs for the theoretical findings in Section \ref{sec:convergence}, including our main results on convergence rate in terms of optimality gap and constraint violation. 
Next, Appendix \ref{app:zero} is dedicated to the proof of the zero constraint violation in Section \ref{sec:zero}.
All other auxiliary lemmas are relegated to Appendix \ref{app:aux}.
Finally, we provide the details of the experiments in Appendix \ref{appendix:experiment}.

\section{Discussions About Assumption \ref{assump:parameterization}}\label{app:3.3}
To leverage the hidden concavity of problem (\ref{eq:problem_theta}) with respect to $\lambda$, it is natural to assume that there exists some desirable correspondence between $\lambda(\theta)$ and $\theta$.
However, as mentioned in Section \ref{sec:convergence}, requiring such correspondence to be one-to-one or invertible is too restrictive.
Although we can show that a one-to-one correspondence indeed exists under the direct parameterization and that the inverse map is Lipschitz continuous as long as there is a universal positive lower bound for the state occupancy measure $d^\pi(\cdot)$ (see Lemma \ref{lemma: Lipschitz continuity of Lambda}), this is not the case for many other parameterizations.
The soft-max policy, defined as 
\begin{equation}\label{eq:soft_max_policy}
    \pi_\theta(a\vert s) =\frac{\exp (\theta_{sa})}{\sum_{a^{\prime}\in \mc{A}} \exp \left(\theta_{sa^{\prime}}\right)},\ \forall\ (s,a)\in \mc{S}\times\mc{A},
\end{equation}
serves as a counterexample.
For a fixed vector $\theta_0\in\mbb{R}^{|\mc{S}||\mc{A}|}$, consider the set of parameters $$\{\theta\in\mbb{R}^{|\mc{S}||\mc{A}|}\mid \theta_{sa}=(\theta_0)_{sa}+k,\ \forall\ (s,a)\in \mc{S}\times\mc{A},\ \forall\ k\in\mbb{R}\}.$$
Then, it is clear that all parameters in the set correspond to the same policy $\pi_{\theta_0}$.
Thus, a one-to-one correspondence does not exist.
This motivates Assumption \ref{assump:parameterization}, which only requires the local existence of a continuous inverse $\lambda^{-1}$.
Assumption \ref{assump:parameterization} is able to accommodate the soft-max policy defined in (\ref{eq:soft_max_policy}).

\begin{lemma}[Lipschitz continuity of $\lambda^{-1}$ under direct parameterization]\label{lemma: Lipschitz continuity of Lambda}Suppose that $d_0 :=\min_{s\in\mc{S},\pi\in \Pi}d^\pi(s) >0$\footnote{Since $d^\pi(s)\geq (1-\gamma)\rho(s)$, this assumption is satisfied when there is an exploratory initial distribution, i.e., $\rho_0:=\min_{s\in \mc{S}} \rho(s) >0$.}.
For every two discounted state-action occupancy measures $\lambda_1, \lambda_2 \in \Lambda$, it holds that
\begin{equation*}
\left\|\lambda^{-1}(\lambda_1) - \lambda^{-1}(\lambda_2)\right\| \leq \frac{\sqrt{2(1+|\mc{A}|)}}{d_0}\|\lambda_1 -\lambda_2\|,
\end{equation*}
where $\lambda^{-1}(\cdot)$ maps a discounted state-action occupancy measure to the corresponding policy, defined as 
$$\pi(a\vert s) = \left[\lambda^{-1}(\lambda^\pi)\right]_{s,a} := \dfrac{\lambda^\pi(s,a)}{\sum_{a^\prime \in \mc{A}}\lambda^\pi(s,a^\prime)}.$$
\end{lemma}
\begin{proof}
Let $d_1(s) = \sum_{a}\lambda_1(s,a)$ and $d_2(s) = \sum_{a}\lambda_2(s,a)$ be the corresponding state occupancy measures.
Then,
\begin{equation*}
\begin{aligned}
\left[\lambda^{-1}(\lambda_1)\right]_{s,a} - \left[\lambda^{-1}(\lambda_2)\right]_{s,a} &= \frac{\lambda_1(s,a)}{d_1(s)}-\frac{\lambda_2(s,a)}{d_2(s)}\\
& = \frac{1}{d_1(s)} \left(\lambda_1(s,a)-\lambda_2(s,a)+\frac{d_2(s)-d_1(s)}{d_2(s)}\cdot\lambda_2(s,a)\right).
\end{aligned}
\end{equation*}
Therefore, one can compute
\begin{equation}\label{eq:lambda_inverse}
\begin{aligned}
&\left\|\lambda^{-1}(\lambda_1) - \lambda^{-1}(\lambda_2)\right\|^2 \\
=\ &\sum_{s,a}\left(\left[\lambda^{-1}(\lambda_1)\right]_{s,a} - \left[\lambda^{-1}(\lambda_2)\right]_{s,a}\right)^2\\
=\ &\sum_{s}\frac{1}{\left[d_1(s)\right]^2}\sum_a\left(\lambda_1(s,a)-\lambda_2(s,a)+\left[d_2(s)-d_1(s)\right]\cdot\frac{\lambda_2(s,a)}{{d_2(s)}}\right)^2\\
\leq\ &\sum_{s}\frac{2}{\left[d_1(s)\right]^2}\left(\sum_a\left[\lambda_1(s,a)-\lambda_2(s,a)\right]^2+\left(\left[d_2(s)-d_1(s)\right]\cdot\frac{\lambda_2(s,a)}{{d_2(s)}}\right)^2\right),
\end{aligned}
\end{equation}
where the last line follows from the inequality $(x+y)^2 \leq 2x^2+2y^2$.
For the second term inside the summation, we have that
\begin{equation}\label{eq:lambda-1_sub1}
\begin{aligned}
\sum_a \left(\left[d_2(s)-d_1(s)\right]\cdot\frac{\lambda_2(s,a)}{{d_2(s)}}\right)^2 &{=} \left[d_2(s)-d_1(s)\right]^2\cdot\left\|\frac{\lambda_2(s,\cdot)}{d_2(s)}\right\|^2\\
&{\overset{(i)}\leq} \left[d_2(s)-d_1(s)\right]^2\cdot\left\|\frac{\lambda_2(s,\cdot)}{d_2(s)}\right\|_1^2\\
&{=} \left[d_2(s)-d_1(s)\right]^2\\
&{=} \left[\sum_{a}\lambda_2(s,a) - \sum_{a}\lambda_1(s,a)\right]^2\\
&{\leq} |\mc{A}|\cdot \sum_a \left[\lambda_2(s,a) - \lambda_1(s,a)\right]^2,
\end{aligned}
\end{equation}
where $(i)$ is due to $\|\cdot\|\leq \|\cdot\|_1$ and the last step follows from the Cauchy-Schwarz inequality.
By substituting (\ref{eq:lambda-1_sub1}) into (\ref{eq:lambda_inverse}) and noting that $d^\pi(s)\geq d_0,\ \forall s\in \mc{S}, \pi\in \Pi$, it holds that
\begin{equation*}
\begin{aligned}
\left\|\lambda^{-1}(\lambda_1) - \lambda^{-1}(\lambda_2)\right\|^2 
&\leq \sum_{s}\frac{2(1+|\mc{A}|)}{\left[d_1(s)\right]^2}\left(\sum_a\left[\lambda_1(s,a)-\lambda_2(s,a)\right]^2\right)\\
&\leq \frac{2(1+|\mc{A}|)}{d_0^2}\sum_{s,a}\left[\lambda_1(s,a)-\lambda_2(s,a)\right]^2\\
&\leq \frac{2(1+|\mc{A}|)}{d_0^2}\left\|\lambda_1-\lambda_2\right\|^2.
\end{aligned}
\end{equation*}
The proof is completed by taking the square root of both sides of the inequality.
\end{proof}

%% file: app/appendix_2_3.tex
\section{Supplementary Materials for Sections \ref{sec:formulation} and \ref{sec:alg}}\label{app:2}
\subsection{Proof of Lemma \ref{lemma:duality}}

\begin{proof}
We note that $\Lambda$, the set of all unnormalized state-action occupancy measures, is a convex polytope having the expression
\begin{equation}\label{eq:Lambda}
\Lambda=\bigg\{\lambda \in \mbb{R}^{|S||A|} \Big\vert \lambda \geq \mb{0}, \sum_{a} \lambda(s, a)= \rho(s)+\dfrac{\gamma}{1-\gamma} \sum_{s^{\prime}, a^{\prime}} \mathbb{P}\left(s\vert s^{\prime}, a^{\prime}\right)\cdot \lambda\left(s^{\prime}, a^{\prime}\right), \forall s \in S\bigg\}.
\end{equation}
Since $\operatorname{cl}\big(\big\{\lambda(\theta)\vert \theta\in \mbb{R}^K\big\}\big) = \Lambda$, the nonconcave problem (\ref{eq:problem_theta}) is equivalent to the convex problem (\ref{eq: main problem}):
\begin{equation*}
\underset{\lambda \in \Lambda}\max \ f(\lambda) \quad \text {s.t.}\quad   {g}(\lambda) \geq {0}.
\end{equation*}
Therefore, the strong duality \textbf{\textit{(I)}} naturally holds under Assumption \ref{assump:slater} (see \cite{boyd2004convex}).

To prove \textbf{\textit{(II)}}, let $C\in \mbb{R}$. For every $\mu\geq {0}$ such that $\mc{D}(\mu)\leq C$, it holds that
\begin{equation}\label{eq:dual bound}
C\geq D(\mu) \overset{(i)}{\geq} f(\lambda(\widetilde \theta)) + \mu g(\lambda(\widetilde \theta)) \overset{(ii)}{\geq} f(\lambda(\widetilde \theta)) + \mu\xi ,
\end{equation}
where (i) follows from the definition of $\mc{D}(\mu)$ and $(ii)$ is due to Assumption \ref{assump:slater}.

Since $\xi >0$, (\ref{eq:dual bound}) gives rise to the bound $ \mu  \leq\big(C- f(\lambda(\widetilde \theta))\big) / \xi$.
Now, by letting $C = f(\lambda(\theta^\star))$, it results from the strong duality that $\left\{\mu \geq 0 \mid D(\mu) \leq C = f(\lambda(\theta^\star))\right\}$ becomes the set of optimal dual variables.
This completes the proof.
\end{proof}

\vspace{5pt}
\subsection{Importance Sampling Weight}\label{app: ISweight}
\begin{lemma}\label{lemma: ISweight}
Given any $\theta\in \mbb{R}^K$, let $\tau = \{(s_k,a_k)\}_{k=0}^{H-1}$ be a trajectory generated under $\pt$.
Then, for any target policy $\theta^\prime$, the IS weight $w(\tau \vert \theta^\prime,\theta)$ satisfies that
\begin{equation}\label{eq: def_p_tau}
\mbb{E}_{\tau\sim \pt,\rho}\brac{w(\tau\vert \theta^\prime, \theta) \widehat{\lambda}(\tau)} = \lambda_H(\theta^\prime),
\quad \mbb{E}_{\tau \sim \pt,\rho}\brac{w(\tau\vert \theta^\prime, \theta)\widehat{d}(\tau,\theta^\prime,r)} = \left[ \nabla_{\theta}\lambda_H(\theta^\prime)\right]^\top r,
\end{equation}
where $\lambda_H(\theta) = \sum_{k=0}^{H-1}\gamma^{k} \mathbb{P}\left(s_{k}=s, a_k = a\vert \pt, s_0 \sim \rho \right)$.
\end{lemma}
\begin{proof}
We first define the probability of having a trajectory $\tau$ with the length $H$ under the policy $\pi_\theta$ as follows
\begin{equation}
    \mbb P(\tau\vert \pi_\theta) = \rho(s_0) \cdot \pi_\theta(a_0 \vert s_0) \cdot \prod_{k = 1}^{H-1} \mbb P(s_k \vert s_{k-1}, a_{k-1}) \pi_\theta(a_k \vert s_k). 
\end{equation}
By the definitions of $w(\tau \vert \theta^\prime, \theta)$ in \eqref{eq:IS_def} and $\mbb P(\tau\vert \pi_\theta)$ in \eqref{eq: def_p_tau}, we deduce that 
\begin{equation}\label{eq:w_p}
    w(\tau \vert \theta^\prime, \theta) = \prod_{k=0}^{H-1} \dfrac{\pi_{\theta^\prime}(a_k \vert s_k)}{\pi_{\theta}(a_k \vert s_k)} = 
    \dfrac{\mbb P(\tau\vert \pi_{\theta^\prime})}{\mbb P(\tau\vert \pi_\theta)}.
\end{equation}
Together with the definition of $\widehat{\lambda}(\tau)$ in \eqref{eq: vanilla_OM_estimator}, we have that 
\begin{equation}
\begin{aligned}
    \mbb{E}_{\tau\sim \pt,\rho}\brac{w(\tau\vert \theta^\prime, \theta) \widehat{\lambda}(\tau)} &= 
    \sum_{\tau} \mbb P(\tau\vert \pi_\theta)\cdot \left(\dfrac{\mbb P(\tau\vert \pi_{\theta^\prime})}{\mbb P(\tau\vert \pi_\theta)} \cdot \sum_{k=0}^{H-1} \gamma^k \cdot \mathbbm{1}(s_k,a_k)\right)\\
    &=\sum_{\tau} \mbb P(\tau\vert \pi_{\theta^\prime}) \cdot \left(\sum_{k=0}^{H-1} \gamma^k \cdot \mathbbm{1}(s_k,a_k)\right) = \lambda_H(\theta^\prime).
\end{aligned}
\end{equation}
With similar steps, we can also derive that 
$\mbb{E}_{\tau \sim \pt,\rho}\brac{w(\tau\vert \theta^\prime, \theta)\widehat{d}(\tau,\theta^\prime,r)} = \big[ \nabla_{\theta}\lambda_H(\theta^\prime)\big]^\top r$.
\end{proof}

\begin{lemma}\label{lemma:ISweight}
Let Assumption \ref{assump:parameterization} hold, and suppose that $\{\theta_t\}_{t=0}^{T-1}$ is the primal sequence generated by Algorithm \ref{alg:pdpg}, i.e., $\theta_{t+1} = \theta_t + \eta_{\theta,t} d_{L,t}$ with $\eta_{\theta,t} = \eta_\theta/\norm{d_{L,t}}$.
Then, it holds that 
\begin{enumerate}
    \item[\textbf{\textit{(I)}}] For every $t\geq 0$ and any trajectory $\tau$ of length $H$, the IS weight satisfies $w(\tau|\theta_t, \theta_{t+1}) \leq \exp\big(2H M_{\pi,1} \cdot \eta_{\theta}\big)$.
    \item[\textbf{\textit{(II)}}] When we choose $\eta_{\theta} = T^{-1/2}$ and $H = \frac{1}{2}\log_{1/\gamma}T$, then $\exists W>0$ such that $w(\tau|\theta_t, \theta_{t+1})\leq W$. In addition, for any trajectory $\tau$ generated under $\pi_{\theta_{t+1}}$, it holds that 
\begin{align}
\mbb E_{\tau \sim \pi_{\theta_{t+1}},\rho}\big[w(\tau|\theta_t, \theta_{t+1})\big] &= 1;\\
\operatorname{Var}_{\tau \sim \pi_{\theta_{t+1}},\rho}\big(w(\tau|\theta_{t},\theta_{t+1}) \big) &\leq \dfrac{C_w }{T},
\end{align}
where $C_w:=H(W+1)\big[(8H+2)M_{\pi, 1}^2+2M_{\pi, 2}\big]$.
\end{enumerate}
\end{lemma}
\begin{proof}
\textbf{\textit{(I)}.} Based on the policy parameterization defined in \eqref{eq:policy_parameterization}, for any policy $\pi_{\theta_t}, \pi_{\theta_{t+1}}$ and any state-action pair $(s,a)\in \mc{S}\times\mc{A}$, it follows that
\begin{equation}
\begin{aligned}
     \dfrac{\pi_{\theta_t}(a\vert s)}{\pi_{\theta_{t+1}}(a\vert s)} &= 
     \dfrac{\exp\big(\psi(\theta_t;s, a)\big)}{\sum_{a^{\prime}\in \mc{A}} \exp \big(\psi\left(\theta_t;s, a^{\prime}\right)\big)} \cdot \dfrac{\sum_{a^{\prime}\in \mc{A}} \exp \big(\psi\left(\theta_{t+1};s, a^{\prime}\right)\big)}{\exp \big(\psi(\theta_{t+1};s, a)\big)}\\[8pt]
     &\leq \exp\big(\psi(\theta_t;s, a) - \psi(\theta_{t+1};s, a)\big) \cdot \max_{a^\prime \in \mc{A}} \left\{\exp\big(\psi\left(\theta_{t+1};s, a^{\prime}\right) -\psi\left(\theta_t;s, a^{\prime}\right)\big)\right\}\\[3pt]
     &\leq \exp\big(2M_{\pi, 1} \norm{\theta_t - \theta_{t+1}}\big)\\[5pt]
     &\leq \exp\big(2M_{\pi, 1} \cdot \eta_{\theta}\big),
\end{aligned}
\end{equation}
where the last two inequalities result from \eqref{eq:assum_para} in Assumption \ref{assump:parameterization} and the update rule of $\theta_t$, respectively. 
Lastly, from the definition of $w(\tau \vert \theta_{t}, \theta_{t+1})$ in \eqref{eq:IS_def} we have that 
\begin{equation}
    w(\tau \vert \theta_{t}, \theta_{t+1}) = \prod_{k=0}^{H-1} \dfrac{\pi_{\theta_{t}} (a_k \vert s_k)}{\pi_{\theta_{t+1}}(a_k \vert s_k)} \leq \exp\big( 2HM_{\pi,1} \cdot \eta_\theta \big).
\end{equation}

\textbf{\textit{(II)}.} 
Using the equality in \eqref{eq:w_p}, we expand the expectation as follows
\begin{equation}
\begin{aligned}
    \mbb E_{\tau \sim \pi_{\theta_{t+1}},\rho}\big[w(\tau|\theta_t, \theta_{t+1})\big]  = \sum_{\tau} \mbb P(\tau|\pi_{\theta_{t+1}}) \cdot \dfrac{\mbb P(\tau\vert \pi_{\theta_t})}{\mbb P(\tau\vert \pi_{\theta_{t+1}})}  = \sum_{\tau} \mbb P(\tau|\pi_{\theta_{t}}) = 1.
\end{aligned}
\end{equation}
Furthermore, the boundedness of $\operatorname{Var}_{\tau \sim \pi_{\theta_{t+1}},\rho}\big(w(\tau|\theta_{t},\theta_{t+1}) \big) \leq C_w/T$  follows the proof of Lemma B.1 in \cite{xu2019sample}, where  $C_w:=H(W+1)\big[(8H+2)M_{\pi, 1}^2+2M_{\pi, 2}\big]$.
\end{proof}

%% file: app/appendix_4.tex
\vspace{10pt}
\section{Supplementary Materials for Section \ref{sec:convergence}}\label{app:4}
\subsection{Proof of Lemma \ref{lemma:boundedness_smoothness}}

\begin{proof}
\textbf{\textit{(I)}.} Being a polytope implies that $\Lambda$ is closed and compact (see (\ref{eq:Lambda})).
Since $f(\cdot)$ and $g(\cdot)$ are concave in $\Lambda \subset \mbb{R}^{|\mc{S}| |\mc{A}|}$, they are also continuous.
Thus, we have that $f(\cdot)$ and $g(\cdot)$ are bounded on $\Lambda$, where we denote the upper bound by $M$. Note that we use the same bound $M$ for both $f(\cdot)$ and $g(\cdot)$ for simplicity. This completes the proof of \textbf{\textit{(I)}}.

 \textbf{\textit{(II)}.} The part \textbf{\textit{(II)}} follows directly from the proof of Lemma 5.3 (iii) in \cite{zhang2021convergence}, where the constant $\ell_\theta$ is defined as follows
 $$
\ell_\theta  =\dfrac{4 \ell_\lambda M_{\pi,1}^{2}}{(1-\gamma)^{4}}+\dfrac{8 M_{\pi,1}^{2} M_{\lambda}}{(1-\gamma)^{3}}+\dfrac{2 M_\lambda \left(M_{\pi,2}+M_{\pi,1}^{2}\right)}{(1-\gamma)^{2}}.
$$

\textbf{\textit{(III)}.} See the proof of Lemma E.3 in  \cite{zhang2021convergence}, where $L_{\theta,\lambda}^2$ is defined as follows
$$L_{\theta,\lambda}^2=\dfrac{8 \ell_\lambda^2 M_{\pi, 1}^2}{(1-\gamma)^6}+16  M_{\pi, 1}^2 \cdot M_{\lambda}^2\left(\dfrac{(H+1)^2}{(1-\gamma)^2}+\dfrac{1}{(1-\gamma)^4}\right).$$

\textbf{\textit{(IV)}.} From the boundedness of $f(\cdot)$ and $g(\cdot)$ on $\Lambda$ in part \textbf{\textit{(I)}} and the fact that $\mu \in U = [0, C_0]$, we have that
\begin{equation}
    \left|L(\lambda(\theta), \mu)\right|=|f(\lambda(\theta))+\mu g(\lambda(\theta))| \leq|f(\lambda(\theta))|+\mu|g(\lambda(\theta))| \leq M+C_0 M, \quad \forall \mu \in U.
\end{equation}
Similarly, as both $f(\lambda(\theta)$ and $g(\lambda(\theta)$ are $\ell_\theta$-smooth from part \textbf{\textit{(II)}}, we have that $L(\lambda(\theta), \mu)$ is $(1+C_0)\ell_\theta$-smooth w.r.t. $\theta$. 
In addition, by the properties in part \textbf{\textit{(III)}}, we further deduce that 
$\norm{\nabla_\theta L(\lambda(\theta),\mu) - \nabla_\theta L(\lambda_H(\theta),\mu)} \leq (1+C_0)L_{\theta,\lambda}\gamma^H.$
\end{proof}

\vspace{10pt}
\subsection{Proof of Lemma \ref{lemma: avg_L}}
\begin{proof}
We first prove \eqref{eq:prop_convergence} in the general concave setting.
Note that computing the primal update in algorithm (\ref{eq: algorithm_upate}) is equivalent to solving the following sub-problem:
\begin{equation}\label{eq:update_subproblem}
\begin{aligned}
\theta_{t+1} &= \theta_t +\eta_{\theta, t} \nabla_\theta L(\lambda(\theta_t),\mu_t)\\
&= \underset{{\theta\in \mbb{R}^K}}{\operatorname{argmax}}\ \left\{ L(\lambda(\theta_t),\mu_t) + (\theta-\theta_t)^\top\nabla_\theta L(\lambda(\theta_t),\mu_t)-\dfrac{1}{2\eta_{\theta,t}}\norm{\theta-\theta_t}^2\right\}\\
&=\underset{\theta\in \mbb{R}^K}{\operatorname{argmax}}\ \left\{ L(\lambda(\theta_t),\mu_t) + (\theta-\theta_t)^\top\nabla_\theta L(\lambda(\theta_t),\mu_t)-\dfrac{(1+C_0) \ell_\theta}{2}\norm{\theta-\theta_t}^2\right\}.
\end{aligned}
\end{equation}
Since $L(\lambda(\theta),\mu)$ is $(1+C_0)\ell_\theta$-smooth w.r.t. $\theta$ by Lemma \ref{lemma:boundedness_smoothness}, we obtain for every $\theta\in \mbb{R}^K$ that
\begin{equation}\label{eq:L_smooth}
\left| L(\lambda(\theta),\mu_t)-L(\lambda(\theta_t),\mu_t)-(\theta-\theta_t)^\top\nabla_\theta L(\lambda(\theta_t),\mu_t)  \right|\leq \dfrac{(1+C_0)\ell_\theta}{2}\norm{\theta-\theta_t}^2.
\end{equation}
Thus, the following ascent property holds
\begin{equation}\label{eq:ascent_property_strongly}
\begin{aligned}
L(\lambda(\theta),\mu_t) &\geq 
L(\lambda(\theta_t),\mu_t)+(\theta-\theta_t)^\top\nabla_\theta L(\lambda(\theta_t),\mu_t)-\dfrac{(1+C_0)\ell_\theta}{2} \norm{\theta-\theta_t}^2\\
&\geq L(\lambda(\theta),\mu_t)-(1+C_0)\ell_\theta\norm{\theta-\theta_t}^2,
\end{aligned}
\end{equation}
where we apply \eqref{eq:L_smooth} two times respectively in the two inequalities above.
On the basis of \eqref{eq:update_subproblem} and \eqref{eq:ascent_property_strongly}, it holds that
\begin{equation}\label{eq: lowerbound_L}
\begin{aligned}
L(\lambda(\theta_{t+1}),\mu_t) 
&\geq L(\lambda(\theta_{t}),\mu_t)+(\theta_{t+1}-\theta_t)^\top\nabla_\theta L(\lambda(\theta_t),\mu_t)-\dfrac{(1+C_0)\ell_\theta}{2} \norm{\theta_{t+1}-\theta_t}^2\\
&= \max_{\theta\in \mbb{R}^K} \left\{\hspace{-2pt}L(\lambda(\theta_t), \mu_t) + (\theta - \theta_t)^\top \nabla_\theta L(\lambda(\theta_t), \mu_t) - \dfrac{(1+C_0)\ell_\theta}{2}\norm{\theta - \theta_t}^2\hspace{-2pt}\right\}\\
&\geq \max_{\theta\in \mbb{R}^K}\left\{ L(\lambda(\theta),\mu_t)-(1+C_0)\ell_\theta \norm{\theta-\theta_t}^2\right\}.
\end{aligned}
\end{equation}
Now, we leverage the local invertibility of $\lambda(\cdot)$ to lower-bound the right-hand side of (\ref{eq: lowerbound_L}).
We first define that
\begin{equation}\label{eq:theta_epsilon}
\theta_{\varepsilon,t} := \lambda^{-1}_{\mc{V}_{\lambda(\theta_t)}}\left((1-\varepsilon)\lambda(\theta_t)+\varepsilon \lambda(\theta^\star)\right).
\end{equation}
According to Assumption \ref{assump:parameterization}, since $\varepsilon\leq \bar\varepsilon$, we have $(1-\varepsilon)\lambda(\theta_t)+\varepsilon \lambda(\theta^\star)\in \mc{V}_{\lambda(\theta_t)}$.
Thus, $\theta_{\varepsilon,t}$ is well-defined and $\theta_{\varepsilon,t}\in \mc{U}_{\theta_t}$.
By definition, the composition of $\lambda:\Theta \rightarrow \Lambda$ and $\lambda^{-1}_{\mc{V}_{\lambda(\theta_t)}}: \mc{V}_{\lambda(\theta_t)} \rightarrow \mc{U}_{\theta_t} $ is the identity map on $\mc{V}_{\lambda(\theta_t)}$.
Together with the facts that $L(\cdot, \mu)$ is concave in $\lambda$, we have
\begin{equation}\label{eq: lowerbound_L_sub1}
\begin{aligned}
   L(\lambda({\theta_{\varepsilon,t}}), \mu_t) &=L\left(\lambda \circ \lambda^{-1}_{\mc{V}_{\lambda(\theta_t)}}\left((1-\varepsilon)\lambda(\theta_t)+\varepsilon \lambda(\theta^\star)\right), \mu_t\right)\\
    &=L\left((1-\varepsilon)\lambda({\theta_t})+\varepsilon \lambda({\theta^\star}), \mu_t\right)\\
    &\geq (1-\varepsilon)L(\lambda({\theta_t}), \mu_t)+\varepsilon L(\lambda({\theta^\star}), \mu_t).
\end{aligned}
\end{equation}
Additionally, the Lipschitz continuity of $\lambda^{-1}_{\mc{V}_{\lambda(\theta_t)}}(\cdot)$ implies that 
\begin{equation}\label{eq: lowerbound_L_sub2}
\begin{aligned}
\norm{\theta_{\varepsilon,t} - \theta_t}^2  
&= \norm{\lambda^{-1}_{\mc{V}_{\lambda(\theta_t)}}\big((1-\varepsilon)\lambda(\theta_t)+\varepsilon \lambda(\theta^\star)\big) - \lambda^{-1}_{\mc{V}_{\lambda(\theta_t)}}\left(\lambda(\theta_t)\right)}^2 \\
&\leq  L_\lambda^2 \norm{(1-\varepsilon)\lambda(\theta_t)+\varepsilon \lambda(\theta^\star) - \lambda(\theta_t)}^2\\
&\leq \varepsilon^2 L_\lambda^2\norm{\lambda({\theta^\star})-\lambda({\theta_t})}^2\\
    & \leq \dfrac{2\varepsilon^2 L_\lambda^2}{(1-\gamma)^2},
\end{aligned}
\end{equation}
where the last inequality uses the diameter of $\Lambda$, i.e., $ \max_{\lambda_1,\lambda_2\in \Lambda} \|\lambda_1-\lambda_2\|\leq \frac{\sqrt{2}}{1-\gamma}$.
By substituting $\theta_{\varepsilon,t}$ into (\ref{eq: lowerbound_L}) and using the inequalities \eqref{eq: lowerbound_L_sub1} and \eqref{eq: lowerbound_L_sub2}, it holds that
\begin{equation*}
\begin{aligned}
    L(\lambda(\theta_{t+1}), \mu_t) &\geq \max_{\theta\in \mbb{R}^K}\left\{L(\lambda(\theta),\mu_t)-(1+C_0) \ell_\theta \norm{\theta-\theta_t}^2\right\}\\
    &\geq L(\lambda(\theta_{\varepsilon,t}),\mu_t)-(1+C_0) \ell_\theta\norm{\theta_{\varepsilon,t}-\theta_t}^2\\
    &\geq (1-\varepsilon) L(\lambda(\theta_t), \mu_t)+\varepsilon L(\lambda(\theta^\star), \mu_t)-\dfrac{2\varepsilon^2 (1+C_0) \ell_\theta L_\lambda^2}{(1-\gamma)^2},
\end{aligned}
\end{equation*}
which implies that
\begin{equation}\label{eq:to_derive_recursion}
L(\lambda(\theta^\star),\mu_t)- L(\lambda(\theta_{t+1}), \mu_t)\leq (1-\varepsilon)\big[ L(\lambda(\theta^\star),\mu_t)-L(\lambda(\theta_{t}), \mu_t) \big]+\dfrac{2\varepsilon^2 (1+C_0) \ell_\theta L_\lambda^2}{(1-\gamma)^2}.
\end{equation}
Consequently, one can obtain the recursion
\begin{equation}\label{eq:recursion_convex}
\begin{aligned}
&\quad L(\lambda(\theta^\star),\mu_{t+1})-L(\lambda(\theta_{t+1}), \mu_{t+1})\\
&= \big[ L(\lambda(\theta^\star),\mu_t)- L(\lambda(\theta_{t+1}), \mu_t)\big] +\big[L(\lambda(\theta^\star),\mu_{t+1})-L(\lambda(\theta^\star), \mu_t) \big] \\
&\quad + \big[L(\lambda(\theta_{t+1}),\mu_t)- L(\lambda(\theta_{t+1}), \mu_{t+1}) \big]\\
&{\overset{(i)}{\leq}} (1\hspace{-2pt}-\hspace{-2pt}\varepsilon)\big[L(\lambda(\theta^\star),\mu_t)\hspace{-2pt}-\hspace{-2pt} L(\lambda(\theta_{t}), \mu_t) \big] \hspace{-2pt}+ \hspace{-2pt}\dfrac{2\varepsilon^2 (1 \hspace{-2pt}+\hspace{-2pt} C_0) \ell_\theta L_\lambda^2}{(1-\gamma)^2} \hspace{-2pt}+\hspace{-2pt} \big[L(\lambda(\theta^\star),\mu_{t+1})\hspace{-2pt}-\hspace{-2pt} L(\lambda(\theta^\star), \mu_t) \big]\\
&\quad+ \big[L(\lambda(\theta_{t+1}),\mu_t)- L(\lambda(\theta_{t+1}), \mu_{t+1}) \big]\\
&{\overset{(ii)}{\leq}} (1-\varepsilon)\big[ L(\lambda(\theta^\star),\mu_t )- L(\lambda(\theta_{t}), \mu_t) \big]+\dfrac{2\varepsilon^2 (1+C_0) \ell_\theta L_\lambda^2}{(1-\gamma)^2} + 2\eta_\mu M^2,
\end{aligned}
\end{equation}
where we use \eqref{eq:to_derive_recursion} in $(i)$ . Step $(ii)$ is due to the bound
\begin{equation}\label{eq:mu_t_mu_t+1}
\begin{aligned}
\big|L(\lambda(\theta),\mu_t)- L(\lambda(\theta),\mu_{t+1})\big| &= \big|(\mu_{t}-\mu_{t+1}) \cdot g(\lambda(\theta))\big|\\
&=\Big|\big[\mu_{t}-\mc{P}_U\left(\mu_t - \eta_\mu\nabla_\mu L(\lambda(\theta_t),\mu_t) \right)\big] \cdot g(\lambda(\theta))\Big|\\
&\leq \big|\eta_\mu\nabla_\mu L(\lambda(\theta_t),\mu_t) \cdot g(\lambda(\theta)) \big|\\
&= \big|\eta_\mu g(\lambda(\theta_t)) \cdot g(\lambda(\theta)) \big|\\
&\leq \eta_\mu M^2,\quad \forall\ \theta\in \mbb{R}^K,
\end{aligned}
\end{equation}
where the two inequalities above result from the non-expansive property of the projection operator and the boundedness of $g(\lambda(\theta))$, i.e., $|g(\lambda(\theta))|\leq M$, respectively.
Utilizing the recursion \eqref{eq:recursion_convex}, we derive that
\begin{equation*}
\begin{aligned}
&\quad L(\lambda(\theta^\star),\mu_{t+1})-L(\lambda(\theta_{t+1}), \mu_{t+1})\\
&\leq (1-\varepsilon)\big[ L(\lambda(\theta^\star),\mu_t)- L(\lambda(\theta_{t}), \mu_t) \big] + 2\varepsilon^2(1+C_0)\ell_\theta L_\lambda^2 + 2\eta_\mu M^2\\
&\leq (1-\varepsilon)^2\big[L(\lambda(\theta^\star),\mu_{t-1})- L(\lambda(\theta_{t-1}), \mu_{t-1}) \big]+(1+1-\varepsilon)\left[\dfrac{2\varepsilon^2 (1+C_0) \ell_\theta L_\lambda^2}{(1-\gamma)^2} + 2\eta_\mu M^2\right]\\
&\leq (1-\varepsilon)^{t+1}\big[ L(\lambda(\theta^\star),\mu_{0})-L(\lambda(\theta_{0}), \mu_{0})\big]+\sum_{i=0}^{t}(1-\varepsilon)^i\left[\dfrac{2\varepsilon^2 (1+C_0) \ell_\theta L_\lambda^2}{(1-\gamma)^2} + 2\eta_\mu M^2\right]\\
&=(1-\varepsilon)^{t+1}\big[L(\lambda(\theta^\star),\mu_{0})- L(\lambda(\theta_{0}), \mu_{0})\big] + \dfrac{1-(1-\varepsilon)^{t+1}}{\varepsilon}\left[\dfrac{2\varepsilon^2 (1+C_0) \ell_\theta L_\lambda^2}{(1-\gamma)^2} + 2\eta_\mu M^2\right],
\end{aligned}
\end{equation*}
which is equivalent to
\begin{equation*}
\begin{aligned}
&\quad L(\lambda(\theta^\star),\mu_{t})-L(\lambda(\theta_{t}), \mu_{t})\\
&\leq (1\hspace{-1pt}-\hspace{-1pt} \varepsilon)^{t}\big[L(\lambda(\theta^\star),\mu_{0})-L(\lambda(\theta_{0}), \mu_{0})\big] \hspace{-2pt}+ \hspace{-2pt}\left({1\hspace{-2pt}-\hspace{-2pt}(1\hspace{-1pt}-\hspace{-1pt}\varepsilon)^{t}}\right)\left(\dfrac{2\varepsilon (1\hspace{-1pt}+\hspace{-1pt}C_0) \ell_\theta L_\lambda^2}{(1-\gamma)^2} + \dfrac{2\eta_\mu M^2}{\varepsilon}\right), \ \forall\ t\geq 0.
\end{aligned}
\end{equation*}
Summing the above inequality over $t=0,1,\dots,T-1$ yields that
\begin{equation*}
\begin{aligned}
& \sum_{t=0}^{T-1}\big[L(\lambda(\theta^\star),\mu_{t})-L(\lambda(\theta_{t}), \mu_{t})\big]\\
\leq& \sum_{t=0}^{T-1} (1-\varepsilon)^{t}\big[L(\lambda(\theta^\star),\mu_{0})- L(\lambda(\theta_{0}), \mu_{0})\big] + \left({1-(1-\varepsilon)^{t}}\right)\left(\dfrac{2\varepsilon (1+C_0) \ell_\theta L_\lambda^2}{(1-\gamma)^2} + \dfrac{2\eta_\mu M^2}{\varepsilon}\right)\\
=& \dfrac{1-(1-\varepsilon)^T}{\varepsilon}\big[ L(\lambda(\theta^\star),\mu_{0})- L(\lambda(\theta_{0}), \mu_{0})\big]+\left(T\hspace{-2pt}-\hspace{-2pt}\dfrac{1\hspace{-2pt}-\hspace{-2pt}(1\hspace{-2pt}-\hspace{-2pt}\varepsilon)^T}{\varepsilon}\right)\left(\dfrac{2\varepsilon (1\hspace{-2pt}+\hspace{-2pt}C_0) \ell_\theta L_\lambda^2}{(1-\gamma)^2}  \hspace{-2pt}+ \hspace{-2pt} \dfrac{2\eta_\mu M^2}{\varepsilon}\right)\\
\leq& \dfrac{1}{\varepsilon}\big[L(\lambda(\theta^\star),\mu_{0})-L(\lambda(\theta_{0}), \mu_{0})\big] + T\left(\dfrac{2\varepsilon (1+C_0) \ell_\theta L_\lambda^2}{(1-\gamma)^2} + \dfrac{2\eta_\mu M^2}{\varepsilon}\right).
\end{aligned}
\end{equation*}
The proof of \eqref{eq:prop_convergence} is completed by dividing both sides of the inequality by $T$.

\vspace{8pt}
Now we proceed to the case where $f(\cdot)$ is $\sigma$-strongly concave with respect to $\lambda$. 
We begin with \eqref{eq: lowerbound_L}:
\begin{equation}\label{eq: lowerbound_sc}
    L(\lambda(\theta_{t+1}),\mu_t) \geq \max_{\theta\in \mbb{R}^K}\left\{ L(\lambda(\theta),\mu_t)-(1+C_0)\ell_\theta \norm{\theta-\theta_t}^2\right\},.
\end{equation}
For $\varepsilon\leq \bar\varepsilon$, we define $\theta_{\varepsilon,t} := \lambda^{-1}_{\mc{V}_{\lambda(\theta_t)}}\left((1-\varepsilon)\lambda(\theta_t)+\varepsilon \lambda(\theta^\star)\right)$ similarly as (\ref{eq:theta_epsilon}).
Due to the fact that $L(\cdot, \mu)$ is $\sigma$-strongly concave in $\lambda$, which results from the $\sigma$-strongly concavity of $f(\cdot)$ and the concavity of $g(\cdot)$, we have that
\begin{equation}\label{eq: lowerbound_L_sub_sc}
\begin{aligned}
   & L(\lambda(\theta_{\varepsilon,t}), \mu_t)\\
   =&L\left(\lambda \circ \lambda^{-1}_{\mc{V}_{\lambda(\theta_t)}}\left((1-\varepsilon)\lambda(\theta_t)+\varepsilon \lambda(\theta^\star)\right), \mu_t\right)\\
    =&L\left((1-\varepsilon)\lambda({\theta_t})+\varepsilon \lambda({\theta^\star}), \mu_t\right)\\
    \geq& (1-\varepsilon)L(\lambda({\theta_t}), \mu_t)+\varepsilon L(\lambda({\theta^\star}), \mu_t) +\dfrac{\sigma}{2}\varepsilon(1-\varepsilon)\norm{\lambda(\theta^\star)-\lambda(\theta_t)}^2.
\end{aligned}
\end{equation}
By Assumption \ref{assump:parameterization}, the Lipschitz continuity of $\lambda^{-1}_{\mc{V}_{\lambda(\theta_t)}}$ implies that
\begin{equation}\label{eq: lip_of_l_theta_sc}
\begin{aligned}
\norm{\theta_{\varepsilon,t} - \theta_t}^2  
&= \norm{\lambda^{-1}_{\mc{V}_{\lambda(\theta_t)}}\left((1-\varepsilon)\lambda(\theta_t)+\varepsilon \lambda(\theta^\star)\right) - \lambda^{-1}_{\mc{V}_{\lambda(\theta_t)}}\left(\lambda(\theta_t)\right)}^2 \\
&\leq  L_\lambda^2 \norm{ (1-\varepsilon)\lambda(\theta_t)+\varepsilon \lambda(\theta^\star) - \lambda(\theta_t)}^2\\
&\leq \varepsilon^2 L_\lambda^2\norm{\lambda({\theta^\star})-\lambda({\theta_t})}^2.
\end{aligned}
\end{equation}
Substitute $\theta_{\varepsilon,t}$ into the right-hand side of (\ref{eq: lowerbound_sc}), we have that
\begin{equation*}
\begin{aligned}
&\quad L(\lambda(\theta_{t+1}), \mu_t) \\
&\geq
 \max_{\theta\in \mbb{R}^K}\left\{L(\lambda(\theta),\mu_t)-(1+C_0)\ell_\theta \norm{\theta-\theta_t}^2\right\}\\
&\geq \max_{0\leq\varepsilon\leq \bar\varepsilon}\left\{L(\lambda(\theta_{\varepsilon,t}),\mu_t)-(1+C_0)\ell_\theta \norm{\theta_{\varepsilon,t}-\theta_t}^2\right\}\\
&\geq \max_{0\leq\varepsilon\leq \bar\varepsilon}\left\{(1\hspace{-2pt}-\hspace{-2pt}\varepsilon) L(\lambda(\theta_t),\mu_t)\hspace{-2pt} +\hspace{-2pt}\varepsilon L(\lambda(\theta^\star),\mu_t)\hspace{-1pt} + \hspace{-1pt}\left(\dfrac{\sigma}{2}\varepsilon(1\hspace{-2pt}-\hspace{-2pt}\varepsilon)-\varepsilon^2 (1\hspace{-2pt}+\hspace{-2pt}C_0)\ell_\theta L_\lambda^2\right)\norm{\lambda(\theta^\star)\hspace{-2pt}-\hspace{-2pt}\lambda(\theta_t)}^2\right\},
\end{aligned}
\end{equation*}
where we use \eqref{eq: lowerbound_L_sub_sc} and \eqref{eq: lip_of_l_theta_sc} in the last inequality.
Consequently,
\begin{equation}\label{eq: lowerbound_L_sub3_sc}
\begin{aligned}
    L(\lambda(\theta^\star), \mu_t)\hspace{-2pt}-\hspace{-2pt}L(\lambda(\theta_{t+1}), \mu_t) &\leq \min_{0\leq\varepsilon\leq \bar\varepsilon}\bigg\{(1\hspace{-2pt}-\hspace{-2pt}\varepsilon)\big[ L(\lambda(\theta^\star), \mu_t) \hspace{-2pt}-\hspace{-2pt} L(\lambda(\theta_t), \mu_t)\big]\\
    &\qquad \quad -\hspace{-2pt}\left(\hspace{-2pt}\dfrac{\sigma \varepsilon (1\hspace{-2pt}-\hspace{-2pt} \varepsilon)}{2} \hspace{-2pt}-\hspace{-2pt} \varepsilon^2 (1+C_0)\ell_\theta L_\lambda^2 \hspace{-2pt} \right) \left\|\lambda(\theta^\star) \hspace{-2pt}-\hspace{-2pt}\lambda(\theta_t)\right\|^2 \hspace{-2pt}\bigg\}.
\end{aligned}
\end{equation}
We note that
\begin{equation}\label{eq:before_tilde_epsilon}
\dfrac{\sigma \varepsilon (1-\varepsilon)}{2} - \varepsilon^2 (1+C_0)\ell_\theta L_\lambda^2\geq 0,\ \text{ if }\ 0\leq \varepsilon \leq \dfrac{\sigma}{\sigma+2  L_\lambda^2 (1+C_0)\ell_\theta }.
\end{equation}
By letting $\tilde{\varepsilon} := \min\{\bar\varepsilon, {\sigma}/{\left(\sigma+2  L_\lambda^2 (1+C_0)\ell_\theta\right)}\} \leq \bar\varepsilon$, it follows from (\ref{eq: lowerbound_L_sub3_sc}) that
\begin{equation}\label{eq: iterative_L_sc}
\begin{aligned}
&\quad L(\lambda(\theta^\star), \mu_t)-L(\lambda(\theta_{t+1}), \mu_t)\\
&\leq (1\hspace{-2pt}-\hspace{-2pt}\tilde\varepsilon)\big[L(\lambda(\theta^\star), \mu_t) \hspace{-2pt}-\hspace{-2pt} L(\lambda(\theta_t), \mu_t)\big] \hspace{-2pt}-\hspace{-2pt}
\left(\dfrac{\sigma \tilde\varepsilon (1-\tilde\varepsilon)}{2} - {\tilde\varepsilon}^2 (1\hspace{-2pt}+\hspace{-2pt}C_0)\ell_\theta L_\lambda^2 \right) \norm{\lambda(\theta^\star)\hspace{-2pt}-\hspace{-2pt}\lambda(\theta_t)}^2\\
&\leq (1\hspace{-2pt}-\hspace{-2pt}\tilde\varepsilon)\big[L(\lambda(\theta^\star), \mu_t) \hspace{-2pt}-\hspace{-2pt} L(\lambda(\theta_t, \mu_t)\big],
\end{aligned}
\end{equation}
where the second inequality results from (\ref{eq:before_tilde_epsilon}).
We rearrange the terms in (\ref{eq: iterative_L_sc}) to obtain
\begin{equation*}
    L(\lambda(\theta^\star), \mu_t) - L(\lambda(\theta_{t+1}), \mu_t) \leq  \dfrac{1-\tilde\varepsilon}{\tilde\varepsilon}\big[L(\lambda(\theta_{t+1}), \mu_t) - L(\lambda(\theta_t), \mu_t)\big],
\end{equation*}
which implies that
\begin{equation*}
\begin{aligned}
L(\lambda(\theta^\star), \mu_t) - L(\lambda(\theta_t), \mu_t) &=  \big[L(\lambda(\theta^\star), \mu_t) - L(\lambda(\theta_{t+1}),\mu_t)\big] + \big[ L(\lambda(\theta_{t+1}),\mu_t)- L(\lambda(\theta_t),\mu_t)\big] \\
&\leq \left(\dfrac{1-\tilde\varepsilon}{\tilde\varepsilon}+1\right)\big[ L(\lambda(\theta_{t+1}),\mu_t)- L(\lambda(\theta_t),\mu_t)\big]\\
&=\dfrac{1}{\tilde\varepsilon}\left[L(\lambda(\theta_{t+1}),\mu_t)- L(\lambda(\theta_t),\mu_t)\right].
\end{aligned}
\end{equation*}
Summing it over $t = 0,\dots, T-1$, we have that
\begin{equation*}
\begin{aligned}
    &\quad \sum_{t=0}^{T-1}[ L(\lambda(\theta^\star), \mu_t) - L(\lambda(\theta_{t}), \mu_t)]\\ 
    &\leq \dfrac{1}{\tilde\varepsilon}\sum_{t=0}^{T-1} \big[L(\lambda(\theta_{t+1}),\mu_t) - L(\lambda(\theta_t),\mu_t)\big] \\
    &= \dfrac{1}{\tilde\varepsilon}\left(L(\lambda(\theta_T),\mu_{T-1}) - L(\lambda(\theta_0),\mu_0) + \sum_{t=0}^{T-2}\big[ L(\lambda(\theta_{t+1}),\mu_{t})-L(\lambda(\theta_{t+1}),\mu_{t+1})\big]\right)\\
    &{\overset{(i)}{\leq}} \dfrac{1}{\tilde\varepsilon}\big[L(\lambda(\theta_T),\mu_{T-1}) - L(\lambda(\theta_0),\mu_0)+ (T-1)\eta_\mu M^2\big],
\end{aligned}
\end{equation*}
where we use (\ref{eq:mu_t_mu_t+1}) to bound the difference $L(\lambda(\theta_{t+1}),\mu_{t})-L(\lambda(\theta_{t+1}),\mu_{t+1})$ in $(i)$.
The proof is completed by dividing both sides of the inequality by $T$.
\end{proof}

\subsection{Proof of Lemma \ref{lemma:obtain_opt_const_exact}}
\begin{proof}
Indeed, Lemma \ref{lemma:obtain_opt_const_exact} is a simplified version of Lemma \ref{lemma:obtain_opt_const_sample}, and its proof simply follows from the proof of Lemma \ref{lemma:obtain_opt_const_sample} in Appendix \ref{app:obtain_opt_const_sample}.
The only subtlety is that in the exact setting, \eqref{eq: u-u} and \eqref{eq: u-u2} become
\begin{equation}
\begin{aligned}
\left(\mu_{t+1}-\mu\right)^2 
&\leq (\mu_t - \mu)^2 - 2\eta_{\mu} (\mu_t-\mu) \cdot g(\lambda(\theta_t)) + (\eta_\mu g(\lambda(\theta_t)))^2\\
&\leq (\mu_t - \mu)^2 - 2\eta_{\mu} (\mu_t-\mu) \cdot g(\lambda(\theta_t)) + \eta_\mu^2 M^2.
\end{aligned}
\end{equation}
Therefore, we can show that in the absence of approximation errors, the optimality gap \eqref{eq:optimality_gap_lemma} and constraint violation \eqref{eq:constraint_violation_lemma} reduce to \eqref{eq:optimality_gap_lemma_exact} and \eqref{eq:constraint_violation_lemma_exact}, respectively.
\end{proof}

\vspace{5pt}
\subsection{Proof of Theorem \ref{thm:exact} (Exact Setting)}
\begin{proof}
The proof of Theorem \ref{thm:exact} relies on Lemmas \ref{lemma: avg_L} and \ref{lemma:obtain_opt_const_exact}.
Combining \eqref{eq:prop_convergence} in Lemma \ref{lemma: avg_L} and \eqref{eq:optimality_gap_lemma_exact} in Lemma \ref{lemma:obtain_opt_const_exact}, we have that
\begin{equation*}
\begin{aligned}
    &\quad \dfrac{1}{T} \sum_{t=0}^{T-1} f(\lambda(\theta^\star)) - f(\lambda(\theta_t))\\
    &\leq \dfrac{L(\lambda(\theta^\star),\mu_{0})-L(\lambda(\theta_{0}), \mu_{0})}{\varepsilon T} + \dfrac{2\varepsilon (1+C_0) \ell_\theta L_\lambda^2}{(1-\gamma)^2} + \dfrac{2 \eta_\mu M^2}{\varepsilon} 
    +\dfrac{\mu_0^2-\mu_T^2}{2T\eta_\mu} + \dfrac{\eta_\mu M^2}{2}\\
    &{\overset{(i)}{\leq}}\dfrac{2M}{\varepsilon T} + \dfrac{2\varepsilon (1+C_0) \ell_\theta L_\lambda^2}{(1-\gamma)^2} + \dfrac{2 \eta_\mu M^2}{\varepsilon} 
    +\dfrac{\mu_0^2-\mu_T^2}{2T\eta_\mu} + \dfrac{\eta_\mu M^2}{2}\\
    &{\overset{(ii)}{\leq}} \dfrac{2M +M^2/2}{T^{2/3}} + \dfrac{2(1+C_0)\ell_\theta L_\lambda^2}{(1-\gamma)^2 T^{1/3}} +\dfrac{2M^2}{T^{1/3}},
\end{aligned}
\end{equation*}
where step $(i)$ follows from the fact that $\mu_0 = 0$ specified in Theorem \ref{thm:exact}, and thereby $L(\lambda(\theta^\star),\mu_{0})-L(\lambda(\theta_{0}), \mu_{0}) = f(\lambda(\theta^\star)) -  f(\lambda(\theta_{0})) \leq 2M$ from Lemma \ref{lemma:boundedness_smoothness}.
Since $\mu_0^2 - \mu_T^2 \leq 0$, the third term in step $(i)$ can be dropped. Then, step $(ii)$ is obtained by specifying $\varepsilon = T^{-1/3}\leq \bar \varepsilon$ and $\eta_\mu = T^{-2/3}$. This completes the proof of the optimality gap in \eqref{eq:optimality_gap}.

Similarly, from  \eqref{eq:constraint_violation_lemma_exact} in Lemma \ref{lemma:obtain_opt_const_exact} and Lemma \ref{lemma: avg_L}, it holds that 
\begin{equation*}
\begin{aligned}
\dfrac{1}{T}\left[\sum_{t=0}^{T-1}-g(\lambda(\theta_t))\right]_+
&\leq  \dfrac{L(\lambda(\theta^\star),\mu_{0})-L(\lambda(\theta_{0}), \mu_{0})}{\varepsilon T} +\dfrac{2\varepsilon (1+C_0) \ell_\theta L_\lambda^2}{(1-\gamma)^2} + \dfrac{2 \eta_\mu M^2}{\varepsilon}\\
&\quad +\dfrac{\max_{\mu\in U}\left\{(\mu_0-\mu)^2 - (\mu_T-\mu)^2\right\}}{2T\eta_\mu} + \dfrac{\eta_\mu M^2}{2}\\
 &\overset{(i)}{\leq} \dfrac{2M}{\varepsilon T} +\dfrac{2\varepsilon (1+C_0) \ell_\theta L_\lambda^2}{(1-\gamma)^2} + \dfrac{2 \eta_\mu M^2}{\varepsilon} + \dfrac{C_0^2}{2T\eta_\mu} +\dfrac{\eta_\mu M^2}{2}\\
 &\overset{(ii)}{\leq} \dfrac{2M+M^2/2}{T^{2/3}} + \dfrac{2(1+C_0)\ell_\theta L_\lambda^2}{(1-\gamma)^2T^{1/3}}+ \dfrac{2M^2+C_0^2/2}{T^{1/3}},
\end{aligned}
\end{equation*}
where in step $(i)$, we apply the inequality 
$\max_{\mu\in U}\left\{(\mu_0-\mu)^2 - (\mu_T-\mu)^2\right\} = \max_{\mu\in U}\big\{(2\mu-\mu_T)\mu_T \big\} \leq C_0^2$. The last step $(ii)$ is from letting $\varepsilon = T^{-1/3}$ and $\eta_\mu = T^{-2/3}$. 
This completes the proof of the constraint violation in \eqref{eq:constraint_violation}.

When $f(\cdot)$ is $\sigma$-strongly concave with respect to $\lambda$, the optimality gap resulted from Lemmas \ref{lemma: avg_L} and \ref{lemma:obtain_opt_const_exact} can be similarly derived as follows
\begin{equation}\label{eq: sum_f_strongly_concave}
\begin{aligned}
    \dfrac{1}{T} \sum_{t=0}^{T-1} f(\lambda(\theta^\star)) \hspace{-2pt}-\hspace{-2pt} f(\lambda(\theta_t)) 
    &\leq \dfrac{L(\lambda(\theta_T),\mu_{T-1})\hspace{-2pt}-\hspace{-2pt} L(\lambda(\theta_{0}), \mu_{0})}{\tilde\varepsilon T} \hspace{-2pt}+\hspace{-2pt} \dfrac{\eta_\mu M^2}{\tilde \varepsilon}\hspace{-1pt} +\hspace{-1pt} \dfrac{\mu_0^2\hspace{-1pt} -\hspace{-1pt} \mu_T^2}{2T\eta_\mu} \hspace{-1pt}+\hspace{-1pt} \dfrac{\eta_\mu M^2}{2}\\
    &\leq \dfrac{(2+C_0)M}{\tilde\varepsilon T} + \left(\dfrac{M^2}{\tilde\varepsilon}+ \dfrac{M^2}{2}\right)\dfrac{1}{\sqrt{T}},
\end{aligned}
\end{equation}
where we specify $\eta_\mu = T^{-1/2}$ and apply the inequality $L(\lambda(\theta_T),\mu_{T-1})-L(\lambda(\theta_{0}), \mu_{0}) = L(\lambda(\theta_T),\mu_{T-1}) - f(\lambda(\theta_{0})) \leq (1+C_0)M + M = (2+C_0)M$.
Finally, we show that the constraint violation satisfies that
\begin{equation*}
\begin{aligned}
    &\quad \dfrac{1}{T}\left[\sum_{t=0}^{T-1}-g(\lambda(\theta_t))\right]_+\\
    &\leq \dfrac{L(\lambda(\theta_T),\mu_{T-1})-L(\lambda(\theta_{0}), \mu_{0})}{\tilde\varepsilon T} + \dfrac{\eta_\mu M^2}{\tilde \varepsilon}+\dfrac{\max_{\mu\in U}\left\{(\mu_0-\mu)^2 - (\mu_T-\mu)^2\right\}}{2T\eta_\mu} + \dfrac{\eta_\mu M^2}{2}\\
    &\leq \dfrac{(2+C_0)M}{\tilde\varepsilon T}+\left(\dfrac{M^2}{\tilde\varepsilon}+ \dfrac{M^2+C_0^2}{2}\right)\dfrac{1}{\sqrt{T}}.
\end{aligned}
\end{equation*}
This completes the proof of Theorem \ref{thm:exact}. 
\end{proof}

\vspace{10pt}
\subsection{Proof of Lemma \ref{lemma:avg_L_sample}}
\begin{proof}
We begin with showing the auxiliary result related to the ascent property of the Lagrangian in Lemma \ref{lemma: ascent}.

\begin{lemma}\label{lemma: ascent}
Suppose that Assumptions \ref{assump:parameterization} and \ref{assump:lipschit_gradient} hold true, and let $\left\{\left(\theta_t, \mu_t\right)\right\}_{t=0}^{T-1}$ be the sequence generated by Algorithm \ref{alg:pdpg}.
Then, for every $0\leq t\leq T-1$, it holds that
\begin{equation*}
\hspace{-2pt} L(\lambda(\theta_{t+1}),\mu_t)\hspace{-2pt}\geq \hspace{-2pt}L(\lambda(\theta_t),\mu_t) + \eta_{\theta}\norm{\nabla_\theta L(\lambda(\theta_t),\mu_t)} - 2\eta_\theta \hspace{-2pt} \norm{e_{L,t}} - \dfrac{(1\hspace{-2pt}+\hspace{-2pt}C_0) \ell_\theta\eta_\theta^2}{2} - 2\eta_\theta(1+C_0)L_{\theta,\lambda} \gamma^H,
\end{equation*}
where $e_{L,t}:= d_{L,t} - \nabla_\theta L(\lambda_H(\theta_t),\mu_t)$.
\end{lemma}
\begin{proof}
By the smoothness of $L(\lambda(\theta), \mu)$ in $\theta$ from Lemma \ref{lemma:boundedness_smoothness}, we have that
\begin{equation}\label{eq: L(lamba)_1}
\begin{aligned}
   \hspace{-5pt} L(\lambda(\theta_{t+1}),\mu_t) 
    \geq& L(\lambda\left(\theta_t), \mu_t\right)+\inner{\nabla_\theta L(\lambda\left(\theta_t), \mu_t\right)}{\theta_{t+1}-\theta_t}-\dfrac{(1+C_0)\ell_\theta}{2}\norm{\theta_{t+1}-\theta_t}^2\\
    \overset{(i)}{=}& L(\lambda\left(\theta_t), \mu_t\right) + \eta_\theta \inner{\nabla_\theta L(\lambda\left(\theta_t), \mu_t\right)}{\dfrac{d_{L,t}}{\norm{d_{L,t}}}} - \dfrac{(1+C_0)\ell_\theta}{2} \eta_\theta^2\\
    = &  L(\lambda\left(\theta_t), \mu_t\right) + \eta_\theta \inner{\nabla_\theta L(\lambda_H\left(\theta_t), \mu_t\right)}{\dfrac{d_{L,t}}{\norm{d_{L,t}}}} - \dfrac{(1+C_0)\ell_\theta}{2} \eta_\theta^2\\
    &+ \eta_\theta\inner{\nabla_\theta L(\lambda(\theta_t), \mu_t) - \nabla_\theta L(\lambda_H(\theta_t), \mu_t)}{\dfrac{d_{L,t}}{\norm{d_{L,t}}}}\\
    \overset{(ii)}{\geq}& L(\lambda(\theta_t), \mu_t)\hspace{-2pt}  +\hspace{-2pt}  \eta_\theta \inner{\nabla_\theta L(\lambda_H(\theta_t), \mu_t)}{\dfrac{d_{L,t}}{\norm{d_{L,t}}}}\hspace{-2pt} -\hspace{-2pt}\dfrac{(1+C_0)\ell_\theta \eta_\theta^2}{2} \hspace{-2pt}- \hspace{-2pt}\eta_\theta (1+C_0)L_{\theta,\lambda} \gamma^H,
\end{aligned}
\end{equation}
where step $(i)$ is due to the update rule of $\theta$ in Algorithm \ref{alg:pdpg}, and step $(ii)$ is due to \eqref{eq: L_smooth} in Lemma \ref{lemma:boundedness_smoothness}. We then bound the second term in the right-hand side of \eqref{eq: L(lamba)_1}:
\begin{equation}\label{eq: L(lamba)_2}
\begin{aligned}
    &\quad \inner{\nabla_\theta L(\lambda_H(\theta_t), \mu_t)}{\dfrac{d_{L,t}}{\norm{d_{L,t}}}} \\
    & = \inner{d_{L, t} - \big[d_{L,t} - \nabla_\theta L(\lambda_H(\theta_t), \mu_t)\big]}{\dfrac{d_{L,t}}{\norm{d_{L,t}}}}\\
    &\geq \norm{d_{L,t}} - \norm{d_{L,t} - \nabla_\theta L(\lambda_H(\theta_t), \mu_t)}\\
    &\overset{(i)}{\geq} \norm{\nabla_\theta L(\lambda_H(\theta_t), \mu_t)} - 2\norm{d_{L,t} - \nabla_\theta L(\lambda_H(\theta_t), \mu_t)}\\
    & \geq \norm{\nabla_\theta L(\lambda(\theta_t), \mu_t)} \hspace{-2pt}-\hspace{-2pt} \norm{\nabla_\theta L(\lambda(\theta_t), \mu_t) \hspace{-2pt}-\hspace{-2pt}  \nabla_\theta L(\lambda_H(\theta_t), \mu_t)} \hspace{-1pt} -\hspace{-1pt} 2\norm{d_{L,t} \hspace{-1pt} -\hspace{-1pt}  \nabla_\theta L(\lambda_H(\theta_t), \mu_t)}\\
    &\overset{(ii)}{\geq} \norm{\nabla_\theta L(\lambda(\theta_t), \mu_t)} - (1+C_0)L_{\theta,\lambda} \gamma^H - 2\norm{d_{L,t} - \nabla_\theta L(\lambda_H(\theta_t), \mu_t)},
\end{aligned}
\end{equation}
where step $(i)$ applies the triangular inequality to the term $d_{L,t} = \nabla_\theta L(\lambda_H(\theta_t), \mu_t) + \big[d_{L,t}- \nabla_\theta L(\lambda_H(\theta_t), \mu_t)\big]$, and step $(ii)$ is again due to \eqref{eq: L_smooth} in Lemma \ref{lemma:boundedness_smoothness}. 
The proof of Lemma \ref{lemma: ascent} is completed by substituting \eqref{eq: L(lamba)_2} back into \eqref{eq: L(lamba)_1}. 
\end{proof} 

Now, we finalize the proof of Lemma \ref{lemma:avg_L_sample}. Recall the definition of $\theta_{\varepsilon,t} := \lambda^{-1}_{\mc{V}_{\lambda(\theta_t)}}\big((1-\varepsilon)\lambda(\theta_t)+\varepsilon \lambda(\theta^\star)\big)$ from \eqref{eq:theta_epsilon}, which is always well-defined as long as $\epsilon \leq \bar\epsilon$ (see Assumption \ref{assump:parameterization}).
Using the fact that $L(\lambda(\theta), \mu)$ is $(1+C_0)\ell_\theta$-smooth with respect to $\theta$ from Lemma \ref{lemma:boundedness_smoothness}, we have that 
\begin{equation}\label{eq: L_lam_sample}
\begin{aligned}
    L(\lambda(\theta_t),\mu_t)
    &\geq L(\lambda(\theta_{\varepsilon,t}), \mu_t) - \inner{\nabla_\theta L(\lambda(\theta_t), \mu_t)}{\theta_{\varepsilon,t}-\theta_t} -\dfrac{(1+C_0)\ell_\theta}{2}\norm{\theta_{\varepsilon,t}-\theta_t}^2\\
    &\geq (1-\varepsilon) L(\lambda(\theta_t), \mu_t) + \varepsilon L(\lambda(\theta^\star),\mu_t) - \inner{\nabla_\theta L(\lambda(\theta_t), \mu_t)}{\theta_{\varepsilon,t} - \theta_t}\\
    &\quad- \dfrac{(1+C_0)\ell_\theta}{2} \norm{\theta_{\varepsilon,t} - \theta_t}^2,
\end{aligned}
\end{equation}
where the last line is due to the concavity of $L(\lambda(\theta), \mu)$ in $\lambda$.
By the local inverse assumption and $L_\lambda$-Lipschitz continuity of $\lambda^{-1}_{\mc{V}_{\lambda(\theta)}}(\cdot)$ for all $\theta \in \mbb{R}^K$ (see Assumption \ref{assump:parameterization}), it holds that
\begin{equation}\label{eq: theta_esp}
\begin{aligned}
    \norm{\theta_{\varepsilon,t} - \theta_t} &= \norm{\lambda^{-1}_{\mc{V}_{\lambda(\theta_t)}}\big((1-\varepsilon)\lambda(\theta_t)+\varepsilon \lambda(\theta^\star)\big) - \lambda^{-1}_{\mc{V}_{\lambda(\theta_t)}}\big(\lambda(\theta_t)\big)}\\
    & \leq L_\lambda \cdot \varepsilon \norm{\lambda(\theta^\star) - \lambda(\theta_t)} \leq \dfrac{\sqrt{2}\varepsilon L_{\lambda}}{1-\gamma}.
\end{aligned}
\end{equation}
Then, we apply the ascent property in Lemma \ref{lemma: ascent} and substitute \eqref{eq: theta_esp} into \eqref{eq: L_lam_sample} to obtain the following
\begin{equation}\label{eq:L_t+1>=}
\begin{aligned}
     &L(\lambda(\theta_{t+1}), \mu_t) 
     \\[3pt]
     \geq& L(\lambda(\theta_t),\mu_t) \hspace{-1pt}+\hspace{-1pt} \eta_{\theta}\norm{\nabla_\theta L(\lambda(\theta_t),\mu_t)} - 2\eta_\theta \norm{e_{L,t}} - \dfrac{(1+C_0) \ell_\theta\eta_\theta^2}{2} - 2\eta_\theta(1+C_0)L_{\theta,\lambda} \gamma^H\\[3pt]
     \geq& (1-\varepsilon) L(\lambda(\theta_t), \mu_t) + \varepsilon L(\lambda(\theta^\star),\mu_t) 
     + \eta_{\theta}\norm{\nabla_\theta L(\lambda(\theta_t),\mu_t)} - 2\eta_\theta \norm{e_{L,t}}\\[3pt]
     &- \dfrac{(1+C_0) \ell_\theta\eta_\theta^2}{2} - 2\eta_\theta(1+C_0)L_{\theta,\lambda} \gamma^H - \inner{\nabla_\theta L(\lambda(\theta_t), \mu_t)}{\theta_{\varepsilon,t} - \theta_t} - \dfrac{(1+C_0)\ell_\theta L_\lambda^2 \varepsilon^2}{(1-\gamma)^2}\\[3pt]
     \overset{(i)}{\geq} &(1-\varepsilon) L(\lambda(\theta_t), \mu_t) + \varepsilon L(\lambda(\theta^\star),\mu_t) 
     + \left(\eta_{\theta} - \dfrac{\sqrt{2}\varepsilon L_\lambda}{1-\gamma}\right) \norm{\nabla_\theta L(\lambda(\theta_t), \mu_t)}\\[3pt]
     &- 2 \eta_\theta\norm{e_{L,t}} - (1+C_0) \left(\dfrac{\ell_\theta \eta_\theta^2}{2} + 2\eta_\theta L_{\theta,\lambda} \gamma^H + \dfrac{\ell_\theta L_\lambda^2 \varepsilon^2}{(1-\gamma)^2}\right),
\end{aligned}
\end{equation}
where in step $(i)$ we use \eqref{eq: theta_esp} again to lower-bound the term $\inner{\nabla_\theta L(\lambda(\theta_t), \mu_t)}{\theta_{\varepsilon,t} - \theta_t}$.
By subtracting $L(\lambda(\theta^\star),\mu_t)$ from both sides of \eqref{eq:L_t+1>=} and rearranging the terms, we derive that
\begin{equation}\label{eq:L_t+1>=_1}
\begin{aligned}
    &L(\lambda(\theta^\star),\mu_t) - L(\lambda(\theta_{t+1}), \mu_t)\\[5pt]
    \leq& (1-\varepsilon) \big[L(\lambda(\theta^\star),\mu_t) - L(\lambda(\theta_{t}), \mu_t)\big] - \left(\eta_{\theta} - \dfrac{\sqrt{2}\varepsilon L_\lambda}{1-\gamma}\right) \norm{\nabla_\theta L(\lambda(\theta_t), \mu_t)}\\[5pt]
    & +2 \eta_\theta\norm{e_{L,t}} +(1+C_0) \left(\dfrac{\ell_\theta \eta_\theta^2}{2} + 2\eta_\theta L_{\theta,\lambda} \gamma^H + \dfrac{\ell_\theta L_\lambda^2 \varepsilon^2}{(1-\gamma)^2}\right).
\end{aligned}
\end{equation}
Note that by the dual variable update, we have that 
$$\left|L(\lambda(\theta), \mu_{t+1}) - L(\lambda(\theta),\mu_t)\right| = \left|(\mu_{t+1} - \mu_t)\cdot g(\lambda(\theta))\right|\leq \eta_\mu M^2, \quad \forall \theta \in \mbb{R}^K.$$ 
Together with \eqref{eq:L_t+1>=_1}, we deduce that
\begin{equation*}
\begin{aligned}
    &L(\lambda(\theta^\star),\mu_{t+1}) - L(\lambda(\theta_{t+1}), \mu_{t+1})\\[5pt]
   =& L(\lambda(\theta^\star),\mu_{t+1}) -L(\lambda(\theta^\star),\mu_t) + L(\lambda(\theta^\star),\mu_t)- L(\lambda(\theta_{t+1}),\mu_{t})\\[5pt]
   &+L(\lambda(\theta_{t+1}),\mu_{t}) - L(\lambda(\theta_{t+1}),\mu_{t+1})\\[5pt]
    \leq& (1-\varepsilon) \big[L(\lambda(\theta^\star),\mu_t) - L(\lambda(\theta_{t}), \mu_t)\big] - \left(\eta_{\theta} - \dfrac{\sqrt{2}\varepsilon L_\lambda}{1-\gamma}\right) \big\|\nabla_\theta L(\lambda(\theta_t), \mu_t)\big\|+2 \eta_\theta\norm{e_{L,t}}\\[5pt]
    &  +(1+C_0) \left(\dfrac{\ell_\theta \eta_\theta^2}{2} + 2\eta_\theta L_{\theta,\lambda} \gamma^H + \dfrac{\ell_\theta L_\lambda^2 \varepsilon^2}{(1-\gamma)^2}\right) + 2\eta_\mu M^2.
\end{aligned}
\end{equation*}
When we choose $\eta_\theta - \frac{\sqrt{2} \varepsilon L_\lambda}{(1-\gamma)} \geq 0$, i.e., $\varepsilon\leq \frac{(1-\gamma)\eta_\theta}{\sqrt{2}L_\lambda}$, then it holds that 
\begin{equation*}
\begin{aligned}
    L(\lambda(\theta^\star),\mu_{t+1}) - L(\lambda(\theta_{t+1}),\mu_{t+1}) \leq& (1-\varepsilon) \big[L(\lambda(\theta^\star),\mu_t) - L(\lambda(\theta_{t}), \mu_t)\big] + 2 \eta_\theta\norm{e_{L,t}}\\[5pt]
    &+(1+C_0) \left(\dfrac{\ell_\theta \eta_\theta^2}{2} + 2\eta_\theta L_{\theta,\lambda} \gamma^H + \dfrac{\ell_\theta L_\lambda^2 \varepsilon^2}{(1-\gamma)^2}\right) + 2\eta_\mu M^2.
\end{aligned}
\end{equation*}
We recursively apply the above inequality to obtain that
\begin{equation}\label{eq:L-L_bound_sample}
\begin{aligned}
    &L(\lambda(\theta^\star),\mu_{t}) - L(\lambda(\theta_{t}),\mu_{t})\\
    \leq& (1-\varepsilon)^t \big[L(\lambda(\theta^\star), \mu_0) - L(\lambda(\theta_0),\mu_0) \big]
     + \sum_{i=0}^{t-1} (1-\varepsilon)^i \cdot 2\eta_\theta\norm{e_{L,t-1-i}}\\
     &+ \sum_{i=0}^{t-1} (1-\varepsilon)^i \left[(1+C_0)\left(\dfrac{\ell_\theta \eta_\theta^2}{2} + 2\eta_\theta L_{\theta,\lambda} \gamma^H + \dfrac{\ell_\theta L_\lambda^2 \varepsilon^2}{(1-\gamma)^2}\right) + 2\eta_\mu M^2 \right]\\
     \leq&(1-\varepsilon)^t \big[L(\lambda(\theta^\star), \mu_0) - L(\lambda(\theta_0),\mu_0) \big]
     + \sum_{i=0}^{t-1} (1-\varepsilon)^i \cdot 2\eta_\theta\norm{e_{L,t-1-i}}\\
     &+ \dfrac{1}{\varepsilon} \left[(1+C_0)\left(\dfrac{\ell_\theta \eta_\theta^2}{2} + 2\eta_\theta L_{\theta,\lambda} \gamma^H
    + \dfrac{\ell_\theta L_\lambda^2 \varepsilon^2}{(1-\gamma)^2}\right) + 2\eta_\mu M^2 \right],
\end{aligned}
\end{equation}
where we use the bound $\sum_{i=0}^{t-1} (1-\varepsilon)^i  = \frac{1 - (1-\varepsilon)^t}{\varepsilon}\leq \frac{1}{\varepsilon}$ in the last line.
Summing the inequality \eqref{eq:L-L_bound_sample} over $t = 0,\dots, T-1$ yields that
\begin{equation}\label{eq:sum_L}
\begin{aligned}
    &\sum_{t=0}^{T-1} L(\lambda(\theta^\star),\mu_{t}) - L(\lambda(\theta_{t}),\mu_{t})\\
    \leq& \dfrac{1}{\varepsilon} \big[L(\lambda(\theta^\star), \mu_0) - L(\lambda(\theta_0),\mu_0) \big] +
    \dfrac{2\eta_\theta}{\varepsilon}\sum_{t=0}^{T-1}\norm{e_{L,t}}\\
    &+ \dfrac{T}{\varepsilon} \left[(1+C_0)\left(\dfrac{\ell_\theta \eta_\theta^2}{2} + 2\eta_\theta L_{\theta,\lambda} \gamma^H
    + \dfrac{\ell_\theta L_\lambda^2 \varepsilon^2}{(1-\gamma)^2}\right) + 2\eta_\mu M^2 \right], 
\end{aligned}
\end{equation}
where the second term $\frac{2\eta_\theta}{\varepsilon}\sum_{t=0}^{T-1}\norm{e_{L,t}}$ is due to 
\begin{equation}
    \sum_{t=0}^{T-1}\sum_{i=0}^{t-1} (1-\varepsilon)^i \cdot 2\eta_\theta\norm{e_{L, t-1-i}} = 2\eta_\theta \sum_{t=0}^{T-1} \left(\sum_{i=0}^{T-1-t} (1-\varepsilon)^i \right) \norm{e_{L,t}} \leq \frac{2\eta_\theta}{\varepsilon}\sum_{t=0}^{T-1}\norm{e_{L,t}}.
\end{equation}
We comment that it is possible to derive a tighter bound in \eqref{eq:sum_L} from \eqref{eq:L-L_bound_sample} by handling the summation $\sum_{i=0}^{t-1} (1-\varepsilon)^i \cdot 2\eta_\theta\norm{e_{L,t-1-i}}$ more carefully.
Nevertheless, the final order of the sample complexity, as well as its dependencies on the problem-related parameters, would not be changed.
Finally, the proof of Lemma \ref{lemma:avg_L_sample} is completed by dividing both sides of \eqref{eq:sum_L} by $T$.
\end{proof}

\subsection{Proof of Lemma \ref{lemma:obtain_opt_const_sample}}\label{app:obtain_opt_const_sample}
\begin{proof}
By definition of the Lagrangian function $L(\lambda(\theta),\mu)$, we have 
\begin{equation}\label{eq: f^*-f}
\begin{aligned}
    \dfrac{1}{T} \sum_{t=0}^{T-1} f(\lambda(\theta^\star))-f(\lambda(\theta_{t}))
    &= \dfrac{1}{T} \sum_{t=0}^{T-1}\big[f(\lambda(\theta^\star))-L(\lambda(\theta_t), \mu_t)\big]+\dfrac{1}{T} \sum_{t=0}^{T-1} \mu_t g(\lambda(\theta_t))\\
    &{\overset{(i)}{=}} \dfrac{1}{T} \sum_{t=0}^{T-1}\big[L(\lambda(\theta^\star), \mu^\star) - L(\lambda(\theta_t),\mu_t)\big] + \dfrac{1}{T} \sum_{t=0}^{T-1} \mu_t g(\lambda(\theta_t))\\
    &{\overset{(ii)}{\leq}} \dfrac{1}{T} \sum_{t=0}^{T-1} \big[ L(\lambda(\theta^\star), \mu_t) - L(\lambda(\theta_t),\mu_t) \big] + \dfrac{1}{T} \sum_{t=0}^{T-1} \mu_t g(\lambda(\theta_t))\\
    &\leq C + \dfrac{1}{T} \sum_{t=0}^{T-1} \mu_t g(\lambda(\theta_t)),
\end{aligned}
\end{equation}
where the equality in step $(i)$ holds due to  complementary slackness that $\mu^\star g(\lambda(\theta^\star))=0$ (see Lemma \ref{lemma:duality}), and step $(ii)$ is due to the fact that $\mu^\star = \operatorname{argmin}_{\mu\geq 0}L(\lambda(\theta^\star), \mu)$.

Then, we upper-bound the second term in the last line of \eqref{eq: f^*-f}. By the update rule of $\mu_t$ in Algorithm \ref{alg:pdpg} and the fact that $\nabla_\mu L(\lambda(\theta_t),\mu_t) = g(\lambda_t)$, we obtain that for any $\mu \in U$
\begin{equation}\label{eq: u-u}
\begin{aligned}
    \left(\mu_{t+1}-\mu\right)^2 
    &=\big[\mc{P}_u\big(\mu_t-\eta_\mu g(\lambda_t)\big)-\mu\big]^2\\
    &\ {\overset{(i)}{\leq}} (\mu_t - \eta_{\mu} g(\lambda_t) - \mu)^2\\
    &= (\mu_t - \mu)^2 - 2\eta_{\mu} (\mu_t-\mu) \cdot g(\lambda_t) + (\eta_\mu g(\lambda_t))^2,
\end{aligned}
\end{equation}
where step $(i)$ follows from the non-expansive property of the projection operator.
Then, using the elementary inequality $(x+y)^2 \leq 2x^2 + 2y^2$ and Lipschitz continuity of $g(\cdot)$, we have that
\begin{equation}
[g(\lambda_t)]^2 \leq 2[g(\lambda(\theta_t))]^2 + 2[g(\lambda(\theta_t)) - g(\lambda_t)]^2 \leq 2M^2 + 2M_\lambda^2 \norm{\lambda(\theta_t)-\lambda_t}^2,
\end{equation}
where we also use the boundedness of $g(\cdot)$ on $\Lambda$ in the last inequality.
We substitute the above inequality into \eqref{eq: u-u} to obtain that for any $\mu \in U$
\begin{equation}\label{eq: u-u2}
\left(\mu_{t+1}-\mu\right)^2 \leq (\mu_t - \mu)^2 - 2\eta_{\mu} (\mu_t-\mu) \cdot g(\lambda_t) + 2\eta_\mu^2 M^2 + 2M_\lambda^2 \norm{\lambda(\theta_t)-\lambda_t}^2\eta_\mu^2.
\end{equation}
Then, by setting $\mu = 0$ and rearranging the terms, we have that 
\begin{equation}\label{eq: mu_tg(lambda_t)}
\begin{aligned}
    \mu_t g(\lambda_t) \leq \dfrac{1}{2 \eta_{\mu}} \left(\mu_t^2 - \mu_{t+1}^2+2 \eta_\mu^2 M^2 + 2M_\lambda^2 \norm{\lambda(\theta_t)-\lambda_t}^2\eta_\mu^2\right).
\end{aligned}
\end{equation}
Finally, using the above inequality and the Lipschitz continuity of $g(\cdot)$, we can bound the term $\mu_tg(\lambda(\theta_t))$ as follows
\begin{equation}\label{eq: uG_1}
\begin{aligned}
    \mu_tg(\lambda(\theta_t)) &= \mu_t g(\lambda_t) + \mu_t\big(g(\lambda(\theta_t)) - g(\lambda_t)\big)\\
    &\leq \hspace{-2pt} \dfrac{1}{2 \eta_{\mu}} \hspace{-2pt}\left(\mu_t^2 \hspace{-2pt}-\hspace{-2pt} \mu_{t+1}^2 \hspace{-2pt}+\hspace{-2pt} 2\eta_\mu^2M^2 \hspace{-2pt}+\hspace{-2pt} 2M_\lambda^2 \norm{\lambda(\theta_t) \hspace{-2pt}-\hspace{-2pt}\lambda_t}^2\eta_\mu^2\right) \hspace{-2pt}+\hspace{-2pt} C_0 M_\lambda \norm{\lambda(\theta_t) \hspace{-2pt}-\hspace{-2pt} \lambda_t}\\
    &{\overset{(i)}{\leq}} \dfrac{1}{2 \eta_{\mu}} \hspace{-2pt}\left(\mu_t^2 \hspace{-2pt}-\hspace{-2pt} \mu_{t+1}^2\right) \hspace{-2pt}+\hspace{-2pt} \eta_\mu M^2 \hspace{-2pt}+\hspace{-2pt} M_\lambda^2 \eta_\mu \hspace{-2pt}\left(2\norm{\lambda(\theta_t) \hspace{-2pt}-\hspace{-2pt} \lambda_H(\theta_t)}^2 \hspace{-2pt}+\hspace{-2pt} 2\norm{\lambda_H(\theta_t) \hspace{-2pt}-\hspace{-2pt} \lambda_t}^2\right) \\
    &\quad + C_0 M_\lambda \left(\norm{\lambda(\theta_t) - \lambda_H(\theta_t)} + \norm{\lambda_H(\theta_t) - \lambda_t}\right),
\end{aligned}
\end{equation}
where $(i)$ is again due to the inequality $(x+y)^2 \leq 2x^2 + 2y^2$.
We sum both sides of \eqref{eq: uG_1} from $t=0$ to $T-1$ and adopt the notation $e_{\lambda,t}:=\lambda_H(\theta_t) - \lambda_t$ to obtain that 
\begin{equation}\label{eq: u_tg}
\begin{aligned}
    \dfrac{1}{T} \sum_{t=0}^{T-1} \mu_t g(\lambda(\theta_t)) 
    &\leq \dfrac{\mu_0^2 - \mu_T^2}{2T\eta_\mu}  +  \eta_\mu M^2 + \dfrac{2\eta_\mu M_\lambda^2}{T} \sum_{t=0}^{T-1}\left( \norm{\lambda(\theta_t) - \lambda_H(\theta_t)}^2 + \norm{e_{\lambda,t}}^2\right) \\
    &\quad + \dfrac{C_0 M_\lambda}{T} \sum_{t=0}^{T-1}\left(\norm{\lambda(\theta_t) - \lambda_H(\theta_t)} + \norm{e_{\lambda,t}}\right)\\
     &{\overset{(i)}{\leq}} \dfrac{\mu_0^2-\mu_T^2}{2T\eta_\mu}+ \eta_\mu M^2 + C_0M_\lambda \gamma^H  + 2\eta_\mu M_\lambda^2 \gamma^{2H}+ \dfrac{C_0M_\lambda}{T}\sum_{t=0}^{T-1} \norm{e_{\lambda,t}} \\
     &\quad + \dfrac{2\eta_\mu M_\lambda^2}{T}\sum_{t=0}^{T-1} \norm{e_{\lambda,t}}^2,
\end{aligned}
\end{equation}
where we further apply the upper bound $\norm{\lambda(\theta_t) - \lambda_H(\theta_t)}\leq \gamma^H$ in $(i)$.
It is clear that the optimality gap in \eqref{eq:optimality_gap_lemma} can be obtained by substituting the upper bound in \eqref{eq: u_tg} into \eqref{eq: f^*-f}.

\vspace{5pt}
Now, we proceed to the constraint violation. 
If $\sum_{t=0}^{T-1} g(\lambda(\theta_t)) \geq 0$, then there is no constraint violation and the bound in \eqref{eq:constraint_violation_lemma} is trivially satisfied. 
Therefore, from now on, we assume $\left[\sum_{t=0}^{T-1} -g(\lambda(\theta_t))\right]_+ > 0$, which implies that $\sum_{t=0}^{T-1}-g(\lambda(\theta_t)) = \left[\sum_{t=0}^{T-1}-g(\lambda(\theta_t))\right]_+$.
Define $\widehat{\mu}:= \mu^\star + 1\geq 1$, as  $\mu^\star \geq 0$. By the boundedness of $\mu^\star$ (see Lemma \ref{lemma:duality}), we have that
\begin{equation*}
    \widehat{\mu} = \mu^\star + 1 \leq \dfrac{f(\lambda(\theta^\star))-f(\lambda(\widetilde\theta))}{\xi} +1 \leq \dfrac{M-f(\lambda(\widetilde\theta))}{\xi} +1=C_0,
\end{equation*}
which implies that $\widehat{\mu} \in U$. Thus, it follows that
\begin{equation}\label{eq: max_term}
\begin{aligned}
    \dfrac{1}{T}\left[\sum_{t=0}^{T-1}-g(\lambda(\theta_t))\right]_+ &= (\widehat{\mu} - \mu^\star) \cdot \dfrac{1}{T}\sum_{t=0}^{T-1} -g(\lambda(\theta_t))\\
    & \leq \max_{\mu \in U} \left\{ (\mu^\star - \mu)\cdot \dfrac{1}{T}\sum_{t=0}^{T-1} g(\lambda(\theta_t)) \right\}
\end{aligned}
\end{equation}
To upper-bound the last line in \eqref{eq: max_term}, we note that
\begin{equation}\label{eq: (mu_star-mu_)grad_L}
\begin{aligned}
    (\mu^\star - \mu) \cdot g(\lambda(\theta_t)) &= (\mu^\star - \mu_t ) \cdot g(\lambda(\theta_t))+ (\mu_t -\mu) \cdot g(\lambda(\theta_t))\\
    &= (\mu^\star - \mu_t)\cdot \nabla_\mu L(\lambda(\theta_t), \mu_t) + (\mu_t -\mu) \cdot g(\lambda(\theta_t))\\
    &{\overset{(i)}{=}} L(\lambda(\theta_t), \mu^\star) - L(\lambda(\theta_t), \mu_t) + (\mu_t - \mu) \cdot g(\lambda(\theta_t))\\
    &{\overset{(ii)}{\leq}} L(\lambda(\theta^\star), \mu^\star) - L(\lambda(\theta_t), \mu_t) + (\mu_t -\mu) \cdot g(\lambda(\theta_t))\\
    &{\overset{(iii)}{\leq}} L(\lambda(\theta^\star), \mu_t) - L(\lambda(\theta_t), \mu_t) + (\mu_t -\mu) \cdot g(\lambda(\theta_t)),
\end{aligned}
\end{equation}
where we use the linearity of $L(\lambda(\theta), \mu)$ with respect to $\mu$ in step $(i)$. 
Step $(ii)$ follows from the fact that $\theta^\star$ maximizes $L(\lambda(\cdot), \mu^\star)$, and step $(iii)$ is due to the fact that $\mu^\star$ minimize $L(\lambda(\theta^\star), \cdot)$ from \eqref{eq:saddle}.
Then, by \eqref{eq: u-u2} and \eqref{eq: uG_1}, we can control the last term in step $(iii)$ of \eqref{eq: (mu_star-mu_)grad_L} as 
\begin{equation}\label{eq: u_t-u}
\begin{aligned}
    &(\mu_t - \mu) \cdot g(\lambda(\theta_t))\\
   =& (\mu_t - \mu) \cdot g(\lambda_t) + (\mu_t - \mu) \cdot \big[g(\lambda(\theta_t)) - g(\lambda_t) \big]\\
   \leq& \dfrac{1}{2\eta_\mu} \hspace{-2pt} \left[ (\mu_t \hspace{-2pt}-\hspace{-2pt} \mu)^2 \hspace{-2pt}-\hspace{-2pt} (\mu_{t+1} \hspace{-2pt}-\hspace{-2pt} \mu)^2 \hspace{-2pt}+\hspace{-2pt} 2 \eta_\mu^2 M^2 \hspace{-2pt}+\hspace{-2pt} 2M_\lambda^2 \norm{\lambda(\theta_t) \hspace{-2pt}-\hspace{-2pt}\lambda_t}^2\eta_\mu^2\right] \hspace{-2pt}+\hspace{-2pt} C_0 M_\lambda \norm{\lambda(\theta_t)\hspace{-2pt}-\hspace{-2pt}\lambda_t}\\
   \leq& \dfrac{1}{2\eta_\mu} \hspace{-2pt}\left[ (\mu_t \hspace{-2pt}- \hspace{-2pt}\mu)^2 \hspace{-2pt}- \hspace{-2pt}(\mu_{t+1}\hspace{-2pt} -\hspace{-2pt} \mu)^2\right]\hspace{-2pt}+\hspace{-2pt} \eta_\mu M^2 \hspace{-2pt}+\hspace{-2pt} 2\eta_\mu M_\lambda^2 
 \left(\gamma^{2H} \hspace{-2pt}+\hspace{-2pt} \norm{e_{\lambda, t}}^2\right) \hspace{-2pt}+\hspace{-2pt} C_0 M_\lambda \left(\gamma^H \hspace{-2pt}+\hspace{-2pt} \norm{e_{\lambda, t}}\right),
\end{aligned}
\end{equation}
where the last inequality is due to similar bounds on $\norm{\lambda(\theta_t)-\lambda_t}$ and $\norm{\lambda(\theta_t)-\lambda_t}^2$ as \eqref{eq: uG_1} and \eqref{eq: u_tg}.
Combining \eqref{eq: (mu_star-mu_)grad_L} and \eqref{eq: u_t-u}, we derive the constraint violation as follows
\begin{equation}
\begin{aligned}
    &\dfrac{1}{T}\left[\sum_{t=0}^{T-1}-g(\lambda(\theta_t))\right]_+ \\
    \leq& \dfrac{1}{T} \sum_{t=0}^{T-1}\big[L(\lambda(\theta^\star), \mu_t)-L(\lambda(\theta_t), \mu_t)\big] + \eta_\mu M^2 + C_0M_\lambda \gamma^H+ 2\eta_\mu M_\lambda^2 \gamma^{2H}\\ &+ \dfrac{C_0M_\lambda}{T}\sum_{t=0}^{T-1} \norm{e_{\lambda,t}}
    + \dfrac{2\eta_\mu M_\lambda^2}{T}\sum_{t=0}^{T-1} \norm{e_{\lambda,t}}^2+ \dfrac{\max_{\mu\in U}\left\{(\mu_0-\mu)^2 - (\mu_T-\mu)^2\right\}}{2T\eta_\mu},
\end{aligned}
\end{equation}
which directly implies \eqref{eq:constraint_violation_lemma}.
\end{proof}

\vspace{5pt}
\subsection{Proof of Theorem \ref{thm:sample} (Sample-based Setting)}
\begin{proof}
The proof of Theorem \ref{thm:sample} primarily utilizes
Lemma \ref{lemma:avg_L_sample}, Lemma \ref{lemma:obtain_opt_const_sample},  and the boundedness of approximation errors $\norm{e_{L,t}}$ and $\norm{e_{\lambda, t}}$ (see Appendix \ref{subsec:control_e}).
We choose the following set of parameters: $\eta_\theta = T^{-1/2}$, $\eta_\mu = T^{-3/4}$, $\alpha_t =(t+1)^{-1/2}$, $H = \frac{1}{2}\log_{1/\gamma}T$ (i.e., $\gamma^H = T^{-1/2}$), $\mu_0 = 0$, and $\varepsilon = \frac{(1-\gamma)\eta_\theta}{\sqrt{2}L_\lambda} = \frac{1-\gamma}{\sqrt{2}L_\lambda}T^{-1/2}$.
We note that for $T \geq \frac{1}{2}\left(\frac{1-\gamma}{L_\lambda \bar{\varepsilon}}\right)^2$, the choice of $\varepsilon$ ensures that $\varepsilon \leq \bar \varepsilon$, and therefore it is consistent with Assumption \ref{assump:parameterization}.
Besides, by viewing the step-sizes as variables, one can also verify that the above parameter choices obtain the optimal convergence rate for both the optimality gap and constraint violation.
Under this set of parameter, we have that $\mbb{E}\left[ \norm{e_{\lambda,t}}^2\right]\lesssim \frac{C_\lambda}{\sqrt{t+2}}$ from \eqref{eq:bound_e_lambda_0}. Then, summing it over $t=0,\dots, T-1$ yields that
\begin{equation}\label{eq:sum_E[e_lam_t^2]}
    \sum_{t=0}^{T-1} \mbb{E}\left[ \norm{e_{\lambda,t}}^2\right]\lesssim C_\lambda \int_{0}^T \dfrac{1}{\sqrt{t+2}}dt \lesssim C_\lambda T^{1/2}.
\end{equation}
Furthermore, since $\mbb{E}\left[ \norm{e_{\lambda,t}}\right] \leq \sqrt{\mbb{E}\left[ \norm{e_{\lambda,t}}^2\right]}$, it holds that $\mbb{E}\left[ \norm{e_{\lambda,t}}\right]\lesssim \frac{\sqrt{C_\lambda}}{{(t+2)}^{1/4}}$. Hence, summing it over $t=0,\dots, T-1$ yields that
\begin{equation}\label{eq:sum_E[e_lam_t]}
\sum_{t=0}^{T-1} \mbb{E}\left[ \norm{e_{\lambda,t}}\right]\lesssim \sqrt{C_\lambda} \int_{0}^T \dfrac{1}{(t+2)^{1/4}}dt \lesssim \sqrt{C_\lambda} T^{3/4}.
\end{equation}
Then, from \eqref{eq:E[e_lt]} in Lemma \ref{lemma:E_elt_bound}, we have that $\mbb{E}\left[\norm{e_{L,t}}\right] \lesssim \frac{C_e (1+C_0)}{(t+1)^{1/4}}$. Summing the term over $t=0, \dots, T-1$ results in that
\begin{equation}\label{eq:sum_E[e_lt]}
\sum_{t=0}^{T-1} \mbb{E}\left[\norm{e_{L,t}}\right] \lesssim C_e (1+C_0)\int_{0}^T \dfrac{1}{(t+1)^{1/4}}dt \lesssim C_e (1+C_0)T^{3/4}.
\end{equation}

Therefore, by \eqref{eq:avg_L_sample} in Lemma \ref{lemma:avg_L_sample} and \eqref{eq:optimality_gap_lemma} in Lemma \ref{lemma:obtain_opt_const_sample}, the optimality gap in the sample-based setting can be upper-bounded as follows
\begin{equation*}\label{eq: optimality_gap_lemma_1}
\begin{aligned}
&\mbb{E}\left[\dfrac{1}{T}\sum_{t=0}^{T-1} f(\lambda(\theta^\star)) - f(\lambda(\theta_t))\right]\\
\leq& \dfrac{L(\lambda(\theta^\star),\mu_{0})-L(\lambda(\theta_{0}), \mu_{0})}{\varepsilon T} + \dfrac{2\eta_\theta }{\varepsilon T}\sum_{t=0}^{T-1}\mbb{E}\left[\norm{e_{L,t}}\right]
\\
& + \dfrac{1}{\varepsilon}\left[(1+C_0)\left(\dfrac{ \ell_\theta \eta_\theta^2}{2} + 2L_{\theta,\lambda} \eta_\theta \gamma^H\right) + 2\eta_\mu M^2 \right]+ \dfrac{(1+C_0) \ell_\theta L_\lambda^2\varepsilon}{(1-\gamma)^2}+\dfrac{\mu_0^2-\mu_T^2}{2T\eta_\mu} + \eta_\mu M^2\\
&+ C_0M_\lambda \gamma^H  + 2\eta_\mu M_\lambda^2 \gamma^{2H}+ \dfrac{C_0M_\lambda}{T}\sum_{t=0}^{T-1} \mbb{E}\left[\norm{e_{\lambda,t}}\right]+ \dfrac{2\eta_\mu M_\lambda^2}{T}\sum_{t=0}^{T-1} \mbb{E}\left[\norm{e_{\lambda,t}}^2\right]\\[5pt]
\overset{(i)}{\lesssim}& \dfrac{L_\lambda M}{(1-\gamma)T^{1/2}}  +\dfrac{L_\lambda C_e (1+C_0)}{(1-\gamma)T^{1/4}} +(1+C_0)\left(\dfrac{\ell_\theta L_\lambda}{(1-\gamma)T^{1/2}} + \dfrac{L_{\theta,\lambda} L_\lambda \gamma^H}{1-\gamma}\right) + \dfrac{L_\lambda M^2}{(1-\gamma)T^{1/4}}\\
&+\dfrac{(1+C_0)\ell_\theta L_\lambda}{(1-\gamma)T^{1/2}}+ \dfrac{M^2}{T^{3/4}} + C_0M_\lambda \gamma^H + \dfrac{M_\lambda^2 \gamma^{2H}}{T^{3/4}}+\dfrac{C_0M_\lambda\sqrt{C_\lambda}}{T^{1/4}} + \dfrac{M_\lambda^2 C_\lambda}{T^{5/4}}\\[5pt]
\overset{(ii)}{\lesssim}& \left[\dfrac{C_0L_\lambda C_e + L_\lambda M^2}{1-\gamma} + C_0 M_\lambda \sqrt{C_\lambda}\right] \cdot \dfrac{1}{T^{1/4}} 
= \mc{O}\left(\dfrac{T^{-1/4}}{(1-\gamma)^4}\right).
\end{aligned}
\end{equation*}
For the first term in step $(i)$, since $\mu_0 = 0$, $\varepsilon = \frac{(1-\gamma)\eta_\theta}{\sqrt{2}L_\lambda} = \frac{1-\gamma}{\sqrt{2}L_\lambda}T^{-1/2}$, and $|f(\lambda)| \leq M$ for all $\lambda \in \Lambda$ from Lemma \ref{lemma:boundedness_smoothness}, we have that 
\begin{equation}
    \dfrac{L(\lambda(\theta^\star),\mu_{0})-L(\lambda(\theta_{0}), \mu_{0})}{\varepsilon T} = \dfrac{f(\lambda(\theta^\star))-f(\lambda(\theta_{0}))}{\varepsilon T} \lesssim \dfrac{L_\lambda M}{(1-\gamma)T^{1/2}}.
\end{equation}
The second term in step $(i)$  is obtained from \eqref{eq:sum_E[e_lt]} under the choice of $\eta_\theta = T^{-1/2}$, i.e.,
\begin{equation}
   \dfrac{2\eta_\theta }{\varepsilon T}\sum_{t=0}^{T-1}\norm{e_{L,t}} \lesssim \dfrac{\sqrt{2}L_\lambda}{(1-\gamma)T}\cdot C_e(1+C_0)T ^{3/4} \lesssim \dfrac{L_\lambda C_e (1+C_0)}{(1-\gamma)T^{1/4}}.
\end{equation}
The remaining terms in step $(i)$ can be similarly derived by substituting the specified parameters into the equation and using the upper bounds derived in \eqref{eq:sum_E[e_lam_t^2]} and \eqref{eq:sum_E[e_lam_t]}.
By only collecting the terms associated with $T^{-1/4}$ and ignoring the problem-independent constants, we arrive at the final bound in $(ii)$.
Finally, we remark that since $C_e = \mc{O}\left(\frac{1}{(1-\gamma)^3}\right)$, the bound in $(ii)$ has the order $\mc{O}\left(\frac{T^{-1/4}}{(1-\gamma)^4}\right)$.

Next, the constraint violation of Algorithm \ref{alg:pdpg} can be derived from  \eqref{eq:avg_L_sample} in Lemma \ref{lemma:avg_L_sample} and \eqref{eq:constraint_violation_lemma} in Lemma \ref{lemma:obtain_opt_const_sample}. Since the derivation closely resembles that of the optimality gap, we will not reiterate some intermediate steps and only write the main inequalities:
\begin{equation*}
\begin{aligned}
&\mbb{E}\left[\dfrac{1}{T} \sum_{t=0}^{T-1}-g(\lambda(\theta_t)) \right]_+\\
\leq& \dfrac{L(\lambda(\theta^\star),\mu_{0})-L(\lambda(\theta_{0}), \mu_{0})}{\varepsilon T} + \dfrac{2\eta_\theta }{\varepsilon T}\sum_{t=0}^{T-1}\mbb{E}\left[\norm{e_{L,t}}\right] +\dfrac{(1+C_0) \ell_\theta L_\lambda^2\varepsilon}{(1-\gamma)^2}\\
 &+\dfrac{1}{\varepsilon}\left[(1+C_0)\left(\dfrac{ \ell_\theta \eta_\theta^2}{2} + 2L_{\theta,\lambda} \eta_\theta \gamma^H\right) + 2\eta_\mu M^2 \right]+\dfrac{\max_{\mu\in U}\left\{(\mu_0-\mu)^2 - (\mu_T-\mu)^2\right\}}{2T\eta_\mu}\\
 &+ \eta_\mu M^2 + C_0M_\lambda \gamma^H + 2\eta_\mu M_\lambda^2 \gamma^{2H}+ \dfrac{C_0M_\lambda}{T}\sum_{t=0}^{T-1} \mbb{E}\left[\norm{e_{\lambda,t}}\right]+ \dfrac{2\eta_\mu M_\lambda^2}{T}\sum_{t=0}^{T-1} \mbb{E}\left[\norm{e_{\lambda,t}}^2\right]\\[5pt]
 \overset{(i)}{\lesssim}& \dfrac{L_\lambda M}{(1-\gamma)T^{1/2}}  +\dfrac{L_\lambda C_e (1+C_0)}{(1-\gamma)T^{1/4}} +(1+C_0)\left(\dfrac{\ell_\theta L_\lambda}{(1-\gamma)T^{1/2}} + \dfrac{L_{\theta,\lambda} L_\lambda \gamma^H}{1-\gamma}\right) +\dfrac{L_\lambda M^2}{(1-\gamma)T^{1/4}}\\
 &+\dfrac{(1+C_0)\ell_\theta L_\lambda}{(1-\gamma)T^{1/2}} + \dfrac{C_0^2}{T^{1/4}} + \dfrac{M^2}{T^{3/4}} + C_0M_\lambda \gamma^H + \dfrac{M_\lambda^2 \gamma^{2H}}{T^{3/4}}+\dfrac{C_0M_\lambda\sqrt{C_\lambda}}{T^{1/4}} + \dfrac{M_\lambda^2 C_\lambda}{T^{5/4}}\\[5pt]
 \lesssim& \left[\dfrac{C_0L_\lambda C_e}{1-\gamma} + C_0^2 + C_0M_\lambda \sqrt{C_\lambda}\right]\cdot \dfrac{1}{T^{1/4}} = \mc{O}\left(\dfrac{T^{-1/4}}{(1-\gamma)^4}\right).
\end{aligned}
\end{equation*}
where in step $(i)$, we again use the inequality $\max_{\mu\in U}\big\{(\mu_0-\mu)^2 - (\mu_T-\mu)^2\big\} = \max_{\mu\in U}\big\{(2\mu-\mu_T)\mu_T \big\} \leq C_0^2$.
This completes the proof of Theorem \ref{thm:sample}.
\end{proof}

\vspace{5pt}
\subsection{Control the Sizes of \texorpdfstring{$\norm{e_{L,t}}$}{} and \texorpdfstring{$\norm{e_{\lambda,t}}$}{}}\label{subsec:control_e}
In this section, we provide upper bounds for the approximation error terms $e_{L,t}$ and $e_{\lambda,t}$, which are modified from that of \cite{barakat2023reinforcement}.
Throughout the section, we assume that Assumption \ref{assump:lipschit_gradient} holds true and will adhere to the choices of $\alpha_t = (t+1)^{-1/2}$ and $\eta_\theta = T^{-1/2}$.
The following technical lemmas will be frequently used in this section.
\begin{lemma}\label{lemma:technical}
Suppose that the sequence $\{x_t\}_{t\geq 0}$ satisfies $x_t \leq (1-\alpha_t)^2 x_{t-1} + \beta_t$, where $\alpha_t = (t+1)^{-1/2}$ and $\beta_t \leq C (t+1)^{-1}$.
Then, it holds that
\begin{equation}\label{eq:lemma_eq}
x_t \leq \dfrac{x_0}{t+1} + \dfrac{12C}{\sqrt{t+2}} = \mc{O}\left(\dfrac{1}{\sqrt{t+2}}\right).
\end{equation}
\end{lemma}
\begin{proof}
We first unroll the recursion for $x_t$ to obtain that
\begin{equation}\label{eq:x_t<=}
\begin{aligned}
    x_t &\leq x_0 \prod_{t^\prime = 1}^t (1-\alpha_{t^\prime})^2 + \sum_{t^\prime = 1}^t \beta_{t^\prime} \cdot \prod_{t^{\prime\prime} = t^{\prime}+1}^t (1-\alpha_{t^{\prime\prime}})^2\\
    &\overset{(i)}{\leq} x_0 \prod_{t^\prime = 1}^t  (1-\alpha_{t^\prime})^2 + \sum_{t^{\prime}=1}^t \beta_{t^\prime} \cdot \exp\left(-2\sum_{t^{\prime \prime} = t^{\prime}+1}^t \alpha_{t^{\prime\prime}}\right).
\end{aligned}
\end{equation}
where step $(i)$ uses the basic inequality $1-x \leq \exp(-x)$.

We first focus on controlling the second term on the right-hand side of \eqref{eq:x_t<=}. 
We begin with bounding the summation $\sum_{t^\prime = 1}^t \alpha_{t^\prime}$ by its integration approximation. Since $\alpha_t = (t+1)^{-1/2}$ is monotone decreasing in $t$, it holds that
\begin{subequations}
\begin{align}
    \sum_{t^\prime = 1}^t \alpha_{t^\prime} &\geq \int_{1}^{t+1} (t^\prime+1)^{-\frac{1}{2}} dt^\prime = 2(t+2)^{\frac{1}{2}} - 2\sqrt{3},\label{eq:alpha_lower}\\
    \sum_{t^\prime = 1}^t \alpha_{t^\prime} &\leq \int_{0}^{t} (t^\prime+1)^{-\frac{1}{2}} dt^\prime = 2(t+1)^{\frac{1}{2}}-2.\label{eq:alpha_upper}
\end{align}
\end{subequations}
Then, we have that 
\begin{equation}\label{eq:beta_t_0}
\begin{aligned}
    \sum_{t^{\prime}=1}^t \beta_{t^\prime} \cdot \exp\Big(-2\sum_{t^{\prime \prime} = t^{\prime}+1}^t \alpha_{t^{\prime\prime}}\Big) 
    =& \exp\Big(-2\sum_{t^\prime =1}^t \alpha_{t^\prime} \Big) \cdot \sum_{t^\prime=1}^t \beta_{t^\prime} \cdot \exp\Big(2\sum_{t^{\prime\prime}=1}^{t^\prime} \alpha_{t^\prime} \Big)\\
    \overset{(i)}{\leq}& \exp\Big(-2\sum_{t^\prime =1}^t \alpha_{t^\prime} \Big) \cdot  \sum_{t^\prime=1}^t \beta_{t^\prime} \cdot \exp\left(4(t^\prime+1)^\frac{1}{2} -4\right)\\
    \leq& C  \exp\Big(\hspace{-3pt}-\hspace{-3pt}2\sum_{t^\prime =1}^t \alpha_{t^\prime} \hspace{-2pt}-\hspace{-2pt} 4 \Big) \hspace{-2pt}\cdot \hspace{-2pt}\underbrace{\sum_{t^\prime=1}^t (t^\prime \hspace{-2pt}+\hspace{-2pt}1)^{-1} \hspace{-2pt} \cdot  \hspace{-2pt}\exp\left(4(t^\prime+1)^\frac{1}{2} \right)}_{\mc T},
\end{aligned}
\end{equation}
where step $(i)$ is obtained by substituting the upper bound of $\sum_{t^{\prime\prime}=1}^{t^\prime} \alpha_{t^{\prime\prime}}$ into \eqref{eq:alpha_upper}. To further bound the term $\mc T$, we consider the function $f(t^\prime) = (t^\prime+1)^{-1} \cdot \exp\left(4(t^\prime+1)^{1/2}\right)$, whose derivative can be computed as follows
\begin{equation}
    f^{\prime}(t^\prime)=(t^\prime+1)^{-\frac{5}{2}} \cdot \exp \left(4(t^\prime+1)^{\frac{1}{2}}\right) \cdot \left(2 t^\prime-(t^\prime+1)^{\frac{1}{2}}+2\right).
\end{equation}
Hence, $f(y)$ is monotone increasing when $y\geq 0$. Therefore, the summation term $\mc T$ can be again upper-bounded by the integration:
\begin{equation}
\begin{aligned}
    \mc{T} = \sum_{t^\prime=1}^t (t^\prime+1)^{-1}\cdot \exp\left(4(t^\prime+2)^\frac{1}{2} \right) &\leq \int_1^{t+1} (t^\prime+1)^{-1}\cdot \exp\left(4(t^\prime+1)^\frac{1}{2} \right) dt^\prime\\
    &\overset{(i)}{\leq} \int_2^{t+2} y^{-1} \exp\left(4y^{\frac{1}{2}} \right) dy =:I,
\end{aligned}
\end{equation}
where in step $(i)$, we perform a change of variables and let $y:= t^\prime +1$, and then we define the entire term as $I$:
Next, our goal is to provide an upper bound for $I$.
\begin{equation}\label{eq:I}
\begin{aligned}
    I &\ = \int_{2}^{t+2} y^{-1} \cdot \dfrac{1}{2}y^{\frac{1}{2}} \cdot d\left(\exp(4y^{\frac{1}{2}})\right)\\
    &\overset{(i)}{=}\dfrac{1}{2} y^{-\frac{1}{2}} \exp\left(4y^{\frac{1}{2}}\right) \bigg\vert_{2}^{t+2} + \dfrac{1}{4} \int_{2}^{t+2} y^{-\frac{3}{2}} \exp\left(4y^{\frac{1}{2}}\right)dy\\
    &\overset{(ii)}{\leq} \dfrac{1}{2} (t+2)^{-\frac{1}{2}}\cdot \exp\left(4(t+2)^{\frac{1}{2}}\right) + \dfrac{1}{4\sqrt{2}} \underbrace{\int_{2}^{t+2}y^{-1} \exp\left(4y^{\frac{1}{2}}\right) dy}_{=I}.
\end{aligned} 
\end{equation}
where step $(i)$ involves integration by parts, and step $(ii)$ is derived as follows
\begin{equation}
    \int_{2}^{t+2} y^{-\frac{3}{2}} \exp\left(4y^{\frac{1}{2}}\right)dy  =  \int_{2}^{t+2} y^{-\frac{1}{2}}\cdot y^{-1} \exp\left(4y^{\frac{1}{2}}\right)dy \leq 2^{-\frac{1}{2}} \int_{2}^{t+2} y^{-1} \exp\left(4y^{\frac{1}{2}}\right)dy.
\end{equation}
By rearranging the terms, \eqref{eq:I} can be further simplified as 
\begin{equation}
    \left(1-\dfrac{1}{4\sqrt{2}}\right) I \leq \dfrac{1}{2}(t+2)^{-\frac{1}{2}} \cdot \exp\left(4(t+2)^{\frac{1}{2}}\right),
\end{equation}
and therefore
\begin{equation}\label{eq:beta_t_1}
    I \leq \dfrac{1}{1-\frac{1}{4\sqrt{2}}}\cdot \dfrac{1}{2}(t+2)^{-\frac{1}{2}} \cdot \exp\left(4(t+2)^{\frac{1}{2}}\right) =  \dfrac{2\sqrt{2}}{4\sqrt{2}-1}(t+2)^{-\frac{1}{2}} \cdot \exp\left(4(t+2)^{\frac{1}{2}}\right).
\end{equation}
Combining \eqref{eq:beta_t_0} and \eqref{eq:beta_t_1}, we obtain that
\begin{equation}\label{eq:beta_t_2}
\begin{aligned}
    &\sum_{t^{\prime}=1}^t \beta_{t^\prime} \cdot \exp\left(-2\sum_{t^{\prime \prime} = t^{\prime}+1}^t \alpha_{t^{\prime\prime}}\right)\\
    \leq& C  \exp\left(-2\sum_{t^\prime =1}^t \alpha_{t^\prime} -4\right) \cdot \sum_{t^\prime=1}^t (t^\prime+1)^{-1}\cdot \exp\left(4(t^\prime+1)^\frac{1}{2} \right)\\
    \overset{(i)}{\leq}& C \exp\left(-4(t+2)^{\frac{1}{2}} +4\sqrt{3}-4\right)\cdot \dfrac{2\sqrt{2}}{4\sqrt{2}-1} (t+2)^{-\frac{1}{2}} \exp\left(4(t+2)^\frac{1}{2} \right)\\
    \leq& \dfrac{12C}{\sqrt{t+2}},
\end{aligned}
\end{equation}
where in step $(i)$, we use \eqref{eq:alpha_lower} to upper-bound the term $\exp\left(-2\sum_{t^\prime =1}^t \alpha_{t^\prime} \right)$.

Finally, we consider the first term $x_0 \prod_{t^\prime = 1}^t (1-\alpha_{t^\prime})^2$ in \eqref{eq:x_t<=}. It holds that
\begin{equation}\label{eq:1-alpha}
\begin{aligned}
    1-\alpha_{t^\prime} = 1-(t^\prime+1)^{-\frac{1}{2}} \leq 1-\dfrac{1}{2}(t^\prime)^{-1} \overset{(i)}{\leq} \left(1+(t^\prime)^{-1}\right)^{-\frac{1}{2}} = \left(\dfrac{t^\prime+1}{t^\prime}\right)^{-\frac{1}{2}} = \dfrac{\alpha_{t^\prime}}{\alpha_{t^\prime-1}},
\end{aligned}
\end{equation}
where step $(i)$ is due to Bernoulli's inequality that $(1+z)^r \geq 1+r z$ for all $z\geq -1$ and $r \leq 0$. Multiplying the above term over $t^\prime = 1, \dots, t$, we have that 
\begin{equation}\label{eq:multi_1-alpha}
\prod_{t^\prime = 1}^t (1-\alpha_{t^\prime})^2 \leq \left[\prod_{t^\prime = 1}^t \dfrac{\alpha_{t^\prime}}{\alpha_{t^\prime-1}} \right]^2  = \alpha_t^2 = \dfrac{1}{t+1}.
\end{equation}
Combining \eqref{eq:x_t<=}, \eqref{eq:beta_t_2} and \eqref{eq:multi_1-alpha}, we arrive at
\begin{equation}
\begin{aligned}
    x_t &\leq x_0 \prod_{t^\prime = 1}^t  (1-\alpha_{t^\prime})^2 + \sum_{t^{\prime}=1}^t \beta_{t^\prime} \cdot \exp\left(-2\sum_{t^{\prime \prime} = t^{\prime}+1}^t \alpha_{t^{\prime\prime}}\right)\leq \dfrac{x_0}{t+1} +\dfrac{12C}{\sqrt{t+2}}.
\end{aligned}
\end{equation}
This completes the derivation of Lemma \ref{lemma:technical}.
\end{proof}

\begin{lemma}\label{lemma:control_e_lambda}
Under Algorithm \ref{alg:pdpg}, let $\left\{e_{\lambda,t}\right\}_{t=0}^{T-1}$ be the sequence defined as $e_{\lambda,t}:= \lambda_H(\theta_t) - \lambda_t$, for all $t\geq 0$.
Then, it holds that
\begin{equation}\label{eq:recursion_e}
\mbb{E}\left[ \norm{e_{\lambda,t}}^2\right]\leq (1-\alpha_t)^2\mbb{E}\left[ \norm{e_{\lambda,t-1}}^2\right] + \dfrac{2C_w \eta_\theta^2}{(1-\gamma)^2} + \dfrac{2\alpha_t^2}{(1-\gamma)^2},
\end{equation}
where $C_w := H(W+1)\left[(8H+2)M_{\pi,1}^2 + 2M_{\pi,2}\right]$.
Furthermore, by letting $\alpha_t = (t+1)^{-1}$, $\eta_\theta = T^{-1/2}$, and $C_\lambda :=\frac{C_w + 1}{(1-\gamma)^2}$, it holds that
\begin{equation}\label{eq:bound_e_lambda_0}
    \mbb{E}\left[ \norm{e_{\lambda,t}}^2\right]\leq \dfrac{25C_\lambda}{\sqrt{t+2}}\lesssim \dfrac{C_\lambda}{\sqrt{t+2}}.
\end{equation}
\end{lemma}

\begin{proof}
To show the recursion in \eqref{eq:recursion_e}, we build a connection between $e_{\lambda,t}$ and $e_{\lambda,t-1}$ as follows
\begin{equation}\label{eq:approx_lamda}
\begin{aligned}
    e_{\lambda,t} 
    & =\lambda_H(\theta_t) - \lambda_t\\
    &\overset{(i)}{=} \lambda_H(\theta_t) - \widehat{\lambda}(\tau_t) - (1-\alpha_t)\big[\lambda_{t-1} - w(\tau_t\vert \theta_{t-1},\theta_t)\widehat{\lambda}(\tau_t)\big]\\
    &\overset{(ii)}{=} (1 \hspace{-2pt}-\hspace{-2pt}\alpha_t)(\lambda_H(\theta_{t-1}) \hspace{-2pt}-\hspace{-2pt} \lambda_{t-1}) + \alpha_t y_t \hspace{-2pt}-\hspace{-2pt} (1 \hspace{-2pt}-\hspace{-2pt} \alpha_t)z_t\\
    & =(1 \hspace{-2pt}-\hspace{-2pt}\alpha_t) e_{\lambda,t-1}  + \alpha_t y_t - (1-\alpha_t) z_t,
\end{aligned}
\end{equation}
where step $(i)$ follows from the update rule of $\lambda_t$ in \eqref{eq:update_lambda} of Algorithm \ref{alg:pdpg}.
In step $(ii)$, we let
\begin{equation}\label{eq: def_ytzt}
y_t:= \lambda_H(\theta_t)- \widehat{\lambda}(\tau_t),\quad z_t:=\lambda_H(\theta_{t-1})- \lambda_H(\theta_t)+ \widehat{\lambda}(\tau_t)\left(1-w\left(\tau_t \vert \theta_{t-1}, \theta_t\right)\right).
\end{equation}
Hence, it holds from \eqref{eq:approx_lamda} that
\begin{equation}\label{eq:E_e_lambda}
\begin{aligned}
\mbb{E}\big[\norm{e_{\lambda,t}}^2\big]=(1-\alpha_t)^2\mbb{E}\big[\norm{e_{\lambda,{t-1}}}^2\big] &+\mbb{E}\big[\norm{\alpha_t y_t - (1-\alpha_t)z_t}^2\big]\\
&+\mbb{E} \big[\inner{(1-\alpha_t)e_{\lambda,{t-1}}}{\alpha_t y_t - (1-\alpha_t)z_t}\big].
\end{aligned}
\end{equation}
We first show that the third expectation term on the right-hand side of \eqref{eq:E_e_lambda} is equal to zero.
Consider the filtration $\left\{\mc{F}_t\right\}_{t\geq 0}$, where the $\sigma$-algebra $\mc{F}_t$ is generated by all trajectories of length $H$ sampled at $0,1,\dots, t$-th periods.
Therefore, for each $t$, the error term $e_{\lambda,t}$ is $\mc{F}_{t}$-measurable.
Since $\tau_t$ is sampled from policy $\pi_{\theta_t}$ and $\mbb{E}[w\left(\tau_t \vert \theta_{t-1}, \theta_t\right) \widehat{\lambda}(\tau_t)] = \lambda_H(\theta_{t-1})$ by Lemma \ref{lemma: ISweight}, we observe that  $\mbb{E}[y_t] = \mbb{E}[z_t]=\mbb{E}[{y_t|\mc{F}_{t-1}}] = \mbb{E}[{z_t|\mc{F}_{t-1}}] =0$.
By the tower rule of conditional expectation, it holds that
\begin{equation}\label{eq:E_e_lambda_1}
\begin{aligned}
   \mbb{E} \big[\hspace{-2pt} \inner{(1\hspace{-2pt}-\hspace{-2pt}\alpha_t) e_{\lambda,{t-1}}}{\alpha_t y_t \hspace{-2pt}-\hspace{-2pt} (1\hspace{-2pt}-\hspace{-2pt}\alpha_t)z_t}\hspace{-2pt} \big] & \hspace{-2pt}=\hspace{-2pt} \mbb{E}\left[\mbb{E}\left[ \inner{(1\hspace{-2pt}-\hspace{-2pt}\alpha_t)e_{\lambda,{t-1}}}{\alpha_t y_t \hspace{-2pt}-\hspace{-2pt} (1\hspace{-2pt}-\hspace{-2pt}\alpha_t)z_t}\big|\mc{F}_{t-1}\right]\right]\\
  &\hspace{-2pt}=\hspace{-2pt} \mbb{E}\left[\inner{(1\hspace{-2pt}-\hspace{-2pt}\alpha_t)e_{\lambda,{t-1}}}{\mbb{E}\left[\alpha_t y_t \hspace{-2pt}-\hspace{-2pt} (1\hspace{-2pt}-\hspace{-2pt}\alpha_t)z_t \big|\mc{F}_{t-1}\right]}\right]\\
    &\hspace{-2pt}=\hspace{-2pt} 0.
\end{aligned}
\end{equation}
Then, by \eqref{eq:E_e_lambda_1}, we can further simplify \eqref{eq:E_e_lambda} as
\begin{equation}\label{eq:E_e_lambda_2}
\mbb{E}\big[\norm{e_{\lambda,t}}^2\big]\leq(1-\alpha_t)^2\mbb{E}\big[\norm{e_{\lambda,{t-1}}}^2\big] +2\alpha_t^2 \mbb{E}\big[\norm{y_t}^2\big] + 2(1-\alpha_t)^2\mbb{E}\big[\norm{z_t}^2\big],
\end{equation}
where we apply the inequality $\norm{a+b}^2 \leq 2\norm{a}^2 + 2\norm{b}^2$.

Now, we proceed to bound the last two terms in \eqref{eq:E_e_lambda_2}:
\begin{equation}\label{eq:E_y_t}
\begin{aligned}
    \mbb{E}\big[\norm{y_t}^2\big] &= \mbb{E} \left[\big\|\lambda_H(\theta_t)- \widehat{\lambda}(\tau_t)\big\|^2\right]\\
    &= \mbb{E}\Big[\big\|\widehat{\lambda}(\tau_t)\big\|^2 -2 \big\langle\widehat{\lambda}(\tau_t),\lambda_H(\theta_t)\big\rangle+ \norm{\lambda_H(\theta_t)}^2 \Big]\\
    &=\mbb{E}\Big[\big\|\widehat{\lambda}(\tau_t)\big\|^2\Big] - 2\big\langle\mbb{E}[\widehat{\lambda}(\tau_t)],\lambda_H(\theta_t)\big\rangle +\norm{\lambda_H(\theta_t)}^2\\
    &\leq \mbb{E}\Big[\big\|\widehat{\lambda}(\tau_t)\big\|^2\Big] \leq \dfrac{1}{(1-\gamma)^2},
\end{aligned}
\end{equation}
where the last inequality stems from the fact that
\begin{equation}\label{eq:lam_bound}
    \big\|\widehat{\lambda}(\tau_t)\big\| \leq \sum_{t=0}^{H-1} \gamma^t\norm{\delta_{s_t, a_t}}=\sum_{t=0}^{H-1} \gamma^t \leq \dfrac{1}{1-\gamma}.
\end{equation}
Next, we bound the last term in \eqref{eq:E_e_lambda_2}. By the definition $z_t$ from \eqref{eq: def_ytzt} and the fact that $\mbb{E}[\widehat{\lambda}(\tau_t)\left(1-w\left(\tau_t \vert \theta_{t-1}, \theta_t\right)\right)] = \lambda_H(\theta_t)- \lambda_H(\theta_{t-1})$, we have that
\begin{equation}\label{eq:E_z_t}
\begin{aligned}
    \mbb{E}\big[\norm{z_t}^2\big] 
    &\overset{(i)}{\leq} \mbb{E}\Big[\big\|\widehat{\lambda}(\tau_t)\left(1-w\left(\tau_t \vert \theta_{t-1}, \theta_t\right)\right)\big\|^2\Big]\\
    &= \mbb{E}\Big[\left(1-w\left(\tau_t \vert \theta_{t-1}, \theta_t\right)\right)^2 \big\|\widehat{\lambda}(\tau_t)\big\|^2\Big]\\
    &\overset{(ii)}{\leq} \dfrac{1}{(1-\gamma)^2} \mbb{E}\big[\left(1-w\left(\tau_t \vert \theta_{t-1}, \theta_t\right)\right)^2\big]\\
    &=\dfrac{1}{(1-\gamma)^2}\left[\operatorname{Var}\big(w\left(\tau_t \vert \theta_{t-1}, \theta_t\right)\big) + \left(\mbb{E}\big[w\left(\tau_t \vert \theta_{t-1}, \theta_t\right)\big] - 1\right)^2 \right]\\
    &= \dfrac{1}{(1-\gamma)^2} \operatorname{Var}\big(w\left(\tau_t \vert \theta_{t-1}, \theta_t\right)\big)\\
     &\leq \dfrac{C_w \eta_\theta^2}{(1-\gamma)^2},
\end{aligned}
\end{equation}
where step $(i)$ is due to the basic fact that $\mbb{E}\big[\norm{X-\mbb{E}\brac{X}}^2\big]\leq \mbb{E}\big[\norm{X}^2\big]$ for any random variable $X$.
Then, inequality $(i)$ stems from \eqref{eq:lam_bound}, and the last two steps follow from Lemma \ref{lemma:ISweight}.
Incorporating \eqref{eq:E_e_lambda_2}, \eqref{eq:E_y_t} and \eqref{eq:E_z_t} completes the proof of \eqref{eq:recursion_e}.

When $\alpha_t = (t+1)^{-1}$ and $\eta_\theta = T^{-1/2}$, we can apply Lemma \ref{lemma:technical} to \eqref{eq:recursion_e} with $\beta_t = \frac{2C_w \eta_\theta^2}{(1-\gamma)^2} + \frac{2\alpha_t^2}{(1-\gamma)^2} \leq \frac{2C_w+2}{(1-\gamma)^2}\cdot \frac{1}{t+1}$ to derive that 
\begin{equation}\label{eq:bound_e_lambda}
\begin{aligned}
\mbb{E}\left[ \norm{e_{\lambda,t}}^2\right] 
&\leq \dfrac{\mbb{E}\left[ \norm{e_{\lambda,0}}^2\right]}{t+1} + 12\cdot \dfrac{2C_w+2}{(1-\gamma)^2}\cdot\frac{1}{\sqrt{t+2}} \\
&\overset{(i)}{\leq} \dfrac{2}{(1-\gamma)^2(t+1)} +  \dfrac{24(C_w+1)}{(1-\gamma)^2}\cdot\frac{1}{\sqrt{t+2}} \\
&\leq \dfrac{25(C_w+1)}{(1-\gamma)^2}\cdot\frac{1}{\sqrt{t+2}}\\
&\lesssim \dfrac{C_w + 1}{(1-\gamma)^2}\cdot \dfrac{1}{\sqrt{t+2}}= \dfrac{C_\lambda}{\sqrt{t+2}},
\end{aligned}
\end{equation}
where inequality $(i)$ uses the fact that $\norm{e_{\lambda,0}}^2 \leq 2\norm{\lambda_0}^2 \leq  \frac{2}{(1-\gamma)^2}$ with probability 1.
This completes the proof of Lemma \ref{lemma:control_e_lambda}.
\end{proof}

\begin{lemma}\label{lemma:bound_e_ft}
Under Algorithm \ref{alg:pdpg}, let $\left\{e_{f,t}\right\}_{t=0}^{T-1}$ be the sequence defined as $e_{f,t}:= d_{f,t} - \nabla_\theta f(\lambda_H(\theta_t))$, for all $t\geq 0$.
Then, it holds that
\begin{equation}\label{eq:E[e_ft]}
\begin{aligned}
\mbb{E}\left[\norm{e_{f,t}}\right] 
&\lesssim \dfrac{C_e}{(t+1)^{1/4}},
\end{aligned}
\end{equation}
where $C_e$ is defined in \eqref{eq:C_e}. The same upper bound also holds for $\left\{e_{g,t}\right\}_{t=0}^{T-1}$, where $e_{g,t}:= d_{g,t} - \nabla_\theta g(\lambda_H(\theta_t))$.
\end{lemma}

\begin{proof}
From the definition of $e_{f,t}$, we can decompose it as
\begin{equation}\label{eq:e_ft}
\begin{aligned}
    e_{f,t} &= d_{f,t} - \nabla_\theta f(\lambda_H(\theta_t))\\
    &=d_{f,t} - \inner{\nabla_\theta \lambda_H(\theta_t)}{r_{f,t-1}} + \inner{\nabla_\theta \lambda_H(\theta_t)}{r_{f,t-1} - \nabla_\lambda f(\lambda_H(\theta_t))},
\end{aligned}
\end{equation}
where $r_{f,t} = \nabla_\lambda f(\lambda_t)$, defined in Algorithm \ref{alg:pdpg}, is the shadow reward estimator at period $t$.
We remark that the quantities $d_{f,t}$ and $r_{f,t-1}$ in \eqref{eq:e_ft} are random, whereas other terms are deterministic.
By letting $\widehat e_{f,t}:= d_{f,t} - \inner{\nabla_\theta \lambda_H(\theta_t)}{r_{f,t-1}}$, we have from \eqref{eq:e_ft} that
\begin{equation}\label{eq:E_e_ft_1}
\mbb{E}\left[\norm{e_{f,t}}\right] \leq \mbb{E}\left[\norm{\widehat e_{f,t}}\right] + \mbb{E}\left[\norm{\inner{\nabla_\theta \lambda_H(\theta_t)}{r_{f,t-1} - \nabla_\lambda f(\lambda_H(\theta_t))} }\right].
\end{equation}
For the second term on the right-hand side of \eqref{eq:E_e_ft_1}, it holds that 
\begin{equation}\label{eq:E_e_ft_2}
\begin{aligned}
    &\inner{\nabla_\theta \lambda_H(\theta_t)}{r_{f,t-1} - \nabla_\lambda f(\lambda_H(\theta_t))}\\
    \overset{(i)}{\leq}\ & \dfrac{2M_{\pi,1}}{(1-\gamma)^2} \norm{r_{f,t-1} - \nabla_\lambda f(\lambda_H(\theta_t))}_{\infty}\\
    \leq \ & \dfrac{2M_{\pi,1}}{(1-\gamma)^2} \big(\norm{r_{f,t-1} - \nabla_\lambda f(\lambda_H(\theta_{t-1}))}_\infty +\norm{\nabla_\lambda f(\lambda_H(\theta_{t-1})) - \nabla_\lambda f(\lambda_H(\theta_t))}_\infty \big)\\
    \overset{(ii)}{\leq}& \dfrac{2M_{\pi,1}}{(1-\gamma)^2} \big(\ell_\lambda \norm{\lambda_{t-1} - \lambda_H(\theta_{t-1})} +\ell_\lambda \norm{\lambda_H(\theta_{t-1})) - \lambda_H(\theta_t)} \big)\\
    \overset{(iii)}{\leq}& \dfrac{2M_{\pi,1}}{(1-\gamma)^2} \big(\ell_\lambda \norm{\lambda_{t-1} - \lambda_H(\theta_{t-1})} + \dfrac{2\ell_\lambda M_{\pi,1}}{(1-\gamma)^2} \norm{\theta_{t-1} - \theta_t}\big)\\
    =\ &  \dfrac{2M_{\pi,1} \ell_\lambda}{(1-\gamma)^2} \norm{e_{\lambda,t-1}} + \dfrac{4M_{\pi,1}^2 \ell_\lambda}{(1-\gamma)^4} \eta_\theta.
\end{aligned}
\end{equation}
In \eqref{eq:E_e_ft_2}, step $(i)$ is derived from the fact that $\inner{\nabla_\theta \lambda_H(\theta_t)}{r} = \nabla_{\theta} \inner{\lambda_H(\theta_t)}{r}$ is the policy gradient of the truncated value function with the reward $r=r_{f,t-1} - \nabla_\lambda f(\lambda_H(\theta_t))$. 
Then, from Lemma \ref{lemma:policy gradient_general}, it holds that
\begin{equation}\label{eq:bound_lamda_r}
\begin{aligned}
    \inner{\nabla_\theta \lambda_H(\theta)}{r} 
    & \hspace{-2pt}=\hspace{-2pt} \mathbb{E}\left[\sum_{k=0}^{H-1} \gamma^k \nabla_\theta \log \pi_\theta(a_k\vert s_k) Q^{\pi_\theta}(r;s_k,a_k) \bigg\vert \pi_\theta, s_0 \hspace{-2pt} \sim \hspace{-2pt} \rho\right]
    \hspace{-2pt} \leq \hspace{-2pt} \dfrac{2M_{\pi,1}}{(1-\gamma)^2} \norm{r}_\infty,
\end{aligned}
\end{equation}
where the last inequality uses the bound on the score function $\nabla_{\theta} \log \pi_{\theta}(a \vert s)$ from Lemma \ref{lemma:parameterization} and the fact that $\left|Q^{\pi_\theta}(r;s,a)\right|\leq \frac{\norm{r}_\infty}{1-\gamma}$. 
Then, step $(ii)$ in \eqref{eq:E_e_ft_2} is by the $\ell_\lambda$-smoothness of function $f(\cdot)$ from Assumption \ref{assump:lipschit_gradient} and the definition $r_{f,t-1} = \nabla_\lambda f(\lambda_{t-1})$.
Step $(iii)$ results from the Lipschitz continuity of $\lambda_H(\theta)$ with respect to $\theta$ stated in Lemma \ref{lemma:lambda_smoothness}. 
Finally, the last step in \eqref{eq:E_e_ft_2} is by the definition $e_{\lambda,t-1} =  \lambda_H(\theta_{t-1}) - \lambda_{t-1}$ and the update rule for $\theta$ in Algorithm \ref{alg:pdpg}, i.e., $\norm{\theta_t-\theta_{t-1}} = \norm{\eta_{\theta,t-1}d_{L,t-1}} = \eta_\theta$.
Incorporating \eqref{eq:E_e_ft_1} and \eqref{eq:E_e_ft_2}, we arrive at
\begin{equation}\label{eq:bound_e_ft}
\mbb{E}\left[\norm{e_{f,t}}\right]\leq \mbb{E}\left[\norm{\widehat e_{f,t}}\right] + \dfrac{2M_{\pi,1}\ell_\lambda}{(1-\gamma)^2}\mbb{E}\left[\norm{ e_{\lambda,t-1}}\right] +\dfrac{4M_{\pi,1}^2\ell_\lambda \eta_\theta}{(1-\gamma)^4}.
\end{equation}
Next, we focus on upper-bounding the term $\mbb{E}\left[\norm{\widehat e_{f,t}}\right]$ by deriving a recursion for the sequence $\left\{\widehat e_{f,t}\right\}_{t=0}^{T-1}$. Similar to the derivation in \eqref{eq:approx_lamda}, we establish the relation between $\widehat e_{f,t}$ and $\widehat e_{f,t-1}$ as follows
\begin{equation}\label{eq:widehat e_ft}
\begin{aligned}
    \widehat e_{f,t} 
    &= d_{f,t} - \inner{\nabla_\theta \lambda_H(\theta_t)}{r_{f,t-1}}\\
    &\overset{(i)}{=}\widehat d(\tau_t,\theta_t,r_{f,t-1}) + (1-\alpha_t)\big[d_{f,t-1} - w(\tau_t\vert \theta_{t-1},\theta_t)\widehat d(\tau_t,\theta_{t-1},r_{f,t-2}) \big]\\ 
    &\quad - \inner{\nabla_\theta \lambda_H(\theta_t)}{r_{f,t-1}}\\
    &\overset{(ii)}{=} (1-\alpha_t)\prth{d_{f,t-1} - \inner{\nabla_\theta \lambda_H(\theta_{t-1})}{r_{f,t-2}}} + \alpha_t \widehat{y}_t + (1-\alpha_t)\widehat{z}_t\\
    &= (1-\alpha_t)\widehat e_{f,t-1} +\alpha_t \widehat{y}_t + (1-\alpha_t)\widehat{z}_t,
\end{aligned}
\end{equation}
where step $(i)$ is by the update rule of $d_{f,t}$ in Algorithm \ref{alg:pdpg}, and in step $(ii)$, we define that
\begin{subequations}
\begin{align}
\widehat{y}_t :=& \widehat{d}(\tau_t,\theta_t,r_{f,t-1}) - \inner{\nabla_\theta \lambda_H(\theta_t)}{r_{f,t-1}}.\label{eq:def_hat_y}\\
\widehat{z}_t :=&  \inner{\nabla_\theta \lambda_H(\theta_{t-1})}{r_{f,t-2}} - \inner{\nabla_\theta \lambda_H(\theta_{t})}{r_{f,t-1}}\\
&+ \widehat{d}(\tau_t,\theta_t,r_{f,t-1}) - w(\tau_t\vert \theta_{t-1},\theta_t)\widehat{d}(\tau_t,\theta_{t-1},r_{f,t-2})\notag. 
\label{eq:def_hat_z}
\end{align}
\end{subequations}
We notice that $\tau_t$ is collected under $\ptt$, yet both $r_{f,t-1}$ and $r_{f,t-2}$ are computed before period $t$.
Thus, the similar reasoning from \eqref{eq:E_e_lambda} to \eqref{eq:E_e_lambda_2} still applies.
We have $\mbb{E}[\widehat{y}_t] = \mbb{E}[\widehat{z}_t]= \mbb{E}[{\widehat{y}_t|\mc{F}_{t-1}}] = \mbb{E}[{\widehat{z}_t|\mc{F}_{t-1}}] =0$, and thereby \eqref{eq:widehat e_ft} further implies that
\begin{equation}\label{eq:E_widehat_e_ft}
\begin{aligned}
\mbb{E}\big[\norm{\widehat{e}_{f,t}}^2\big]
&\leq (1-\alpha_t)^2\mbb{E}\big[\norm{\widehat{e}_{f,{t-1}}}^2\big] + \mbb{E}\big[\norm{\alpha_t \widehat{y}_t + (1-\alpha_t)\widehat{z}_t}^2\big]\\
&\leq (1-\alpha_t)^2\mbb{E}\big[\norm{\widehat{e}_{f,{t-1}}}^2\big] +2\alpha_t^2 \mbb{E}\big[\norm{\widehat{y}_t}^2\big] + 2(1-\alpha_t)^2\mbb{E}\big[\norm{\widehat{z}_t}^2\big].
\end{aligned}
\end{equation}
We first focus on the term $\mbb{E}\big[\norm{\widehat{y}_t}^2\big]$ in \eqref{eq:E_widehat_e_ft}. By the definition in \eqref{eq:def_hat_y}, we observe that
\begin{equation}\label{eq:expect_hat_y_no_square}
\begin{aligned}
    \norm{\widehat{y}_t} &\leq \big\|\widehat{d}(\tau_t,\theta_t,r_{f,t-1})\big\| +\norm{\inner{\nabla_\theta \lambda_H(\theta_t)}{r_{f,t-1}}}\\
    &\overset{(i)}{\leq} \dfrac{2 M_\lambda M_{\pi,1}}{(1-\gamma)^2} + \dfrac{2 M_\lambda M_{\pi,1}}{(1-\gamma)^2}\\
    &= \dfrac{4M_\lambda M_{\pi,1}}{(1-\gamma)^2},
\end{aligned}
\end{equation}
where the second term in $(i)$ follows from the bound in \eqref{eq:bound_lamda_r} with $\norm{r}_\infty = \norm{r_{f,t-1}}_\infty = \norm{\nabla_\lambda f(\lambda_{t-1})}_\infty\leq M_\lambda$, i.e., $M_\lambda$-Lipschitz continuity of $f(\cdot)$ in Assumption \ref{assump:lipschit_gradient}.
The first term in $(i)$ comes from an upper bound on $\big\|\widehat{d}(\tau_t,\theta_t,r_{f,t-1})\big\|$, which can be derived as follows
\begin{equation}\label{eq:d_tau_132}
\begin{aligned}
    \big\|\widehat{d}(\tau_t,\theta_t,r_{f,t-1})\big\| 
    &\ \overset{(i)}{=}\norm{\sum_{h=0}^{H-1}\left(\sum_{k=h}^{H-1} \gamma^k r_{f,t-1}\left(s_k, a_k\right)\right) \nabla \log \pi_{\theta_t}\left(a_h \vert s_h\right)}\\
    &\ \leq \sum_{h=0}^{H-1} \sum_{k=h}^{H-1} \gamma^k\big\|r_{f, t-1}\big\|_{\infty} \cdot \big\|\nabla \log \pi_{\theta_t}(a_h \vert s_h)\big\|\\
    &\ \leq M_\lambda \sum_{h=0}^{H-1} \sum_{k=h}^{H-1}  \gamma^k \big\|\nabla \log \pi_{\theta_t}(a_h \vert s_h)\big\|\\
    &\overset{(ii)}{\leq} 2 M_\lambda M_{\pi,1} \sum_{h=0}^{H-1} \sum_{k=h}^{H-1} \gamma^k\\
    &\ = 2 M_\lambda M_{\pi,1} \sum_{k=0}^{H-1}\sum_{h=0}^{k} \gamma^k\\
    &\ \leq 2 M_\lambda M_{\pi,1} \sum_{k=0}^{H-1}  (k+1)\gamma^k\\
    &\ \leq \dfrac{2 M_\lambda M_{\pi,1}}{(1-\gamma)^2}.
\end{aligned}
\end{equation}
where $(i)$ is by the definition of the policy gradient estimator in \eqref{eq:pg_estimator}, and $(ii)$ applies the bound on the score function $\nabla_{\theta} \log \pi_{\theta}(a \vert s)$ from Lemma \ref{lemma:parameterization}. Hence, \eqref{eq:expect_hat_y_no_square} indicates that
\begin{equation}\label{eq:expect_hat_y_square}
    \mbb{E}\big[\norm{\widehat{y}_t}^2\big] \leq \dfrac{16(M_\lambda M_{\pi,1})^2}{(1-\gamma)^4}.
\end{equation}
\vspace{5pt}Next, we focus on the term $\mbb{E}\big[\norm{\widehat{z}_t}^2\big]$. Since $\mbb{E}\left[\widehat{z}_t\right] = 0$, it follows from the same reasoning as \eqref{eq:E_z_t} that
\begin{equation}\label{eq:E_hat_z}
\begin{aligned}
    \mbb{E}\big[\norm{\widehat{z}_t}^2\big] &\leq \mbb{E}\Big[\big\| \widehat{d}(\tau_t, \theta_t, r_{f,t-1}) - w(\tau_t\vert \theta_{t-1},\theta_t)\widehat{d}(\tau_t,\theta_{t-1},r_{f,t-2})\big\|^2 \Big]\\[8pt]
    &=\mbb{E}\Big[\big\|\widehat{d}(\tau_t, \theta_t, r_{f,t-1}) - \widehat{d}(\tau_t, \theta_{t-1}, r_{f,t-1})\\
    &\quad+ \widehat{d}(\tau_t, \theta_{t-1}, r_{f,t-1}) - \widehat{d}(\tau_t, \theta_{t-1}, r_{f,t-2})\\
    & \quad + (1-w(\tau_t\vert \theta_{t-1},\theta_t)) \widehat{d}(\tau_t,\theta_{t-1},r_{f,t-2})\big\|^2 \Big]\\[5pt]
    &\leq 3\mbb{E}\Big[\big\|\widehat{d}(\tau_t, \theta_t, r_{f,t-1}) - \widehat{d}(\tau_t, \theta_{t-1}, r_{f,t-1})\big\|^2\Big]\\
    &\quad+ 3\mbb{E}\Big[\big\|\widehat{d}(\tau_t, \theta_{t-1}, r_{f,t-1}) - \widehat{d}(\tau_t, \theta_{t-1}, r_{f,t-2})\big\|^2\Big]\\
    &\quad +3\mbb{E}\Big[(1-w(\tau_t\vert \theta_{t-1},\theta_t))^2 \big\|\widehat{d}(\tau_t,\theta_{t-1},r_{f,t-2})\big\|^2\Big].
\end{aligned}
\end{equation}
where we apply the inequality $\norm{a+b+c}^2\leq 3\norm{a}^2+3\norm{b}^2+3\norm{c}^2$ in the last step.
Below, we derive some upper bounds for the three terms on the right-hand side of \eqref{eq:E_hat_z} separately.

{\bf 1. The first term of \eqref{eq:E_hat_z}}:
\begin{equation}
\begin{aligned}    
    \big\|\widehat{d}(\tau_t, \theta_t, r_{f,t-1}) - \widehat{d}(\tau_t, \theta_{t-1}, r_{f,t-1})\big\| &\overset{(i)}{\leq} \dfrac{2(M_{\pi,1}^2+M_{\pi,2})}{(1-\gamma)^2} \norm{r_{f,t-1}}_\infty \cdot \norm{\theta_t-\theta_{t-1}}\\
    &= \dfrac{2(M_{\pi,1}^2+M_{\pi,2})}{(1-\gamma)^2} \norm{\nabla_\lambda f(\lambda_{t-1})}_\infty \cdot \eta_\theta\\
    & \leq \dfrac{2M_\lambda(M_{\pi,1}^2+M_{\pi,2})}{(1-\gamma)^2}\eta_\theta,
\end{aligned}
\end{equation}
where inequality $(i)$ follows from Lemma \ref{lemma:lambda_smoothness}, and last line is due to  $M_\lambda$-Lipschitz continuity of $f(\cdot)$ from Assumption \ref{assump:lipschit_gradient}. Therefore, with $\eta_\theta = T^{-1/2}$, we have that
\begin{equation}\label{eq:first_term}
\begin{aligned}
    \mbb{E}\Big[\big\|\widehat{d}(\tau_t, \theta_t, r_{f,t-1}) - \widehat{d}(\tau_t, \theta_{t-1}, r_{f,t-1})\big\|^2\Big] \leq& \dfrac{4M_\lambda^2 (M_{\pi,1}^2+M_{\pi,2})^2}{(1-\gamma)^4}\eta_\theta^2\\
    =& \dfrac{4M_\lambda^2 (M_{\pi,1}^2+M_{\pi,2})^2}{(1-\gamma)^4T}.
\end{aligned}
\end{equation}

{\bf 2. The second term of \eqref{eq:E_hat_z}}:
\begin{equation}\label{eq:second_term}
\begin{aligned}
    &\ \ \quad  \big\|\widehat{d}(\tau_t, \theta_{t-1}, r_{f,t-1}) - \widehat{d}(\tau_t, \theta_{t-1}, r_{f,t-2})\big\| \\
    &\ =\big\|\widehat{d}(\tau_t, \theta_{t-1}, r_{f,t-1} - r_{f,t-2})\big\|\\
    &\ \overset{(i)}{\leq} \sum_{h=0}^{H-1} \sum_{k=h}^{H-1} \gamma^k\big\|r_{f, t-1}-r_{f,t-2}\big\|_{\infty} \cdot \big\|\nabla \log \pi_{\theta_{t-1}}(a_h \vert s_h)\big\|\\
    &\overset{(ii)}{\leq} \dfrac{2M_{\pi,1}}{(1-\gamma)^2} \norm{r_{f,t-1} - r_{f,t-2}}_\infty\\
    &\ = \dfrac{2M_{\pi,1}}{(1-\gamma)^2} \norm{\nabla_\lambda f(\lambda_{t-1}) - \nabla_\lambda f(\lambda_{t-2})}_\infty\\
    &\overset{(iii)}{\leq} \dfrac{2M_{\pi,1}\ell_\lambda}{(1-\gamma)^2} \norm{\lambda_{t-1} - \lambda_{t-2}},
\end{aligned}
\end{equation}
where steps $(i)$ and $(ii)$ follow from the same reasoning as \eqref{eq:d_tau_132}, and step $(iii)$ uses the $\ell_\lambda$-smoothness of $f(\cdot)$ from Assumption \ref{assump:lipschit_gradient}.
Utilizing \eqref{eq:second_term}, we obtain that 
\begin{equation}\label{eq:second_term_final}
    \mbb{E}\Big[\big\|\widehat{d}(\tau_t, \theta_{t-1}, r_{f,t-1}) - \widehat{d}(\tau_t, \theta_{t-1}, r_{f,t-2})\big\|^2 \Big] \leq \dfrac{4M_{\pi,1}^2 \ell_\lambda^2}{(1-\gamma)^4} \cdot \mbb{E} \big[\norm{\lambda_{t-1} - \lambda_{t-2}}^2\big].
\end{equation}
Hence, it remains to show the boundedness of $\mbb{E} \big[\norm{\lambda_{t-1} - \lambda_{t-2}}^2\big]$. Based on the update rule of $\lambda_t$, we observe that
\begin{equation}
\begin{aligned}
    \lambda_{t} - \lambda_{t-1} &= \widehat{\lambda}(\tau_t) + (1-\alpha_t)\big[\lambda_{t-1} - w(\tau_t\vert \theta_{t-1},\theta_t)\widehat{\lambda}(\tau_t)\big] - \lambda_{t-1}\\
    &= \alpha_t(\widehat{\lambda}(\tau_t) - \lambda_{t}) + \alpha_t(\lambda_{t} - \lambda_{t-1}) + (1-\alpha_t)(1-w(\tau_t\vert \theta_{t-1},\theta_t))\widehat{\lambda}(\tau_t).
\end{aligned}
\end{equation}
By rearranging the terms, we obtain that
\begin{equation}
\begin{aligned}
    \lambda_{t} - \lambda_{t-1} &= \dfrac{\alpha_t}{1-\alpha_t}(\widehat{\lambda}(\tau_t) - \lambda_{t}) +(1-w(\tau_t\vert \theta_{t-1},\theta_t))\widehat{\lambda}(\tau_t)\\
    &=\dfrac{\alpha_t}{1-\alpha_t} (\lambda_H(\theta_t) - \lambda_t) + \dfrac{\alpha_t}{1-\alpha_t}(\widehat{\lambda}(\tau_t) - \lambda_H(\theta_t)) + (1-w(\tau_t\vert \theta_{t-1},\theta_t))\widehat{\lambda}(\tau_t)\\
    &= \dfrac{\alpha_t}{1-\alpha_t} e_{\lambda,t} + \dfrac{\alpha_t}{1-\alpha_t}(\widehat{\lambda}(\tau_t) - \lambda_H(\theta_t)) + (1-w(\tau_t\vert \theta_{t-1},\theta_t))\widehat{\lambda}(\tau_t).
\end{aligned}
\end{equation}
Using the above equality, we further arrive at
\begin{equation}\label{eq:E_lam-lam}
\begin{aligned}
    &\quad \mbb{E}\big[\norm{\lambda_{t} - \lambda_{t-1}}^2\big] \\
    &\leq \dfrac{3\alpha_t^2 \mbb{E}\big[\norm{e_{\lambda,t}}^2\big]}{(1-\alpha_t)^2}  + \dfrac{3\alpha_t^2 \mbb{E} \Big[\big\|\lambda_H(\theta_t) - \widehat{\lambda}(\tau_t)\big\|^2\Big] }{(1-\alpha_t)^2} +3\mbb{E}\Big[\big\|(1-w(\tau_t\vert \theta_{t-1},\theta_t))\widehat{\lambda}(\tau_t)\big\|^2\Big]\\
    &\overset{(i)}{\leq}\dfrac{75\alpha_t^2 C_\lambda}{(1-\alpha_t)^2\sqrt{t+2}} + \dfrac{3 \alpha_t^2}{(1-\alpha_t)^2(1-\gamma)^2} +\dfrac{3C_w \eta_\theta^2}{(1-\gamma)^2}\\
    & \overset{(ii)}{\leq} \dfrac{75C_\lambda}{(t+1)^{3/2}} + \dfrac{3}{(1-\gamma)^2(t+1)} + \dfrac{3C_w}{(1-\gamma)^2T}\\
    &\leq \dfrac{78C_\lambda}{t+1}\\ 
    &\lesssim \dfrac{C_\lambda}{t+1},
\end{aligned}
\end{equation}
where in step $(i)$, we respectively use the bound for $\mbb{E}\big[\norm{e_{\lambda,t}}^2\big]$ from \eqref{eq:bound_e_lambda_0} in Lemma \ref{lemma:control_e_lambda}, the bound for $\mbb{E} \Big[\big\|\lambda_H(\theta_t) - \widehat{\lambda}(\tau_t)\big\|^2\Big]$ in \eqref{eq:E_y_t}, and the bound for $\mbb{E}\Big[\big\|(1-w(\tau_t\vert \theta_{t-1},\theta_t))\widehat{\lambda}(\tau_t)\big\|^2\Big]$ in \eqref{eq:E_z_t}.
Then, inequality $(ii)$ comes from the choices of $\alpha_t = (t+1)^{-1/2}$ and $\eta_\theta = T^{-1/2}$.
Applying \eqref{eq:E_lam-lam} to \eqref{eq:second_term_final}, we obtain the following inequality for all $t\geq 1$:
\begin{equation}\label{eq:second_term_final_final}
    \mbb{E}\Big[\big\|\widehat{d}(\tau_t, \theta_{t-1}, r_{f,t-1}) - \widehat{d}(\tau_t, \theta_{t-1}, r_{f,t-2})\big\|^2 \Big] \leq \dfrac{4(M_{\pi,1} \ell_\lambda)^2}{(1-\gamma)^4} \cdot \dfrac{78C_\lambda}{t}\leq \dfrac{624(M_{\pi,1} \ell_\lambda)^2 C_\lambda}{(1-\gamma)^4(t+1)}.
\end{equation}
The above upper bound also apply to the case $t=0$, where the term $\widehat{d}(\tau_t, \theta_{t-1}, r_{f,t-1}) - \widehat{d}(\tau_t, \theta_{t-1}, r_{f,t-2})$ actually does not show up in Eq. \eqref{eq:E_hat_z}.

\vspace{5pt}
{\bf 3. The third term of \eqref{eq:E_hat_z}}:
\begin{equation}\label{eq:third_term}
\begin{aligned}
    \mbb{E}\Big[(1-w(\tau_t\vert \theta_{t-1},\theta_t))^2 \big\|\widehat{d}(\tau_t,\theta_{t-1},r_{f,t-2})\big\|^2\Big] &\overset{(i)}{\leq} \dfrac{4(M_\lambda M_{\pi,1})^2}{(1-\gamma)^4} \cdot \mbb{E}\big[(1-w(\tau_t\vert \theta_{t-1},\theta_t))^2\big]\\
    &\overset{(ii)}{\leq}  \dfrac{4(M_\lambda M_{\pi,1})^2 C_w \eta_\theta^2}{(1-\gamma)^4}\\
    &= \dfrac{4(M_\lambda M_{\pi,1})^2 C_w }{(1-\gamma)^4T},
\end{aligned}
\end{equation}
where step $(i)$ follows from \eqref{eq:d_tau_132}, and step $(ii)$ applies Lemma \ref{lemma:ISweight} in a similar way as \eqref{eq:E_z_t}.

Combining the upper bounds for the three terms shown in \eqref{eq:first_term}, \eqref{eq:second_term_final_final}, and \eqref{eq:third_term}, we finalize the bound for $\mbb{E}\big[\norm{\widehat{z}_t}^2\big]$ in \eqref{eq:E_hat_z} as follows
\begin{equation}\label{eq:expect_hat_z_square}
\begin{aligned}
    \mbb{E}\big[\norm{\widehat{z}_t}^2\big] &\leq \dfrac{12 M_\lambda^2(M_{\pi,1}^2 +M_{\pi,2})^2}{(1-\gamma)^4T} + \dfrac{1872(M_{\pi,1} \ell_\lambda)^2 C_\lambda}{(1-\gamma)^4(t+1)} +  \dfrac{12(M_\lambda M_{\pi,1})^2 C_w }{(1-\gamma)^4T}.
\end{aligned}
\end{equation}
Finally, by substituting the bounds for $\mbb{E}\big[\norm{\widehat{y}_t}^2\big]$ in \eqref{eq:expect_hat_y_square} and $\mbb{E}\big[\norm{\widehat{z}_t}^2\big]$ in \eqref{eq:expect_hat_z_square} into \eqref{eq:E_widehat_e_ft}, we conclude that 
\begin{equation}\label{eq:e_hat_recursion}
\begin{aligned}
    \hspace{-20pt}\mbb{E}\big[\norm{\widehat{e}_{f,t}}^2\big] &\leq(1-\alpha_t)^2\mbb{E}\big[\norm{\widehat{e}_{f,{t-1}}}^2\big] + 2\alpha_t^2  \cdot \dfrac{16(M_\lambda M_{\pi,1})^2}{(1-\gamma)^4}\\
    &\quad + \hspace{-2pt}2(1\hspace{-2pt}-\hspace{-2pt}\alpha_t)^2\left(\hspace{-2pt}\dfrac{12 M_\lambda^2(M_{\pi,1}^2 \hspace{-2pt}+\hspace{-2pt} M_{\pi,2})^2}{(1-\gamma)^4T} \hspace{-2pt}+\hspace{-2pt} \dfrac{1872(M_{\pi,1} \ell_\lambda)^2 C_\lambda}{(1-\gamma)^4(t+1)} \hspace{-2pt}+\hspace{-2pt}  \dfrac{12(M_\lambda M_{\pi,1})^2 C_w }{(1-\gamma)^4T}\hspace{-2pt}\right)\\
    &\leq (1-\alpha_t)^2\mbb{E}\big[\norm{\widehat{e}_{\lambda,{t-1}}}^2\big] + \dfrac{32(M_\lambda M_{\pi,1})^2}{(1-\gamma)^4} \cdot \dfrac{1}{t+1}\\
   &\quad + \dfrac{24}{(1-\gamma)^4} \left(\dfrac{M_\lambda^2\left[C_w M_{\pi,1}^2  +(M_{\pi,1}^2+M_{\pi,2})^2\right]}{T} +\dfrac{156(M_{\pi, 1} \ell_\lambda)^2 C_\lambda}{t+1}\right).
\end{aligned}
\end{equation}
Therefore, $\mbb{E}\left[ \norm{\widehat e_{f,t}}^2\right]$ satisfies the recursion $\mbb{E}\left[ \norm{\widehat e_{f,t}}^2\right] \leq (1-\alpha_t)^2\mbb{E}\left[ \norm{\widehat e_{f,t-1}}^2\right] + \beta_t$ with
\begin{equation}
\beta_t \leq \underbrace{\dfrac{32 (M_\lambda M_{\pi,1})^2 + 24\left\{M_\lambda^2\left[C_wM_{\pi,1}^2 + (M_{\pi,1}^2+M_{\pi,2})^2 \right]  + 156(M_{\pi,1}\ell_\lambda)^2C_\lambda \right\}}{(1-\gamma)^4}}_{=:C_\beta}\cdot \dfrac{1}{t+1},
\end{equation}
where we define $C_\beta$ to be the coefficient of $1/(t+1)$ in the above upper bound.
Hence, by applying Lemma \ref{lemma:technical}, we have that
\begin{equation}\label{eq:bound_hat_e_ft}
\begin{aligned}
\mbb{E}\left[ \norm{\widehat e_{f,t}}^2\right] 
&\ \leq \dfrac{\mbb{E}\left[ \norm{\widehat e_{f,0}}^2\right]}{t+1} + \dfrac{12C_\beta}{\sqrt{t+2}}\\
&\overset{(i)}{\leq} \dfrac{16(M_\lambda M_{\pi,1})^2}{(1-\gamma)^4(t+1)}+ \dfrac{12C_\beta}{\sqrt{t+2}}\\
&\ \lesssim \underbrace{\dfrac{M_\lambda^2\left[(1+C_w)M_{\pi,1}^2 + (M_{\pi,1}^2+M_{\pi,2})^2 \right] +(M_{\pi,1} \ell_\lambda)^2 C_\lambda}{(1-\gamma)^4}}_{=:C_{\widehat e}}\cdot \dfrac{1}{\sqrt{t+2}},
\end{aligned}
\end{equation}
where step $(i)$ uses the upper bound $\norm{\widehat e_{f,0}} \leq \norm{d_{f,0}} + \norm{ \inner{\nabla_\theta \lambda_H(\theta_0)}{r_{f,-1}}}\leq \frac{4M_\lambda M_{\pi,1}}{(1-\gamma)^2}$ with the derivations identical to \eqref{eq:expect_hat_y_no_square}.

Lastly, combining \eqref{eq:bound_e_lambda}, \eqref{eq:bound_e_ft}, and \eqref{eq:bound_hat_e_ft}, we conclude that
\begin{equation}
\begin{aligned}
& \hspace{18pt} \mbb{E}
\left[\norm{e_{f,t}}\right]\\[3pt] 
&\lesssim \dfrac{\sqrt{C_{\widehat e}}}{(t+2)^{1/4}} + \dfrac{M_{\pi,1}\ell_\lambda \sqrt{C_\lambda}}{(1-\gamma)^2}\cdot \dfrac{1}{(t+1)^{1/4}} + \dfrac{M_{\pi,1}^2\ell_\lambda }{(1-\gamma)^4}\cdot \dfrac{1}{\sqrt{T}}\\
&\lesssim \left(\sqrt{C_{\widehat e}} + \dfrac{M_{\pi,1}\ell_\lambda \sqrt{C_\lambda}}{(1-\gamma)^2} \right)\dfrac{1}{(t+1)^{1/4}}\\
&=: \underbrace{\dfrac{\sqrt{M_\lambda^2\left[(1 \hspace{-2pt}+\hspace{-2pt} C_w)M_{\pi,1}^2 \hspace{-2pt}+\hspace{-2pt} (M_{\pi,1}^2 \hspace{-2pt}+\hspace{-2pt} M_{\pi,2})^2 \right] +(M_{\pi,1} \ell_\lambda)^2 C_\lambda} + M_{\pi,1}\ell_\lambda \sqrt{C_\lambda}}{(1-\gamma)^2}}_{=:C_e} \cdot \dfrac{1}{(t+1)^{1/4}}.
\end{aligned}
\end{equation}
Since function $g(\lambda(\theta))$ satisfies the same property as $f(\lambda(\theta))$, we know that the same upper bound also applies to $\norm{e_{g,t}}$.
This completes the proof of Lemma \ref{lemma:bound_e_ft}.
\end{proof}

\begin{lemma}\label{lemma:E_elt_bound}
Under Algorithm \ref{alg:pdpg}, let $\left\{e_{L,t}\right\}_{t=0}^{T-1}$ be the sequence defined as $e_{L,t}:= d_{L,t} - \nabla_\theta L(\lambda_H(\theta_t),\mu_t)$, for all $t\geq 0$.
Then, it holds that
    \begin{equation}\label{eq:E[e_lt]}
\begin{aligned}
    \mbb{E}\left[\norm{e_{L,t}}\right] = \mbb{E}\left[\norm{e_{f,t} + \mu_t e_{g,t}}\right] \leq \mbb{E}\left[\norm{e_{f,t}}\right] + \mu_t\mbb{E}\left[\norm{e_{g,t}}\right] 
    \lesssim \dfrac{C_e (1+C_0)}{(t+1)^{1/4}},
\end{aligned}
\end{equation}
where $C_e$ is defined in \eqref{eq:C_e}.

\begin{proof}
To control $e_{L,t}$, we notice the following decomposition:
\begin{equation}\label{eq:e_lt}
\begin{aligned}
e_{L,t} &= d_{L,t} - \nabla_\theta L(\lambda_H(\theta_t),\mu_t)\\
&= (d_{f,t} + \mu_t d_{g,t}) - \left[\nabla_\theta f(\lambda_H(\theta_t)) + \mu_t \nabla_\theta g(\lambda_H(\theta_t))\right]\\
&= \left[ d_{f,t} - f(\lambda_H(\theta_t)) \right] + \mu_t \left[d_{g,t} - g(\lambda_H(\theta_t))\right]\\
&= e_{f,t} + \mu_t e_{g,t}.
\end{aligned}
\end{equation}
Then, according to Lemma \ref{lemma:bound_e_ft}, we have that
\begin{equation}
\begin{aligned}
    \mbb{E}\left[\norm{e_{L,t}}\right] = \mbb{E}\left[\norm{e_{f,t} + \mu_t e_{g,t}}\right] \leq \mbb{E}\left[\norm{e_{f,t}}\right] + \mu_t\mbb{E}\left[\norm{e_{g,t}}\right] 
    \overset{(i)}{\lesssim} \dfrac{C_e (1+C_0)}{(t+1)^{1/4}},
\end{aligned}
\end{equation}
where step $(i)$ is by the fact that $\mu_t \in [0, C_0]$. This completes the proof of Lemma \ref{lemma:E_elt_bound}.
\end{proof}
\end{lemma}

%% file: app/appendix_5.tex
\vspace{10pt}
\section{Supplementary Materials for Section \ref{sec:zero}}\label{app:zero}
\subsection{Proof of Theorem \ref{thm:zero}}
\begin{proof}
Recall the pessimistic problem \eqref{eq:prob_pessimistic_new} and the modified algorithm in \eqref{eq:algorithm_update_zero}. We begin with general arguments that apply to both concave and strongly concave cases.
Firstly, by the assumption $\delta<\xi$, it holds that
\begin{equation}\label{eq:new_bound_pessimistic}
    g_\delta(\lambda(\theta)) = g(\lambda(\theta))- \delta \ \ \Rightarrow \ \ \left|g_\delta(\lambda(\theta))\right|\leq M+\xi \ \text{ and } g_\delta(\lambda(\widetilde\theta))\geq  \xi-\delta > 0,
\end{equation}
which gives the constraint upper bound and strict feasibility for the pessimistic problem \eqref{eq:prob_pessimistic_new}.

Let $\theta_\delta^\star$ be an optimal solution to the pessimistic problem (\ref{eq:prob_pessimistic_new}). Then,
\begin{equation}\label{eq:zero_violation_difference}
\begin{aligned}
&\dfrac{1}{T} \sum_{t=0}^{T-1} \left[f(\lambda(\theta^\star)) - f(\lambda(\theta_t))\right]\\
=& \dfrac{1}{T} \sum_{t=0}^{T-1} \left[f(\lambda(\theta^\star)) - f(\lambda(\theta_\delta^\star))\right] +\dfrac{1}{T} \sum_{t=0}^{T-1} \left[f(\lambda(\theta^\star_\delta)) - f(\lambda(\theta_t))\right]\\
=& \left[f(\lambda(\theta^\star)) - f(\lambda(\theta_\delta^\star))\right] + \dfrac{1}{T} \sum_{t=0}^{T-1} \left[f(\lambda(\theta^\star_\delta)) - f(\lambda(\theta_t))\right].
\end{aligned}
\end{equation}
To upper-bound the first term in (\ref{eq:zero_violation_difference}), let $\theta_\delta$ be a policy parameter that satisfies
\begin{equation}\label{eq:lambda_theta_delta}
\lambda(\theta_\delta) := \dfrac{\xi-\delta}{\xi}\lambda(\theta^\star) +\dfrac{\delta}{\xi}\lambda(\widetilde\theta),
\end{equation}
where we assume $0< \delta< \xi$.
It is easy to verify that $\theta_\delta$ is a feasible point to \eqref{eq:prob_pessimistic_new}:
\begin{equation*}
\begin{aligned}
g_\delta(\lambda(\theta_\delta)) &= g\left(\lambda(\theta_\delta)\right) - \delta\\
&=g\left(\dfrac{\xi-\delta}{\xi}\lambda(\theta^\star) +\dfrac{\delta}{\xi}\lambda(\widetilde\theta)\right)-\delta\\
&\ {\overset{(i)}{\geq}} \dfrac{\xi-\delta}{\xi}g\left(\lambda(\theta^\star)\right) + \dfrac{\delta}{\xi}g\left(\lambda(\widetilde\theta) \right)-\delta\\
&\ {\overset{(ii)}{\geq}} 0 + \dfrac{\delta}{\xi}\cdot \xi - \delta\\
&\ =0,
\end{aligned}
\end{equation*}
where $(i)$ follows from the concavity of $g(\cdot)$ and $(ii)$ uses the feasibility of $\theta^\star$ to (\ref{eq:problem_theta}) as well as Assumption \ref{assump:slater}.
We remark that the policy corresponds to $\lambda(\theta_\delta)$ is unique and given by 
$$
\pi_{\theta_\delta}(a\vert s) = \dfrac{\lambda(\theta_\delta;s,a)}{\sum_{a^\prime \in \mc{A}}\lambda(\theta_\delta;s,a^\prime)}.
$$
On the other hand, due to the assumption of over-parameterization (see Assumption \ref{assump:parameterization}), $\theta_\delta$ may not be unique, and it suffices to choose one such $\theta_\delta$ that satisfies (\ref{eq:lambda_theta_delta}).

The feasibility of $\theta_\delta$ to (\ref{eq:prob_pessimistic_new}) implies that $f(\lambda(\theta_\delta^\star))\geq f(\lambda(\theta_\delta))$.
Consequently, 
\begin{equation}\label{eq:F-F_theta_delta}
\begin{aligned}
f(\lambda(\theta^\star)) - f(\lambda(\theta_\delta^\star))&\leq f(\lambda(\theta^\star))-f(\lambda(\theta_\delta))\\
&=f\left(\lambda(\theta^\star)\right)-f\left(  \dfrac{\xi-\delta}{\xi}\lambda(\theta^\star) +\dfrac{\delta}{\xi}\lambda(\widetilde\theta)\right)\\
&\leq f\left(\lambda(\theta^\star)\right)- \left(\dfrac{\xi-\delta}{\xi}f\left(\lambda(\theta^\star)\right) + \dfrac{\delta}{\xi}f\left(\lambda(\widetilde\theta) \right)\right)\\
&=\dfrac{\delta}{\xi} \left[f\left(\lambda(\theta^\star)\right) - f\left(\lambda(\widetilde\theta) \right)\right]\\
&\leq \dfrac{2\delta M}{\xi}.
\end{aligned}
\end{equation}
By \eqref{eq:new_bound_pessimistic}, for any $\delta<\xi$, choosing $C_{0,\delta}\hspace{-2pt}=1+\left(M-f(\lambda(\widetilde \theta))\right)/{(\xi-\delta)}$ ensures that the optimal dual variable of problem \eqref{eq:prob_pessimistic_new} belongs to the dual feasible region $U=[0,C_{0,\delta}]$ (see Lemma \ref{lemma:duality}).

Now, we show the first part of Theorem \ref{thm:zero}. When $f(\cdot)$ is a general concave function, we apply \eqref{eq:optimality_gap} and \eqref{eq:constraint_violation} in Theorem \ref{thm:exact} to obtain that
\begin{equation}
\begin{aligned}
&\dfrac{1}{T} \sum_{t=0}^{T-1} \left[f(\lambda(\theta_\delta^\star)) - f(\lambda(\theta_t))\right] \leq \dfrac{2M +(M \hspace{-2pt}+\hspace{-2pt}\xi)^2/2}{T^{2/3}} + \dfrac{2 (1+C_{0,\delta})\ell_\theta L_\lambda^2}{(1-\gamma)^2T^{1/3}}+\dfrac{2(M\hspace{-2pt}+\hspace{-2pt}\xi)^2}{T^{1/3}},\\
&\dfrac{1}{T}\left[\sum_{t=0}^{T-1} - g_\delta(\lambda(\theta_t))\right]_+ \leq \dfrac{2M+(M\hspace{-2pt}+\hspace{-2pt}\xi)^2/2}{T^{2/3}} + \dfrac{2(1+C_{0,\delta})\ell_\theta L_\lambda^2}{(1-\gamma)^2T^{1/3}}+ \dfrac{2 (M\hspace{-2pt}+\hspace{-2pt}\xi)^2 + C_{0, \delta}^2/2}{T^{1/3}},
\end{aligned}
\end{equation}
where we replace the upper bound $M$ on $|g(\cdot)|$ by the new upper bound $M+\xi$ for $|g_\delta(\cdot)|$.
Therefore, together with (\ref{eq:zero_violation_difference}) and (\ref{eq:F-F_theta_delta}), we have the following optimality gap for the original problem \eqref{eq:problem_theta}:
\begin{equation}\label{eq:opt_zero_proof}
\begin{aligned}
&\dfrac{1}{T} \sum_{t=0}^{T-1} \left[f(\lambda(\theta^\star)) - f(\lambda(\theta_t))\right]
\\
\leq& \dfrac{2\delta M}{\xi} + \dfrac{2M+(M+\xi)^2/2}{T^{2/3}} + \dfrac{2 (1+C_{0,\delta})\ell_\theta L_\lambda^2}{(1-\gamma)^2T^{1/3}}+\dfrac{2(M+\xi)^2}{T^{1/3}}\\
=& \mc{O}(\delta) + \mc{O}\left(T^{-1/3}\right).
\end{aligned}
\end{equation}
For the constraint violation, we have that
\begin{equation}\label{eq:const_zero_proof}
\begin{aligned}
&\dfrac{1}{T}\left[\sum_{t=0}^{T-1} -g(\lambda(\theta_t))\right]_+\\
=& \dfrac{1}{T}\left[\sum_{t=0}^{T-1}-\big(g(\lambda(\theta_t))-\delta\big)-\delta\right]_+\\
=& \left[\dfrac{1}{T}\left[\sum_{t=0}^{T-1}-\big(g(\lambda(\theta_t))-\delta\big)\right]_+-\delta\right]_+\\
=&\left[\dfrac{1}{T}\left[\sum_{t=0}^{T-1}-g_\delta(\lambda(\theta_t))\right]_+-\delta\right]_+\\
\leq& \left[\dfrac{2M+(M+\xi)^2/2}{T^{2/3}} + \dfrac{2(1+C_{0,\delta})\ell_\theta L_\lambda^2}{(1-\gamma)^2T^{1/3}}+ \dfrac{2 (M+\xi)^2 + C_{0, \delta}^2/2}{T^{1/3}}-\delta\right]_+.
\end{aligned}
\end{equation}
By choosing $\delta$ such that
\begin{equation}\label{eq:delta_convex}
\dfrac{2M+(M+\xi)^2/2}{T^{2/3}} + \dfrac{2(1+C_{0,\delta})\ell_\theta L_\lambda^2}{(1-\gamma)^2T^{1/3}}+ \dfrac{2 (M+\xi)^2 + C_{0, \delta}^2/2}{T^{1/3}}-\delta = 0,
\end{equation}
the constraint violation \eqref{eq:const_zero_proof} becomes 0.
Such constant $\delta$ must exist since the summation of the first three terms in \eqref{eq:delta_convex} decreases in the order of $\mc{O}(T^{-1/3})$ as $T$ increases.
As \eqref{eq:delta_convex} implies $\delta =\mc{O}(T^{-1/3})$, the convergence rate of the optimality gap (\ref{eq:opt_zero_proof}) remains $\mc{O}(T^{-1/3})$.
Finally, we remark that the requirement $\delta<\xi$ is naturally satisfied when $T$ is reasonably large.

\vspace{5pt}
When $f(\cdot)$ is $\sigma$-strongly concave, we apply \eqref{eq:optimality_gap_sc} and \eqref{eq:constraint_violation_sc} in Theorem \ref{thm:exact} to obtain that
\begin{equation*}
\begin{aligned}
&\dfrac{1}{T} \sum_{t=0}^{T-1} \left[f(\lambda(\theta^\star)) - f(\lambda(\theta_t))\right] \leq \dfrac{2M + C_{0,\delta}(M+\xi)}{\tilde\varepsilon T} +  \left(\dfrac{(M+\xi)^2}{\tilde\varepsilon}+ \dfrac{(M+\xi)^2}{2}\right)\dfrac{1}{\sqrt{T}},\\
&\dfrac{1}{T}\left[\sum_{t=0}^{T-1}-g(\lambda(\theta_t))\right]_+ \leq \dfrac{2M+C_{0,\delta}(M+\xi)}{\tilde\varepsilon T} +  \left(\dfrac{(M+\xi)^2}{\tilde\varepsilon}+ \dfrac{(M+\xi)^2+C_{0,\delta}^2}{2}\right)\dfrac{1}{\sqrt{T}}.
\end{aligned}
\end{equation*}
We note that, similar to \eqref{eq: sum_f_strongly_concave}, the numerator term $2M+C_{0,\delta}(M+\xi)$ comes from the new upper bound on the term $L_\delta(\lambda(\theta_T),\mu_{T-1})-L_\delta(\lambda(\theta_{0}), \mu_{0})$.
Together with (\ref{eq:zero_violation_difference}) and (\ref{eq:F-F_theta_delta}), we derive the optimality gap as
\begin{equation}\label{eq:opt_zero_sc_proof}
\begin{aligned}
&\dfrac{1}{T} \sum_{t=0}^{T-1} \left[f(\lambda(\theta^\star)) - f(\lambda(\theta_t))\right]\\
\leq& \dfrac{2\delta M}{\xi}+\dfrac{2M + C_{0,\delta}(M+\xi)}{\tilde\varepsilon T} +  \left(\dfrac{(M+\xi)^2}{\tilde\varepsilon}+ \dfrac{(M+\xi)^2}{2}\right)\dfrac{1}{\sqrt{T}}\\
=& \mc{O}(\delta) + \mc{O}\left(T^{-1/2}\right).
\end{aligned}
\end{equation}
For the constraint violation, similarly as (\ref{eq:const_zero_proof}), we have that
\begin{equation}\label{eq:const_zero_sc_proof}
\begin{aligned}
&\dfrac{1}{T}\left[\sum_{t=0}^{T-1}-g(\lambda(\theta_t))\right]_+\\
\leq& 
\left[\dfrac{2M+C_{0,\delta}(M+\xi)}{\tilde\varepsilon T} +  \left(\dfrac{(M+\xi)^2}{\tilde\varepsilon}+ \dfrac{(M+\xi)^2+C_{0,\delta}^2}{2}\right)\dfrac{1}{\sqrt{T}} -\delta\right]_+.
\end{aligned}
\end{equation}
We choose $\delta$ such that
\begin{equation}\label{eq:delta_sc}
\dfrac{2M+C_{0,\delta}(M+\xi)}{\tilde\varepsilon T} +  \left(\dfrac{(M+\xi)^2}{\tilde\varepsilon}+ \dfrac{(M+\xi)^2+C_{0,\delta}^2}{2}\right)\dfrac{1}{\sqrt{T}} -\delta = 0,
\end{equation}
which guarantees the constraint violation in \eqref{eq:const_zero_sc_proof} to be zero.
As \eqref{eq:delta_sc} implies $\delta =\mc{O}(T^{-1/2})$, the convergence rate of the optimality gap \eqref{eq:opt_zero_sc_proof} is $\mc{O}(T^{-1/2})$. This completes the proof of Theorem \ref{thm:zero}.
\end{proof}

%% file: app/appendix_aux.tex
\vspace{5pt}
\section{Auxiliary Lemmas}\label{app:aux}
In this section, we present a few auxiliary lemmas that we need for proof of the main results in this paper.
These lemmas are standard results on Markov decision processes. We refer the reader to Section \ref{sec:formulation} for necessary definitions and \cite{agarwal2021theory} for the proofs of these results.

\begin{lemma}[Policy gradient under general parameterization]\label{lemma:policy gradient_general}
Let $V^{\pi_\theta}(r)$ be the value function under policy $\pi_\theta$ with an arbitrary reward function $r:\mc{S}\times \mc{A}\rightarrow \mbb{R}$. The gradient of $V^{\pi_\theta}(r)$ with respect to $\theta$ can be given by the following three equivalent forms:
\begin{equation*}
\begin{aligned}
\nabla_{\theta} V^{\pi_{\theta}}\left(r\right)&=\dfrac{1}{1-\gamma} \mathbb{E}_{s \sim d^{\pi_{\theta}}} \mathbb{E}_{a \sim \pi_{\theta}(\cdot \vert s)}\left[\nabla_{\theta} \log \pi_{\theta}(a \vert s)\cdot Q^{\pi_{\theta}}(r;s, a)\right]\\
&=\mathbb{E}\left[\sum_{k=0}^\infty \gamma^k \nabla_\theta \log \pi_\theta(a_k\vert s_k)\cdot Q^{\pi_\theta}(r;s_k,a_k) \bigg\vert \pi_\theta, s_0 \sim \rho\right]\\
&=\mathbb{E}\left[\sum_{k=0}^\infty \gamma^k \cdot r(s_k,a_k)\cdot \left(\sum_{k^\prime = 0}^k \nabla_\theta \log \pi_\theta(a_{k^\prime}\vert s_{k^\prime}) \right)\bigg\vert \pi_\theta, s_0 \sim \rho\right]
\end{aligned}
\end{equation*}
\end{lemma}

\begin{lemma}[Policy gradient under direct parameterization]\label{lemma:policy gradient_direct}
Let $V^\pi(r)$ be the value function under policy $\pi$ with an arbitrary reward function $r:\mc{S}\times \mc{A}\rightarrow \mbb{R}$. The gradient of $V^\pi(r)$ with respect to $\pi$ is given by
\begin{equation*}
\frac{\partial V^{\pi}(r)}{\partial \pi(a \vert s)}=\frac{1}{1-\gamma} d^{\pi}(s)\cdot  Q^{\pi}(r;s, a),\ \forall \  (s,a)\in \mc{S}\times \mc{A}.
\end{equation*}
\end{lemma}

\begin{lemma}[Smoothness of $V^\pi(r)$ w.r.t. $\pi$]\label{lemma:value function smoothness}
Let $V^\pi(r)$ be the value function under policy $\pi$ with an arbitrary reward function $r:\mc{S}\times \mc{A}\rightarrow \mbb{R}$.
For every two policies $\pi$ and $\pi^\prime$, it holds that
\begin{equation*}
\left\|\nabla_{\pi} V^{\pi}\left(r\right)-\nabla_{\pi} V^{\pi^{\prime}}\left(r\right)\right\| \leq \frac{4 \gamma|\mathcal{A}|}{(1-\gamma)^{3}}\cdot \|r\|_\infty\cdot \left\|\pi-\pi^{\prime}\right\|.
\end{equation*}
\end{lemma}

\begin{lemma}[Performance difference]\label{lemma:performance difference}
Let $V^\pi(r)$ be the value function under policy $\pi$ with an arbitrary reward function $r:\mc{S}\times \mc{A}\rightarrow \mbb{R}$.
For every two policies $\pi$ and $\pi^\prime$, it holds that
\begin{equation*}
\begin{aligned}
V^{\pi^{\prime}}(r)-V^{\pi}(r) &= \left\langle r, \lambda^{\pi^\prime} - \lambda^\pi\right\rangle\\
&=\frac{1}{1-\gamma} \sum_{s\in\mc{S}} d^{\pi}(s) \sum_{a\in\mc{A}}\left(\pi^{\prime}(a \vert s)-\pi(a \vert s)\right) \cdot Q^{\pi^{\prime}}(r;s, a)\\
&=\frac{1}{1-\gamma}\sum_{s\in \mc{S}}d^{\pi^\prime}(s)\sum_{a\in\mc{A}}\pi^\prime(a\vert s) \cdot A^\pi(r;s,a),
\end{aligned}
\end{equation*}
where $A^{\pi}\left(r; s, a\right)$ denotes the advantage function with reward $r(\cdot,\cdot)$ under policy $\pi$, defined as
\begin{equation}\label{eq:advantage function}
A^{\pi}\left(r; s, a\right):= Q^\pi(r;s,a) - \mbb{E}_{a\sim \pi(\cdot\vert s)}\brac{Q^\pi(r;s,a)},\ \forall (s,a)\in \mc{S}\times\mc{A}.
\end{equation}
\end{lemma}

\begin{lemma}\label{lemma:parameterization}
Under the general soft-max parameterization \eqref{eq:policy_parameterization}, let \eqref{eq:assum_para} of Assumption \ref{assump:parameterization} hold. Then, for every $\theta \in \Theta$ and $(s,a) \in \mathcal{S} \times \mathcal{A}$, we have that 
\begin{align}
    \norm{\nabla_\theta\log \pi_\theta (a\vert s)} &\leq 2M_{\pi, 1};\\
    \norm{\nabla_\theta^2 \log \pi_\theta (a\vert s)} &\leq 2(M_{\pi, 1}^2 + M_{\pi, 2}).
\end{align}
\end{lemma}

\begin{lemma}\label{lemma:lambda_smoothness}
Let Assumptions \ref{assump:parameterization} and \ref{assump:lipschit_gradient} hold. Then, for every $\theta_1, \theta_2 \in \mbb R^k$, we have that
\begin{align}
    &\norm{\lambda_H(\theta_1) - \lambda_H(\theta_2)}\leq \dfrac{2M_{\pi,1}}{(1-\gamma)^2}\norm{\theta_1-\theta_2};\\
    &\big\|\widehat{d}(\tau, \theta_1, r) - \widehat{d}(\tau, \theta_2,r)\big\| \leq \dfrac{2(M_{\pi, 1}^2 + M_{\pi, 2})}{(1-\gamma)^2} \norm{r}_\infty \cdot\norm{\theta_1 - \theta_2}, \ \forall r \in \mbb R^{|\mc{S}|\times |\mc{A}|}.
\end{align}
\end{lemma}

%% file: app/appendix_exp.tex
\vspace{5pt}
\section{Experiment Details}
\label{appendix:experiment}
This section elaborates on the details of the environment and experimentation from Section \ref{sec: Numerical Experiment}.

\subsection{Details on Environments}

The environments are $8 \times 8$ and $20 \times 20$ tabular MDPs where $(0,0)$ is a fixed initial state (marked as $S$) and they have a fixed goal point at $(7,7)$ and $(19,19)$, respectively (See Figure \ref{fig:gridworld8x8_env} for the coordinate). For the coordination, we set the upper left point as the origin point, i.e., $(0,0)$, and each block has a length of $1$. The agent executes one of the four actions of up, left, right, and down. For every step, the agent receives a $-0.2$ reward, and if it reaches the goal state, it receives a $+100$ reward.

\vspace{5pt}
\subsection{Details on Experiments}
We elaborate on the experiment's hyperparameters in this subsection.
\subsubsection{$8 \times 8$ gridworld}
\begin{itemize}
    \item Iteration number $T= 10000$.
    \item Trajectory length $H=14$.
    \item Gamma $\gamma = 0.9$.
    \item Initial primal step-size $\eta_\theta \in \{ 0.23,0.24,0.25\}$.
    \item Initial dual step-size $\eta_\mu = 0.1$.
    \item Step size $\alpha_t = 0.1$ for all $ t \geq 1$.
    \item Initial dual variable $\mu_0 = 1$.
    \item Dual variable interval upper bound $C_0= 10$ (Line 14 of Algorithm \ref{alg:pdpg}).
    \item Constraint right-hand side value $d_0 \in \{ 0.001, 0.005, 0.01, 0.05 \}$ (see \eqref{eq:exp_tracking}).
\end{itemize}

\vspace{5pt}
\subsubsection{$20 \times 20$ gridworld}
\begin{itemize}
    \item Iteration number $T= 10000$.
    \item Trajectory length $H=38$.
    \item Gamma $\gamma = 0.9$.
    \item Initial primal step-size $\eta_\theta=0.22 $.
    \item Initial dual step-size $\eta_\mu \in \{ 0.01, 0.02, 0.03, 0.04 ,0.05\}$.
    \item Step size $\alpha_t \in \{ 0.03, 0.05, 0.07, 0.09 \}$ for all $t\geq 1$.
    \item Initial dual variable $\mu_0 = 1$.
    \item Dual variable interval upper bound $C_0 = 10$ (Line 14 of Algorithm \ref{alg:pdpg}).
    \item Constraint right-hand side value $d_0 \in \{ 0.001, 0.005, 0.01 \}$ (see \eqref{eq:exp_tracking}).
\end{itemize}